\documentclass[a4paper, 11pt]{article}
\usepackage[american]{babel}

\usepackage{authblk}
\usepackage{booktabs}
\usepackage{enumitem}
\usepackage{multirow}

\RequirePackage{fancyhdr}
\RequirePackage{xcolor} 
\RequirePackage{algorithm}
\RequirePackage{algorithmic}
\RequirePackage{eso-pic} 
\RequirePackage{forloop}
\RequirePackage{url}

\usepackage[mono=false]{libertine}
\usepackage[top=2.5cm, bottom=2.5cm, left=2cm, right=2cm]{geometry}

\usepackage{natbib} 
\usepackage{xfrac}
\usepackage{amsmath, amsfonts, amssymb}
\usepackage{bbm}
\renewcommand{\vec}[1]{\boldsymbol{#1}}
\newcommand{\mat}[1]{\boldsymbol{#1}}
\newcommand{\dd}{{\rm d}}
\newcommand{\RR}{\mathbb{R}}

\DeclareMathOperator{\prox}{prox}

\DeclareMathOperator{\Bias}{Bias}
\DeclareMathOperator{\Var}{Var}
\DeclareMathOperator{\Pois}{Pois}

\newcommand{\channel}{\vec{g}_{\rm out}}

\newcommand{\denoiser}{\vec{f}_a}

\newcommand{\reals}{\mathbb{R}}

\newcommand{\mean}[2]{\mathbb{E}_{#1}\left[#2\right]}

\newcommand{\dataset}{\mathcal{D}}

\newcommand{\hatw}{\hat{\Vec{\theta}}}
\newcommand{\werm}{\hat{\vec{\theta}}_{\lambda}}
\newcommand{\wstar}{\Vec{\theta}_{\star}}

\newcommand{\bo}{\rm bo}
\newcommand{\pb}{\rm pb}
\newcommand{\rb}{\rm rb}
\newcommand{\rr}{\rm rr}
\newcommand{\fr}{\rm fr}
\newcommand{\jk}{\rm jk}

\newcommand{\Ss}{\rm ss}

\newcommand{\varianceOnXY}{\Var_{\dataset}(\hat{\vec{\theta}}_{\lambda})}
\newcommand{\varianceOnY}{\Var_{\dataset|\mat{X}}(\hat{\vec{\theta}}_{\lambda})}
\newcommand{\variancePairBootstrap}{\widehat{\Var_{\pb}}}
\newcommand{\varianceResidualBootstrap}{\widehat{\Var_{\rb}}}
\newcommand{\varianceSubsampling}{\widehat{\Var_{\Ss}}}
\newcommand{\varianceJackknife}{\widehat{\Var_{\jk}}}
\newcommand{\varianceBO}{\Var_{\bo}}

\newcommand{\biasOnXY}{\Bias^2_{\dataset}(\hat{\vec{\theta}}_{\lambda})}
\newcommand{\biasOnY}{\Bias^2_{\dataset|\mat{X}}(\hat{\vec{\theta}}_{\lambda})}
\newcommand{\biasPairBootstrap}{\widehat{\Bias^2_{\pb}}}
\newcommand{\biasResidualBootstrap}{\widehat{\Bias^2_{\rb}}}
\newcommand{\biasSubsampling}{\widehat{\Bias^2_{\Ss}}}
\newcommand{\biasJackknife}{\widehat{\Bias^2_{\jk}}}

\usepackage{microtype}
\usepackage{graphicx}
\usepackage{subfigure}
\usepackage{booktabs} 

\usepackage{hyperref}
\hypersetup{%
    colorlinks, breaklinks=true,    urlcolor=orange, linkcolor=blue, citecolor=blue
}

\usepackage{amsmath}
\usepackage{amssymb}
\usepackage{mathtools}
\usepackage{amsthm}

\usepackage[capitalize,noabbrev]{cleveref}

\usepackage{pgfplots}
\pgfplotsset{compat=newest}
\pgfplotsset{scaled y ticks=false}
\usepgfplotslibrary{groupplots}
\usepgfplotslibrary{dateplot}
\usepackage{tikz}

\title{Analysis of Bootstrap and Subsampling \\ in High-dimensional Regularized Regression}

\author[1]{Lucas Clart\'e}
\author[1,2]{Adrien Vandenbroucque}
\author[1,2,3]{Guillaume Dalle}
\author[4]{Bruno Loureiro}
\author[2]{Florent Krzakala}
\author[1]{Lenka Zdeborov\'a}

\affil[1]{
\'Ecole Polytechnique F\'ed\'erale de Lausanne (EPFL)\\
Statistical Physics of Computation laboratory\\
CH-1015 Lausanne, Switzerland
}
\affil[2]{
\'Ecole Polytechnique F\'ed\'erale de Lausanne (EPFL)\\
Information, Learning and Physics laboratory\\
CH-1015 Lausanne, Switzerland
}
\affil[3]{
\'Ecole Polytechnique F\'ed\'erale de Lausanne (EPFL)\\
Information and Network Dynamics laboratory\\
CH-1015 Lausanne, Switzerland
}
\affil[4]{
D\'epartement d'Informatique, \'Ecole Normale Sup\'erieure - PSL \& CNRS, 45 rue d’Ulm, F-75230 Paris cedex 05, France
}

\theoremstyle{plain}
\newtheorem{theorem}{Theorem}[section]
\newtheorem{proposition}[theorem]{Proposition}

\theoremstyle{definition}

\theoremstyle{remark}
\newtheorem{remark}[theorem]{Remark}

\date{}

\begin{document}


\maketitle

\begin{abstract}
We investigate popular resampling methods for estimating the uncertainty of statistical models, such as subsampling, bootstrap and the jackknife, and their performance in high-dimensional supervised regression tasks. We provide a tight asymptotic description of the biases and variances estimated by these methods in the context of generalized linear models, such as ridge and logistic regression, taking the limit where the number of samples $n$ and dimension $d$ of the covariates grow at a comparable fixed rate $\alpha\!=\! n/d$. Our findings are three-fold: i) resampling methods are fraught with problems in high dimensions and exhibit the double-descent-like behavior typical of these situations; ii) only when $\alpha$ is large enough do they provid
e consistent and reliable error estimations (we give convergence rates); iii) in the over-parametrized regime $\alpha\!<\!1$ relevant to modern machine learning practice, their predictions are not consistent, even with optimal regularization. 
\end{abstract}
%
\section{Introduction}
\label{sec:intro}

Estimating and quantifying errors is a central aspect of statistical practice. Nevertheless, a solid understanding of how uncertainty can be reliably quantified in modern machine learning practice is largely missing, despite being a key endeavor towards a reliable use of these methods across sensitive applications. This paper delves into a comprehensive mathematical analysis of conventional resampling methods to estimate uncertainty, such as subsampling, the bootstrap and the jackknife, specifically in the context of high-dimensional regression and classification tasks. 

Let $Z_{1},\cdots, Z_{n}\sim p_{\theta}$ denote $n$ independent samples from a parametric probability distribution. Given an estimator $\hat{\theta}$ of $\theta$ (e.g. the maximum likelihood estimator), one is interested not only in the absolute performance of $\hat{\theta}$ but also in estimating how reliable it is, e.g. error bars. In particular,  even if the estimator is consistent, i.e. $\hat{\theta}\!\to\!\theta$ when $n\!\to\!\infty$, having access only to a finite amount of data $n$ introduces uncertainty in our estimation $\theta$. A central question in statistics is \emph{how to quantify this uncertainty} \citep{wasserman2004all}.

A classical family of non-parametric methods developed to address this question are \emph{resampling methods} \citep{tibshirani1993introduction,james2023resampling}, which consist in estimating the statistics of interest from the empirical distribution $p_{n} = \sfrac{1}{n}\sum_{i=1}^{n}\delta_{Z_{i}}$. Our goal is to investigate the statistical properties of three popular resampling methods in the context of the most widespread machine learning task: \emph{supervised learning}. Here the samples are given by pairs $Z_{i} = (\vec{x}_{i}, y_{i})$ from a joint distribution $p_{\theta}(\vec{x},y)$, with $\vec{x}_{i}\in\mathbb{R}^{d}$ being the covariates and $y_{i}\in\mathcal{Y}\subset\mathbb{R}$ the labels. Given the parameter $\hat \theta$ learned by a fitting model, say ridge or logistic regression, the goal is to estimate the actual bias and variance of $\hat \theta$. 

We focus on the {\it high-dimensional} regime, where both the number of samples $n$ and their dimension $d$ are comparatively large, with a fixed ratio $\alpha=n/d$. We provide a tight asymptotic description of the biases and variances estimated by resampling methods for generalized linear models, such as ridge and logistic regression or any M-estimator. We show that resampling methods are fraught with problems in high-dimensions, either overestimating or underestimating the mean and variances. Reliable error estimation can only be reached in the regime when $\alpha\gg 1$, for which we provide asymptotic rates of convergences. However, in the overparametrized regime $\alpha < 1$, relevant to modern machine learning practice, the predictions of resampling methods are clearly off, even when optimally regularizing.

\section{Setting \& motivation}
\label{sec:setting}
We consider the class of generalized linear estimation problems, where the goal is to estimate a parameter $\vec{\theta}_{\star}\in\mathbb{R}^{d}$ from $n$ independent samples $\mathcal{D}=\{(\vec{x}_{i},y_{i})_{i\in[n]}\}$ drawn from the following distribution:
\begin{align}
    \label{eq:def_model}
    y_{i}\sim p(\cdot|\vec{\theta}_{\star}^{\top}\vec{x}_{i}), && \vec{x}_{i}\sim\mathcal{N}(0,\sfrac{1}{d}\mat{I}_{d})
\end{align}
for a general likelihood $p(y|z)$. Therefore, in this case, the joint distribution reads $p_{\vec{\theta}_{\star}}(\vec{x},y) = p(y|\vec{\theta}_{\star}^{\top}\vec{x})p(\vec{x})$. For concreteness, we assume $\vec{\theta}_{\star}\sim\mathcal{N}(0,\mat{I}_{d})$. In the following, we focus on the (regularized) maximum likelihood estimator:
\begin{align}
    \label{eq:def_erm}
    \hat{\vec{\theta}}_{\lambda}(\mathcal{D}) = \underset{\vec{\theta}\in\mathbb{R}^{d}}{\rm argmin}\sum\limits_{i=1}^{n}-\log p\left(y_{i}|\vec{\theta}^{\top}\vec{x}_{i}\right) + \frac{\lambda}{2}\|\vec{\theta}\|^{2}_{2}
\end{align}
also known as \emph{empirical risk minimizer} in the context of supervised machine learning, where the loss function coincides with minus the empirical log-likelihood: ${\ell(y,z) = -\log p(y|z)}$. When it is clear from the context, we omit the training data dependence $\mathcal{D}$ in the MLE estimator and write $\hat{\vec{\theta}}_{\lambda}$.

We will focus on two particular examples of generalized linear estimation: ridge and logistic regression. Ridge regression is a regression problem $\mathcal{Y}=\mathbb{R}$, which corresponds to the Gaussian likelihood $p(y|z) = \mathcal{N}(z|y, \Delta)$ (or equivalently the square loss function $\ell(y,z)=\frac{1}{2\Delta}(y-z)^{2}$) for $\Delta>0$. Instead, logistic regression is a binary classification problem $\mathcal{Y}=\{-1,+1\}$ which corresponds to a logit likelihood $p(y|z) = \sigma(yz)$ for $\sigma(t) = (1+e^{-t})^{-1}$ the logistic function (this corresponds to the logistic or cross-entropy loss function $\ell(y,z) = \log(1+e^{-yz})$).

Note that the estimation problem introduced above is well-specified, and therefore enjoys strong mathematical guarantees in the classical statistical regime where $n\to\infty$ at fixed $d$. For instance, a well-known result is the asymptotic normality of the MLE for $\lambda=0$ \citep{wasserman2004all}: 
\begin{align}
\label{eq:consistency}
    \sqrt{n}\left(\hat{\vec{\theta}}_{0} - \vec{\theta}_{\star}\right) \overset{(d)}{\to} \mathcal{N}(0, \mathcal{I}^{-1}), && n\to\infty
\end{align}
where $\mathcal{I}\in\mathbb{R}^{d\times d}$ is the Fisher information matrix, in particular implying consistency and calibration of the maximum likelihood estimator. However, those guarantees break down when the number of samples is comparable with the dimension of the covariates $n=\Theta(d)$. This is precisely the regime of interest in our work, and applying it to resampling methods will be our goal in the following.

\subsection{What statisticians want}
``Bias'' and ``variance'' depend on the underlying data sampling process, and therefore, different notions co-exist, whether one takes, for instance, a frequentist or Bayesian viewpoint. Below, we define these different quantities, which resampling methods try to approximate.

\paragraph{Frequentist bias and variance --- } In the classical frequentist approach, the statistician seeks to estimate the bias and variance with respect to the data sampling process. This induces the classical \emph{bias-variance decomposition} of the mean squared error for the estimator $\hat{\vec{\theta}}_{\lambda}$:
\begin{equation}
    {\rm MSE}(\werm)\!=\!\frac 1d  \mean{\dataset, \wstar}{\|\werm -\wstar  \|^2} \!\!=\!  \biasOnXY \!+\! \varianceOnXY \nonumber 
\end{equation}
with: 
\begin{align}
    \biasOnXY &=  \frac1d\left\lVert\mean{\dataset, \wstar}{\werm} - \vec{\theta}_{\star}\right\lVert^2 \label{eq:def_true_bias}\\
   \varianceOnXY &= \frac1d\mean{\dataset, \wstar}{\left\lVert\werm-\mean{\dataset, \wstar}{\werm}\right\lVert^{2}}. \label{eq:def_true_variance}
\end{align}
We emphasize that in this case, the expectations are taken with respect to sampling of the full data set ${\mathcal{D} = \{(\vec{x}_{i},y_{i})_{i\in[n]}\}\sim p^{\otimes n}_{\wstar}}$.

\paragraph{Conditional bias and variance --- }
Alternatively, in a supervised learning setting one can define the bias and variance only with respect to the sampling of the labels $y_{i}\sim p(\cdot|\vec{x}_{i}^{\top}\vec{\theta}_{\star})$, i.e. conditionally on the covariates $\vec{x}_{i}$. This is known as a \emph{fixed design} analysis. We will refer to the corresponding notions as {\it conditional} bias and variance:
\begin{align}
    \biasOnY &=  \frac1d\left\lVert\mathbb{E}_\mathcal{D}[\hat{\vec{\theta}}_{\lambda}|\mat{X}] - \vec{\theta}_{\star}\right\lVert^2\label{eq:def_true_cond_bias}\\
   \varianceOnY &= \frac1d\mathbb{E}_{\mathcal{D}}\left\lVert\hat{\vec{\theta}}_{\lambda}-\mathbb{E}[\hat{\vec{\theta}}_{\lambda}|\mat{X}]\right\lVert^{2},\label{eq:def_true_cond_variance}
\end{align}
where for convenience we defined the covariate matrix $\mat{X}\in\mathbb{R}^{n\times d}$ with rows given by the covariates $\vec{x}_{i}\in\mathbb{R}^{d}$.

\paragraph{Bayesian estimator and variance ---}
Finally, it is natural to compare the maximum likelihood estimator above with the best estimator (in mean squared error) conditioned on the full training data $\mathcal{D}$, also known as the \emph{Bayes-optimal} estimator. It requires, however, the knowledge of the {\it a priori} distribution of the ``true'' weights.
\begin{align}
    \hat{\vec{\theta}}_{\rm bo} = \underset{\hat{\vec{\theta}}\in\mathbb{R}^{d}}{\rm argmin}~\mathbb{E}\left[\lVert \hat{\vec{\theta}} - \vec{\theta}_{\star}\lVert^{2}\right] = \mathbb{E}[\vec{\theta}|\mathcal{D}]
\end{align}
where the conditional expectation is taken with respect to the posterior distribution:
\begin{align}
\label{eq:def_bo}
    p(\vec{\theta}|\mathcal{D}) \propto \mathcal{N}(\vec{\theta}|0,\mat{I}_{d})\prod\limits_{i=1}^{n} p(y_{i}|\vec{\theta}^{\top}\vec{x}_{i}) 
\end{align}
Note that, by definition, $\hat{\vec{\theta}}_{\rm bo}$ is an unbiased and calibrated estimator of $\vec{\theta}_{\star}$ \citep{clarte2023theoretical}. Nevertheless, it captures the irreducible variance due to the fact we have a finite sample $\mathcal{D}$ of the population distribution: 

\begin{equation}
    \varianceBO = \frac1d\mathbb{E}\left[\left\lVert \vec{\theta} -\vec{\theta}_{\rm bo} \right\lVert^{2}|\mathcal{D}\right]
\end{equation}
where, again, the expectation is taken over the posterior distribution $p(\vec{\theta}|\mathcal{D})$.

\subsection{Resampling estimates} \label{sec:resampling_estimates}
A central problem in statistics is the estimation of the biases \eqref{eq:def_true_bias} \& \eqref{eq:def_true_cond_bias} and variances \eqref{eq:def_true_variance} \& \eqref{eq:def_true_cond_variance}, which involve population expectations, from a finite number of samples $\mathcal{D}=\{(\vec{x}_{i},y_{i})_{i\in[n]}\}$. Resampling methods are a popular class of statistical procedures that fit a family of estimators $\hatw_{k} \equiv \werm(\mathcal{D}_{k}^{\star})$ from resampled data $\mathcal{D}^{\star}_{k}$ generated from the original samples $\mathcal{D}=\{(\vec{x}_{i},y_{i})_{i\in[n]}\}$, and from which the bias and variance of $\hat{\vec{\theta}}_{\lambda}$ can be estimated:
\begin{align}
    \widehat{\Bias}^{2} &= \frac1d\left\lVert \frac{1}{B}\sum\limits_{k=1}^{B}\hatw_{k} - \werm \right\lVert^{2}, \label{eq:def:bias}\\ 
    \widehat{\Var} &= \frac{1}{dB}\sum\limits_{k=1}^{B}\left\lVert \hatw_{k}-
    \frac{1}{B}\sum\limits_{k=1}^{B}\hatw_{k}\right\lVert^{2}\label{eq:def:var}
\end{align}
In this work, we will focus on the following methods:
\begin{description}
    \item[- Pair bootstrap:] Consists in resampling $\mathcal{D}_{k}^{\star}$ from $\mathcal{D}$ with sample replacements, or in other words, sampling ${\mathcal{D}^{\star}_{k} = \{(\vec{x}^{\star}_{k,i},y^{\star}_{k,i})_{i\in[n]}\}\sim p^{\otimes n}_{n}}$ from the empirical distribution. For simplicity, we always assume $B=n$ for the bootstrap in the following.
    
    \item[- Residual bootstrap:] Akin to the pair bootstrap method, but for the conditional distribution $p(y|z)$. In practice, one first fits an estimator $\hat{\vec{\theta}}_{\lambda}(\mathcal{D})$ on the original samples (the MLE \eqref{eq:def_erm} in our setting), and given a statistical model for $\hat{p}(y|z)$, one resamples only the labels from $\hat{p}(y|\hat{\vec{\theta}}_{\lambda}(\mathcal{D})^\top\vec{x}_i)$, generating new datasets $\dataset^{\star}_k = \{ \Vec{x}_i, y^{\star}_{k,i} \}_{i = 1}^n$. This allows for the estimation of conditional statistical errors. 
    
    \item[- Subsampling:] Consists of generating new datasets $\mathcal{D}_{k}^{\star}$ of a smaller size $\lfloor r  n \rfloor$  by subsampling $\mathcal{D}$ without replacement, where $r\in(0,1)$. While bootstrap creates datasets of the right size but from the wrong distribution (as elements of $\dataset$ are duplicated), subsampling relies on data of the wrong size but from the right distribution.\footnote{Since $\mathcal{D}_{k}^{\star}$ are independent conditionally on $\mathcal{D}$.}
    
    \item[- Jackknife:] Consists of creating $B=n$ datasets $\mathcal{D}^{\star}_{k} = \{(\vec{x}_{i}, y_{i})_{i\neq k}\}$, each of which leaves a single sample out. Note that when $n\!\to\!\infty$, as in our high-dimensional regime, this is equivalent to subsampling with $r\!\to\!1$.
\end{description}
For notational convenience, we will refer to these statistics  as $\widehat{\Bias^2_k}, \widehat{\Var_t}$ with $t\!\in\!\{\pb, \rb, \Ss, \jk\}$ for  pair (pb) and residual bootstrap (rb), subsampling (ss) and jackknife (jk). \looseness=-1

\section{Contributions \& related work}
The resampling methods above have been widely studied in the classical statistical literature, with whole books dedicated to proving their mathematical soundness \citep{10.1214/aos/1176344552, 10.1214/ss/1177013815, Davison_Hinkley_1997}. However, as discussed in \cref{sec:setting} most of the classical guarantees hold in the regime where the quantity of data $n$ available to the statistician is large in comparison with data dimension $d$ --- a regime that falls short in the context of modern machine learning practice. Of particular importance was the work of \citet{ElKaroui2018} who have pointed out the lack of consistency of the bootstrap method for {\it unregularized} least squares, in the {\it underparametrized regime} $n>d$.  One of our goals in this manuscript is to fill the gap, providing a complete evaluation of the aforementioned methods (beyond bootstrap), including the effect of regularization and over-parametrization. 

More precisely, our \textbf{main contributions} are:
\begin{itemize}
    \item We provide a closed-form expression for the biases and variances in the proportional high-dimensional limit where $n,d \to \infty$ at fixed rate $\alpha=n/d$ for all the cases discussed in \cref{sec:setting}: the pair and residual biases and variances and their bootstrap, subsample, and jackknife estimates. Our result holds for generic log-concave likelihoods (corresponding to convex losses) and convex regularizers. 
    
    \item Our formulas are derived from mapping to an Approximate Message Passing (AMP) scheme admitting a rigorous asymptotic characterization in terms of \emph{state evolution} equations \citep{bayati2011dynamics,bayati2011lasso,JMLR:v15:javanmard14a,emami2020generalization,loureiro2021learning}. We believe this derivation has an interest on its own, as we show how simultaneously tracking {\it coupled} AMP trajectories provides the biases and variances for all the resampling methods. Our construction is quite generic and can be extended to other variants of interest.
    \item Our examination into the effectiveness and limitations of these methods yields three key insights. Firstly, we demonstrate that resampling techniques face significant challenges in high-dimensional contexts, resulting in a double-descent behavior typical of such scenarios. Secondly, we find that these methods yield consistent and reliable error estimates only when the ratio $\alpha$ is sufficiently large, for which we also present convergence rates. Thirdly, in the overparametrized regime where $\alpha\!<\!1$, the predictions remain inconsistent despite optimal regularization.
\end{itemize}

\paragraph{Further related work ---} Resampling methods are a classical topic in statistics. The jackknife method was introduced in \citet{6c956df0-ca97-3419-9961-dcc097853946}, refined by \citet{10.1214/aoms/1177706647} and analysed by \citet{efron1981jackknife}. Bootstrap was introduced by \citet{10.1214/aos/1176344552}, and studied in the context of least squares estimation in \citet{10.1214/aos/1176345638, 10.1214/aos/1176350142}.

The impact of high-dimensionality for the bootstrap method was first investigated by \citep{ElKaroui2018} in the context of unregularized $M$-estimation with $n>d$, where it was shown that methods that pair bootstrap under-estimates the true variance, while residual bootstrap overestimates it. 

The asymptotic theory of high-dimensional statistical generalized linear problems has witnessed a burst of activity over the last decades. Pioneered by the statistical physics community in the late 80s \citep{Gardner_1989, Opper_1990, NIPS1991_8eefcfdf, PhysRevA.45.6056, 6796373}, it is now an established field of research encompassing applications to machine learning, statistics, and signal processing among others \citep{bayati2011lasso, Karoui2013a, Donoho2016, pmlr-v40-Thrampoulidis15, thrampoulidis2018precise, 10.1214/17-AOS1549, sur_modern_2018, 10.1214/18-AOS1789, pmlr-v125-gerbelot20a, 9745052, loureiro2021learning, Loureiro_2022, Bellec2023, Bellec2023b}. 
Bayes-optimal generalization guarantees for generalized linear models were established by \citet{6566160, PhysRevX.2.021005, doi:10.1073/pnas.1802705116, NEURIPS2020_7ec0dbee}.
\cite{candes_phase_2018} have shown that, besides not being well-defined when $n<d$, the unregularized maximum likelihood estimator is biased \citep{Karoui2013a, Karoui2013b, Bean2013,sur_modern_2018, Bellec2022b} for $n>d$. One consequence is that the variance of the MLE underestimates the true variance of $\vec{\theta}_{\star}$, leading to an overconfident prediction \citep{bai2021dont, bai2021understanding, clarte2023theoretical}. Indeed, \citet{clarte2023theoretical, clarte2022overparametrized} highlighted the importance of properly regularizing the MLE in the high-dimensional regime, showing that cross-validation over $\lambda$ can mitigate some of these issues. \citet{clarte2023ec} showed that post-training \textit{temperature scaling} can mitigate overconfidence, regardless of the regularization used.

Bagging (the combination of subsampling with ensembling) has been studied in the high-dimensional regime by \citep{NIPS1995_1019c809, PhysRevE.55.811, pmlr-v108-lejeune20b, JMLR:v24:23-0887, pmlr-v202-du23d, pmlr-v202-chen23am, ando2023highdimensional, patil2023asymptotically}. Ensembling has also been investigated in the context of the random features model as a tool to decouple the different sources of randomness \citep{pmlr-v119-d-ascoli20a, JMLR:v22:20-1211, NEURIPS2020_7d420e2b, Loureiro2022_ensembling}. The performance of \textit{AdaBoost} \cite{Schapire1999Brief} and its link to minimum $\ell_1$-norm classifiers were studied in \cite{Liang2022Precise}. The performance of bootstrap averaging has been studied in the context of Gaussian Processes and Support Vector Machines using the replica method by \cite{NIPS2002_f12ee973, NIPS2003_2c6ae45a}. A replicated AMP algorithm for computing bootstrap averages of GLMs was proposed by \cite{takahashi2019replicated} and studied in the context of LASSO \citep{JMLR:v20:18-109} and Elastic Net \citep{takahashi2023role}.

Finally, we note that resampling methods in the context of generalized linear models are not just theoretical abstractions but are used in machine learning practice. For instance, \citet{Musil2019Fast} use subsampling to estimate the uncertainty in kernel regression for the energy of molecular compounds. Their observation that subsampling yields a better uncertainty estimation than Bootstrap or Gaussian processes is one motivation for the present work.

\section{Main technical results}
\label{sec:technical}
The key observation in the results that follow is that in order to asymptotically characterize the biases and variances associated with any of the resampling methods in \cref{sec:setting}, it is sufficient to characterize only a few correlations. For example, the resampling variance \eqref{eq:def:var}:
\begin{align}
    \widehat{\Var} = \frac1d\left(\frac{1}{B}\sum\limits_{k=1}^{B}\lVert\hat{\vec{\theta}}_{k}\lVert^{2} - \frac{1}{B^2}\sum\limits_{k,k'=1}^{B}\hat{\vec{\theta}}_{k}^{\top}\hat{\vec{\theta}}_{k'}\right).
\end{align}
Assuming the data sets $\mathcal{D}_{k}^{\star}$ are independently resampled from $\mathcal{D}$, it is then enough to characterize the norm of $\hat{\vec{\theta}}_{1}$ and the correlation between two independent (conditionally on $\mathcal{D}$) resampled estimators $\hat{\vec{\theta}}_{1}^{\top}\hat{\vec{\theta}}_{2}$ - with all the rest being statistically similar. The results that follow precisely characterize these quantities asymptotically.

Finally, the methods defined in \cref{sec:setting} naturally divide into two categories: estimators for the statistics of the joint distribution $p_{\wstar}(\vec{x},y)$ (we refer to them as \emph{pair resampling}) and for the conditional distribution $p(y|\wstar^{\top}\vec{x})$ (we refer to them as \emph{conditional} or \emph{residual resampling}). Below, we start by discussing our results for the former. 

\subsection{Pair resampling}
The key idea is to reframe the regularized MLE problem \eqref{eq:def_erm} as a \textit{weighted empirical risk minimization} (wERM) problem:
\begin{equation}
    \werm\left(\dataset,\vec{p} \right) = \arg\min_{\Vec{\theta}\in\reals^d} \sum_{i=1}^{n} - p_i \log p \left( y_i | \Vec{\theta}^{\top} \Vec{x}_i \right) + \frac{\lambda}{2} \| \Vec{\theta} \|^2
    \label{eq:def_weighted_erm}
\end{equation}
where for each sample $(\vec{x}_{i},y_{i})\in\mathcal{D}$, we have introduced a sample weight $p_{i}$. When $p_{i}=1$ for all $i\in[n]$, this reduces to standard MLE \eqref{eq:def_erm}, which we sometimes refer to as full resampling. However, by taking the $p_{i}$'s at random from a judiciously chosen distribution, we can asymptotically cover all pair resampling methods from \cref{sec:setting}. 

Indeed, it is immediate to see that by choosing $p_{i}\in\{0,1\}$ at random from a Bernoulli distribution with probability $r\in(0,1]$, the wERM \eqref{eq:def_weighted_erm} asymptotically corresponds to doing subsampling. Intuitively, this can be seen as throwing a coin for each sample $i\in[n]$ in order to decide whether to include it in the subsampled batch $\mathcal{D}^{\star}_{\rm ss}$, which on average will contain precisely $r$ samples. The jackknife estimator can then be obtained as the $r\to 1^{-}$ limit of subsampling.

Similarly, pair bootstrap is asymptotically equivalent to taking $p_{i}\sim \Pois(1)$ independently. Indeed, for finite $n$, pair bootstrap exactly corresponds to taking $\vec{p}\in\mathbb{R}^{n}$ from the multinomial distribution ${\rm Multinomial}(n,\sfrac{1}{n})$. As $n\to\infty$, this is marginally equivalent to choosing $p_{i}\sim\Pois(1)$ independently~\citep{ElKaroui2018}.

To summarize, each resampling method can be thought of as applying sampling weights which are \text{i.i.d.}, with distributions defined as
\begin{align}
    \begin{cases}
        \mu_{\pb}(p) &\vcentcolon= \frac{1}{ep!}\\
        \mu_{\Ss(r)}(p) &\vcentcolon= r^p(1-r)^{1-p} \text{ for } r\in(0,1).
    \end{cases}
\end{align}
We note that a key assumption which permits to retrieve our result is that for a particular resampling method, the sample weights $p_i, \: i\in[n]$ are \textit{i.i.d.}. We are now ready to state our first two results for pair resampling. For the sake of clarity, we state our results for ridge regression and refer to~\cref{appendix:se_pair_resampling} for the derivation of our results and a statement for general convex loss and penalties.

In the following, the asymptotic values of correlations needed to compute biases and variances will be referred to as \textit{overlaps}. For $\rm{t}\!\in\!\{\pb, \Ss, \jk\}$, these overlaps read:
\begin{align}
\label{eq:def:overlaps}
    \begin{cases}
        Q_{11}^{\rm t} \!\!\!\! &\vcentcolon= \lim_{n, d\to\infty} \mathbb{E}_{\wstar, \dataset, \Vec{p} } \left[ \| \werm(\dataset, \Vec{p}) \|^2 \right] \\
        Q_{12}^{\rm t}\!\!\!\! &\vcentcolon= \lim_{n, d\to\infty}\mathbb{E}_{\wstar, \dataset } \left[ \| \mathbb{E}_{\Vec{p}} [ \werm(\dataset, \Vec{p}) ] \|^2 \right]\\
        Q_{11}^{\fr}\!\!\!\! &\vcentcolon= \lim_{n, d\to\infty} \mathbb{E}_{\wstar, \dataset} \left[ \| \werm(\dataset) \|^2 \right] \\
        Q_{12}^{\fr}\!\!\!\! &\vcentcolon= \lim_{n, d\to\infty}\mathbb{E}_{\wstar} \left[ \| \mathbb{E}_{\dataset} [ \werm(\dataset) ] \|^2 \right]\\
        Q_{12}^{\fr, \rm{t}}\!\!\!\! &\vcentcolon= \lim_{n, d\to\infty}\mathbb{E}_{\wstar, \dataset, \Vec{p}} \left[\werm(\dataset)^\top\werm(\dataset, \Vec{p}) \right]\\
        m_1^{\rm t}\!\!\!\! &\vcentcolon= \lim_{n, d\to\infty}\mathbb{E}_{\wstar, \dataset, \Vec{p}} \left[ \werm(\dataset, \Vec{p})^\top \wstar \right]\\
        m_1^{\fr}\!\!\!\! &\vcentcolon= \lim_{n, d\to\infty}\mathbb{E}_{\wstar, \dataset} \left[ \werm(\dataset)^\top \wstar \right]
    \end{cases},
\end{align}
where $\vec{p}=(p_1, \dots, p_n)\stackrel{\text{i.i.d.}}{\sim}\mu_{\rm t}$.

\begin{theorem}[Biases and Variances for pair resampling in ridge regression]\label{thm:pair_resampling}
    Let $\dataset = \{(\Vec{x}_{i}, y_{i})_{i\in[n]}\}$ denote $n$ independent samples drawn from model~\eqref{eq:def_model} with log-concave likelihood $p(y|z)$.
    In the high-dimensional proportional regime $n, d\to\infty$ with $\sfrac{n}{d}=\alpha$, the overlaps of interest \eqref{eq:def:overlaps} are given by the unique solution $\Vec{m} \in \mathbb{R}^2$, $\mat{Q} \in \mathbb{R}^{2 \times 2}, \mat{V} \in \mathbb{R}^2$ to the following set of self-consistent equations:
    \begin{eqnarray}
    &\begin{cases}
        \Vec{m} \!\!\!\!&= \left( \lambda \mat{I}_2 + \hat{\mat{V}} \right)^{-1} \hat{\vec{m}} \\
        \mat{Q}  \!\!\!\!   &= \left( \lambda \mat{I}_2 + \hat{\mat{V}} \right)^{-1} \left( \hat{\vec{m}} \hat{\vec{m}}^\top + \hat{\mat{Q}} \right) \left( \lambda \mat{I}_2 + \hat{\mat{V}} \right)^{-1\top} \\
        \mat{V}   \!\!\!\!    &= \left( \lambda \mat{I}_2 + \hat{\mat{V}} \right)^{-1} 
    \end{cases}
    \label{eq:thm_se_overlaps}
    \\
    &\begin{cases}
        \hat{\vec{m}}\!\!\!\! &= \alpha \mathbb{E}_{\Vec{p}} \left[ \mat{G}(\Vec{p}) \right] \mathbf{1}_2 \\
        \hat{\mat{Q}}   \!\!\!\!    &= \alpha \mathbb{E}_{\Vec{p}} \left[ \mat{G}(\Vec{p}) \left( (v_{\star} + \Delta) \mathbf{1}_{2 \times 2} + \mat{B Q B}^{\top} \right) \mat{G}(\Vec{p})^{\top} \right]\\
        \hat{\mat{V}}   \!\!\!\!    &= \alpha \mathbb{E}_{\Vec{p}} \left[ \mat{G}(\Vec{p}) \right]\\
    \end{cases},
    \label{eq:thm_se_hat_overlaps}
\end{eqnarray}
for a careful choice of the joint distribution of $\Vec{p} = (p_1, p_2)$. 
In the above, $\mat{G}(\Vec{p}) = (\mat{I}_2 + \mat{P V})^{-1} \mat{P}$ with $\mat{P} = \mathrm{Diag}(\Vec{p}), \;\mat{B} = \begin{pmatrix}
    \Vec{m}^\top \\ \vec{m}^\top
\end{pmatrix} \mat{Q}^{-1} - \mat{I}_2$ and $v_{\star} = 1 - \Vec{m}^\top Q^{-1} \Vec{m}$.

Then, the following holds:
\begin{itemize}
    \item the variance of resampling method $\mathrm{t}\in\{\pb, \Ss, \jk\}$ is given by
    \begin{equation}
        \widehat{\Var_t} = Q_{11}^{\mathrm{t}}-Q_{12}^{\mathrm{t}},
    \end{equation}
    where overlaps with superscript $\mathrm{t}$ are obtained by solving~\eqref{eq:thm_se_overlaps}, \eqref{eq:thm_se_hat_overlaps} using joint distribution $\mu(p_1, p_2) = \mu_{\rm t}(p_1)\cdot\mu_{\rm t}(p_2)$.
    \item the true variance is given by
    \begin{equation}
        \varianceOnXY = Q^{\fr}_{11}-Q^{\fr}_{12},
    \end{equation}
    where overlaps with superscript $\fr$ (indicating full resampling) are obtained by solving~\eqref{eq:thm_se_overlaps}, \eqref{eq:thm_se_hat_overlaps} using joint distribution $$\mu(p_1, p_2)= (\mathbbm{1}(p_1 = 0, p_2 = 1) + \mathbbm{1}(p_1 = 1 , p_2 = 0)).$$
    \item the squared bias of resampling method $\mathrm{t}$ is given by
    \begin{equation}
        \widehat{\Bias^2_{\rm t}} = Q^{\fr}_{11} + Q^{\rm t}_{12} - 2 Q^{\rm t, \rm fr}_{12},
    \end{equation}
    where overlaps with superscript $\mathrm{t}, \fr$ are obtained by solving~\eqref{eq:thm_se_overlaps}, \eqref{eq:thm_se_hat_overlaps} using distribution $\mu(p_1, p_2) = \mu_{\rm t}(p_1)\cdot\mathbbm{1}\{p_2=1\}$ for $p_1, p_2$.
    \item the true squared bias is given by
    \begin{equation}
        \biasOnXY=1 - 2 m^{\rm fr}_1 + Q^{\rm fr}_{12}.
    \end{equation}
\end{itemize}
\end{theorem}

\looseness=-1
\subsection{Conditional resampling}

Similar to pair resampling, we leverage the fact that the conditional bias and variance, together with the estimates by residual bootstrap, can be written in terms of correlations between estimators. The key difference here is that the covariate $\vec{x}_i$ remain constant, and only the labels are resampled. Focusing on linear regression, in the case of residual resampling, the labels are sampled from the true distribution $y^{\star}_i \sim \mathcal{N}(\wstar^{\top} \Vec{x}_i, \Delta)$, whereas for residual bootstrap, we use the ERM estimator to approximate this distribution and $y^{\star}_i \sim \mathcal{N}(\werm^{\top} \Vec{x}_i, \tilde{\Delta})$ with $\tilde{\Delta}$ an estimator of $\Delta$. 
Similarly to pair bootstrap, we now just need the correlation between $B$ estimators $\hatw_{\lambda, k}$ trained on resampled datasets $\dataset^{\star}_k = \{(\vec{x}_{i},y_{i,k}^{\star})_{i=1}^{n}\}$. This can be done by considering the minimization problem~\eqref{eq:def_erm_residual}.  Despite minimizing each $\hatw_{\lambda, k}$ independently, they see the same covariates $\vec{x}_{i}$. In \cref{appendix:residual_bootstrap}, we discuss how this correlation can be exactly captured by designing a particular approximate message passing, and also provide more details and an extension to more generic losses. As in the previous section, we first define the overlaps of interest 
\begingroup
\allowdisplaybreaks
\begin{align}
\label{eq:def:res:overlaps}
    \begin{cases}
        Q_{11}^{\rb} &\vcentcolon= \lim_{n, d\to\infty} \mathbb{E}_{\wstar, \dataset } \left[ \mean{\vec{y}^\star|\dataset}{\| \werm(\mat{X}, \vec{y}^\star) \|^2} \right] \\
        Q_{12}^{\rb} &\vcentcolon= \lim_{n, d\to\infty}\mathbb{E}_{\wstar, \dataset } \left[ \| \mathbb{E}_{\vec{y}^\star|\dataset} [ \werm(\mat{X}, \vec{y}^\star) ] \|^2 \right]\\
        Q_{11}^{\rr} &\vcentcolon= \lim_{n, d\to\infty} \mathbb{E}_{\wstar, \dataset} \left[ \| \werm \|^2 |\mat{X}\right] \\
        Q_{12}^{\rr} &\vcentcolon= \lim_{n, d\to\infty}\mathbb{E}_{\wstar} \left[ \| \mathbb{E}_{\dataset} [ \werm|\mat{X} ] \|^2 \right]\\
        m_{1}^{\rb} &\vcentcolon= \lim_{n, d\to\infty}\mathbb{E}_{\wstar, \dataset} \left[\werm(\dataset)^\top\mean{\vec{y}^\star|\dataset}{\werm(\mat{X}, \vec{y}^\star)} \right]\\
        m_1^{\rr} &\vcentcolon= \lim_{n, d\to\infty}\mathbb{E}_{\wstar} \left[ \mean{\dataset}{\werm|\mat{X}}^\top \wstar \right].
    \end{cases}
\end{align}
\endgroup
and the minimization problem for conditional resampling
\begin{equation}
    \hatw_{\lambda, k} = \arg\min\limits_{\Vec{\theta}}\sum_{i=1}^{n} - \log p(y^{\star}_{k,i} | \Vec{\theta}^{\top} \Vec{x}_i) + \sfrac{\lambda}{2} \| \Vec{\theta} \|^2.
    \label{eq:def_erm_residual}
\end{equation}

\begin{theorem}[Biases and Variances for conditional resampling in ridge regression]\label{thm:conditional_resampling}
    Let $\dataset = \{(\Vec{x}_{i}, y_{i})_{i\in[n]}\}$ denote $n$ independent samples drawn from model~\eqref{eq:def_model} with log-concave likelihood $p(y|z)$.
    In the high-dimensional proportional regime $n, d\to\infty$ with $\sfrac{n}{d}=\alpha$, the overlaps of interest \eqref{eq:def:res:overlaps} are given by the unique solution $\Vec{m} \in \mathbb{R}^2$, $\mat{Q} \in \mathbb{R}^{2 \times 2}, \mat{V} \in \mathbb{R}^2$ to the following set of self-consistent equations:

\begin{align}
    \begin{cases}
        \Vec{m} &= \Tilde{\rho} \left( \lambda \mat{I}_2 + \hat{\mat{V}} \right)^{-1} \hat{\vec{m}} \\
        \mat{Q}       &= \left( \lambda \mat{I}_2 + \hat{\mat{V}} \right)^{-1} \left( \Tilde{\rho}  \hat{\vec{m}} \hat{\vec{m}}^\top + \hat{\mat{Q}} \right) \left( \lambda \mat{I}_2 + \hat{\mat{V}} \right)^{-1\top} \\
        \mat{V}       &= \left( \lambda \mat{I}_2 + \hat{\mat{V}} \right)^{-1} 
    \end{cases}
    \label{eq:se_overlaps_conditional}
\end{align}
and
\begin{align}
    \begin{cases}
        \hat{\vec{m}} &= \alpha \mat{G} \mathbf{1}_2 \\
        \hat{\mat{Q}} &= \alpha \mat{G} \left( v_{\star} \mathbf{1}_{2 \times 2} + \Tilde{\Delta} \mat{I}_2 + \mat{B Q B}^{\top} \right) \mat{G}^{\top}\\
        \hat{\mat{V}} &= \alpha \mat{G}
    \end{cases}
    \label{eq:se_hat_overlaps_conditional}
\end{align}
where $\mat{G} = \left(\mat{I}_B + \mat{V}\right)^{-1}$, $\mat B$ is defined as in Theorem~\ref{thm:pair_resampling} and $v_{\star} = \Tilde{\rho} - \Vec{m}^{\top} \mat{Q}^{-1} \Vec{m}$ where $\Tilde{\rho}$ depends on the method and is defined below.
Then, the following holds:
\begin{itemize}
    \item the variance of residual bootstrap is given by 
    \begin{equation}
        \varianceResidualBootstrap = Q_{11}^{\rb} - Q_{12}^{\rb},
    \end{equation}
    where $\mat{Q}^{\rb}$ solves~\eqref{eq:thm_se_overlaps}, \eqref{eq:thm_se_hat_overlaps} using $\Tilde{\rho} = Q_{11}^{\fr}$ and $\Tilde{\Delta} = \sfrac{(1 + \Delta - 2 m_{1}^{\fr} + Q_{11}^{\fr})}{(1 + V_{11}^{\fr})^2}$. Note that the overlaps with superscript $\fr$ are specified in~\cref{thm:pair_resampling}.
    \item the true variance $\varianceOnY$ is given by
    \begin{equation}
        \varianceOnY = Q^{\rr}_{11}-Q^{\rr}_{12},
    \end{equation}
    where $\mat{Q}^{\rr}$ is obtained by solving~\eqref{eq:se_overlaps_conditional} and \eqref{eq:se_hat_overlaps_conditional} with $\Tilde{\rho} = 1, \Tilde{\Delta} = \Delta$.
    
    \item the squared bias of residual bootstrap
    \begin{equation}
        \biasResidualBootstrap = Q^{\fr}_{11} + Q^{\rb}_{12} - 2 m^{\rb}_{1}
    \end{equation}
    \item the true conditional squared bias is given by
    \begin{equation}
        \biasOnY = 1 - 2 m^{\rr}_1 + Q^{\rr}_{12}.
    \end{equation}
\end{itemize}
\end{theorem}

The details for the derivations of~\cref{thm:conditional_resampling} are shown in~\cref{appendix:residual_bootstrap}. Compared to pair resampling, residual resampling does not involve introducing sample weights, only the labels are resampled from a conditional distribution. However, for residual bootstrap, the main idea is that the target weights $\wstar$ are replaced by $\werm$. Moreover, for ridge regression, we approximate the variance $\Delta$ by the averaged residual:
\begin{equation}
    \Tilde{\Delta} = \frac{1}{n}\sum_{i = 1}^n (y_i - \werm^{\top} \vec{x}_i)^2
\end{equation}

In the high-dimensional regime, the analytical expression of this training error is given by the overlaps of state-evolution, and $\Tilde{\Delta} = \sfrac{(1 + \Delta - 2 m_{1}^{\fr} + Q_{11}^{\fr})}{(1 + V_{11}^{\fr})^2}$. The derivation of this expression can be found in \citet{loureiro2021learning}. We end this section by observing that so far, we considered only the variance on the weights. However, one could be interested in other types of variances such as \textit{predictive variance}, which we discuss in~\cref{appendix:other_variances}.

\section{Discussions and main findings}
\label{sec:discussions}
\begin{table}[t]
\parbox{.5\textwidth}{
        \centering
        \begin{tabular}{c | c | c }
            \multicolumn{3}{c}{Pair resampling rates}\\
            \toprule             
                  & Rate & Error \\
            \midrule
            $\varianceOnXY$ & $\sfrac{1}{\alpha}$ &  --\\
            $\varianceSubsampling$ & $\sfrac{1}{\alpha}$ & $\sfrac{1}{\alpha}$\\
            $\varianceJackknife$ & $\sfrac{1}{\alpha}$ & $\sfrac{1}{\alpha^2}$\\
            $\variancePairBootstrap$ & $\sfrac{1}{\alpha}$ & $\sfrac{1}{\alpha^3}$\\
            \midrule
            $\biasOnXY$ & $\sfrac{1}{\alpha^2}$ &  --\\
            $\biasSubsampling$ & $\sfrac{1}{\alpha^2}$ & $\sfrac{1}{\alpha^2}$\\
            $\biasJackknife$ & $\sfrac{1}{\alpha^2}$ & $\sfrac{1}{\alpha^3}$\\
            $\biasPairBootstrap$ & $\sfrac{1}{\alpha^4}$ & $\sfrac{1}{\alpha^2}$\\
            \bottomrule
        \end{tabular}
        }
 \parbox{.5\textwidth}{
        \centering
        \begin{tabular}{c | c | c}
            \multicolumn{3}{c}{Residual resampling rates}\\
            \toprule             
              & Rate & Error \\
            \midrule
            $\varianceOnY$ & $\sfrac{1}{\alpha}$ & -- \\
            $\varianceResidualBootstrap$ & $\sfrac{1}{\alpha}$ &  $\sfrac{1}{\alpha^2}$\\
            \midrule
            $\biasOnY$ & $\sfrac{1}{\alpha^2}$ & -- \\
            $\biasResidualBootstrap$ & $\sfrac{1}{\alpha^2}$ & $\sfrac{1}{\alpha^3}$ \\
            \bottomrule
        \end{tabular}
        }
    \caption{Summary of large $\alpha$ rates for ridge regression (see~\cref{appendix:large_alpha_rates} for details).}
    \label{table:large_alpha_rates}
\end{table}

In this section we discuss the consequences of the technical results from~\cref{sec:technical} on the performance of resampling methods, and compare with empirical values. We refer to~\cref{appendix:numerics} for more details on the plots.

\subsection{Ridge regression} 
\label{sec:ridge_numerical_results}

\begin{figure*}[t]
    \centering
    \def\figwidth{0.32\columnwidth}
    \def\figheight{0.32\columnwidth}
    
    \input{Figures/ridge/sigma=1lambda=0.01/ridge_regression_lambda=0.01_variance}
\begin{tikzpicture}

\tikzstyle{every node}=[font=\tiny]
\begin{axis}[
width=\figwidth,
height=\figheight,
legend cell align={left},
legend style={fill opacity=0.5, text opacity=1, draw=none, at={(0.5,1.4)}, anchor=north},
legend columns=2, 
log basis x={10},
log basis y={10},
tick align=outside,
tick pos=left,
x grid style={white!69.0196078431373!black},
xlabel={\(\displaystyle n / d\)},
xmajorgrids,
xmin=0.0630957344480193, xmax=1584.89319246111,
xmode=log,
xtick style={color=black},
y grid style={white!69.0196078431373!black},
ymajorgrids,
ymin=8.79523097270906e-06, ymax=8.12213042084835,
ymode=log,
ytick style={color=black}
]
\addplot [draw=blue, fill=blue, forget plot, mark=*, only marks, mark size=1pt]
table{%
x  y
0.1 0.0955757199188433
0.158489319246111 0.168005556701651
0.251188643150958 0.29632631806705
0.398107170553497 0.572509939403609
0.630957344480193 1.40620517331222
1 4.3502516574617
1.58489319246111 1.49080092054981
2.51188643150958 0.571974809083435
3.98107170553497 0.316064308863492
6.30957344480193 0.170152746071062
10 0.100653867226234
15.8489319246111 0.0604086529816402
25.1188643150958 0.0402169818021407
39.8107170553497 0.0248281188474615
63.0957344480193 0.0151216436450233
100 0.00975632663841655
};
\addplot [draw=blue, fill=blue, forget plot, mark=*, only marks, mark size=1pt]
table{%
x  y
0.1 1.64211219641278e-05
0.158489319246111 4.31293109703146e-05
0.251188643150958 9.47512674393128e-05
0.398107170553497 0.000322011071281431
0.630957344480193 0.00243931094548415
1 0.347749797490979
1.58489319246111 0.465706905579717
2.51188643150958 0.299254815780222
3.98107170553497 0.245563178545702
6.30957344480193 0.143700237150156
10 0.0892631666259161
15.8489319246111 0.0531990027174608
25.1188643150958 0.0393570872432478
39.8107170553497 0.0215955690448371
63.0957344480193 0.0153829330933823
100 0.00930386750042774
};
\addplot [semithick, black]
table {%
0.1 0.951249219725039
0.120679264063933 0.941479149581631
0.145634847750124 0.929830259620891
0.175751062485479 0.915978098023596
0.212095088792019 0.899559775422429
0.255954792269954 0.880178451897206
0.308884359647748 0.857413732304305
0.372759372031494 0.830840736193387
0.449843266896945 0.800061166989237
0.542867543932386 0.764749750660854
0.655128556859551 0.724718172485006
0.79060432109077 0.679994913816788
0.954095476349994 0.630911893083236
1.15139539932645 0.578177441501334
1.38949549437314 0.522903447622853
1.67683293681101 0.466551797298133
2.02358964772516 0.410783386850318
2.44205309454865 0.35723383338175
2.94705170255181 0.307282514152105
3.55648030622313 0.26189171569943
4.29193426012878 0.22155793280544
5.17947467923121 0.18636415136042
6.25055192527397 0.156088059021772
7.54312006335462 0.130319652852411
9.10298177991522 0.108559466140178
10.9854114198756 0.0902877537860572
13.2571136559011 0.0750068229272322
15.9985871960606 0.0622632036977974
19.3069772888325 0.051656537798877
23.2995181051537 0.0428405721044067
28.1176869797423 0.0355199321742889
33.9322177189533 0.0294449659099199
40.9491506238043 0.0244059848050846
49.4171336132384 0.020227616829027
59.6362331659464 0.0167636171513071
71.9685673001152 0.0138922732833311
86.8511373751353 0.0115124279798814
104.811313415469 0.00954008641021109
126.48552168553 0.00790554910133878
152.641796717523 0.00655100441585887
184.206996932672 0.00542851546288103
222.29964825262 0.00449834164283569
268.269579527972 0.00372754192700753
323.745754281764 0.00308881412652284
390.693993705462 0.00255953115458984
471.486636345739 0.00212094134705143
568.98660290183 0.00175750519621309
686.6488450043 0.00145634538867079
828.642772854684 0.00120679088313369
1000 0.000999999000002028
};
\addlegendentry{$\Var_{\rm bo}$}
\addplot [semithick, blue]
table {%
0.1 0.108422438144413
0.120679264063933 0.133767103486768
0.145634847750124 0.165896666670245
0.175751062485479 0.207107514870522
0.212095088792019 0.260762938832645
0.255954792269954 0.331985100842307
0.308884359647748 0.428959038050959
0.372759372031494 0.565571980272228
0.449843266896945 0.767237860690658
0.542867543932386 1.08512845924867
0.655128556859551 1.63450738168487
0.79060432109077 2.6878614001999
0.954095476349994 4.3329767778829
1.15139539932645 3.70506612977992
1.38949549437314 2.18542680232547
1.67683293681101 1.37795942531813
2.02358964772516 0.940776745143141
2.44205309454865 0.677562638315907
2.94705170255181 0.505734537756847
3.55648030622313 0.386949652945393
4.29193426012878 0.301384078994228
5.17947467923121 0.237853112233166
6.25055192527397 0.189595920400761
7.54312006335462 0.152295343366242
9.10298177991522 0.123069948603246
10.9854114198756 0.0999258235312862
13.2571136559011 0.0814414965992962
15.9985871960606 0.0665782205252898
19.3069772888325 0.0545611039598888
23.2995181051537 0.0448020258384668
28.1176869797423 0.0368481180359314
33.9322177189533 0.0303464091470532
40.9491506238043 0.0250189808389455
49.4171336132384 0.0206451405285614
59.6362331659464 0.0170483868891171
71.9685673001152 0.0140867191524415
86.8511373751353 0.0116453246603405
104.811313415469 0.00963098801738993
126.48552168553 0.00796776675927735
152.641796717523 0.00659361254906854
184.206996932672 0.00545770775039256
222.29964825262 0.00451834980543342
268.269579527972 0.00374125964959049
323.745754281764 0.00309822152376993
390.693993705462 0.00256598398775176
471.486636345739 0.00212536834117427
568.98660290183 0.00176054280458959
686.6488450043 0.00145842991778289
828.642772854684 0.0012082215166066
1000 0.00100098094118273
};
\addlegendentry{$\varianceOnY$}
\addplot [semithick, blue, dashed]
table {%
0.1 2.62329624555468e-05
0.120679264063933 3.3945062028605e-05
0.145634847750124 4.46855906589549e-05
0.175751062485479 6.01467273307343e-05
0.212095088792019 8.33575062159364e-05
0.255954792269954 0.000120139558941773
0.308884359647748 0.000182725267916489
0.372759372031494 0.000299947178494742
0.449843266896945 0.000551015894257656
0.542867543932386 0.00120520653008116
0.655128556859551 0.00351182725620158
0.79060432109077 0.016659450669493
0.954095476349994 0.139649167276653
1.15139539932645 0.522962276286141
1.38949549437314 0.615748516751516
1.67683293681101 0.556707118739856
2.02358964772516 0.475998494566605
2.44205309454865 0.400147160800669
2.94705170255181 0.334142983582394
3.55648030622313 0.27815485486054
4.29193426012878 0.231165993936463
5.17947467923121 0.191932281651697
6.25055192527397 0.159263957147099
7.54312006335462 0.132105738237639
9.10298177991522 0.109550397863539
10.9854114198756 0.0908296949289868
13.2571136559011 0.0752983211750557
15.9985871960606 0.062416743957268
19.3069772888325 0.0517351417026206
23.2995181051537 0.04287916136077
28.1176869797423 0.0355376269883358
33.9322177189533 0.0294520875784865
40.9491506238043 0.0244080056109663
49.4171336132384 0.020227368482747
59.6362331659464 0.0167625144183861
71.9685673001152 0.0138909851041786
86.8511373751353 0.0115112410859497
104.811313415469 0.00953909928574692
126.48552168553 0.0079047733015789
152.641796717523 0.00655041593900196
184.206996932672 0.00542807964263181
222.29964825262 0.0044980243222843
268.269579527972 0.00372731375885138
323.745754281764 0.00308865160502991
390.693993705462 0.00255941623045586
471.486636345739 0.00212086054026406
568.98660290183 0.00175744863245075
686.6488450043 0.00145630593590318
828.642772854684 0.00120676344394877
1000 0.000999979960341735
};
\addlegendentry{$\varianceResidualBootstrap$}
\end{axis}

\end{tikzpicture}
\begin{tikzpicture}

\definecolor{color0}{rgb}{1,0.647058823529412,0}

\tikzstyle{every node}=[font=\tiny]
\begin{axis}[
width=\figwidth,
height=\figheight,
legend cell align={left},
legend style={fill opacity=0.8, text opacity=1, draw=none, at={(0.5,1.65)}, anchor=north},
log basis x={10},
log basis y={10},
tick align=outside,
tick pos=left,
x grid style={white!69.0196078431373!black},
xlabel={\(\displaystyle n / d\)},
xmajorgrids,
xmin=2, xmax=1000,
xmode=log,
xtick style={color=black},
y grid style={white!69.0196078431373!black},
ymajorgrids,
ymin=1e-9, ymax=40.236713538847,
ymode=log,
ytick style={color=black}
]
\addplot [semithick, red]
table {%
0.1 -0.148753578189284
0.120679264063933 -0.179417151362426
0.145634847750124 -0.216554172150152
0.175751062485479 -0.261724476891353
0.212095088792019 -0.317070145437651
0.255954792269954 -0.385730266597605
0.308884359647748 -0.472701879271375
0.372759372031494 -0.5867518812119
0.449843266896945 -0.744998390803021
0.542867543932386 -0.984892808943437
0.655128556859551 -1.39818743256941
0.79060432109077 -2.21533117584859
0.954095476349994 -3.45381883233227
1.15139539932645 -2.1284110711224
1.38949549437314 0.893206965314274
1.67683293681101 2.62037187212038
2.02358964772516 1.19284880664675
2.44205309454865 0.475388839925367
2.94705170255181 0.209385110552047
3.55648030622313 0.0997720057793046
4.29193426012878 0.0501580462595259
5.17947467923121 0.0261585356612382
6.25055192527397 0.0139953905303496
7.54312006335462 0.00762389735317481
9.10298177991522 0.00420647262679719
10.9854114198756 0.00234211589146782
13.2571136559011 0.00131253173675538
15.9985871960606 0.000738938447906401
19.3069772888325 0.000417369240718335
23.2995181051537 0.000236281004002636
28.1176869797423 0.000133978567830151
33.9322177189533 7.60550643486013e-05
40.9491506238043 4.32073229429397e-05
49.4171336132384 2.45593896357477e-05
59.6362331659464 1.39648106544099e-05
71.9685673001152 7.94252489544345e-06
86.8511373751353 4.51806856216486e-06
104.811313415469 2.57035746320522e-06
126.48552168553 1.46239348453037e-06
152.641796717523 8.32058532296287e-07
184.206996932672 4.73429195957387e-07
222.29964825262 2.69378452322222e-07
268.269579527972 1.5327600788595e-07
323.745754281764 8.72141642238589e-08
390.693993705462 4.96249592529807e-08
471.486636345739 2.823663047824e-08
568.98660290183 1.60666296933343e-08
686.6488450043 9.14188336142985e-09
828.642772854684 5.20170295725819e-09
1000 2.95974611486116e-09
};
\addlegendentry{$\variancePairBootstrap - \varianceOnXY$}
\addplot [semithick, green!50!black]
table {%
0.1 0.0186433567432589
0.120679264063933 0.0283401405654154
0.145634847750124 0.0431849874253311
0.175751062485479 0.0661436461870397
0.212095088792019 0.102172668134852
0.255954792269954 0.159902113824078
0.308884359647748 0.255212345922092
0.372759372031494 0.41959291874435
0.449843266896945 0.722117367589371
0.542867543932386 1.33629563242659
0.655128556859551 2.78195336423321
0.79060432109077 6.91303458714761
0.954095476349994 17.479052814987
1.15139539932645 13.9107203693544
1.38949549437314 4.94780519650281
1.67683293681101 1.97020765287763
2.02358964772516 0.919241994986872
2.44205309454865 0.477869241149897
2.94705170255181 0.267173274559983
3.55648030622313 0.157174856247191
4.29193426012878 0.0959539406295012
5.17947467923121 0.0602383857399086
6.25055192527397 0.0386476916958396
7.54312006335462 0.0252314177257252
9.10298177991522 0.0167112556354382
10.9854114198756 0.0112045989027979
13.2571136559011 0.00759376105581436
15.9985871960606 0.0051970451072693
19.3069772888325 0.00358940512820944
23.2995181051537 0.00250096182872013
28.1176869797423 0.00175775140925535
33.9322177189533 0.00124620844068521
40.9491506238043 0.000891396158997668
49.4171336132384 0.00064340896001867
59.6362331659464 0.000468742361029326
71.9685673001152 0.000344740782884366
86.8511373751353 0.000255983923475468
104.811313415469 0.000191911496422051
126.48552168553 0.000145248647953171
152.641796717523 0.000110954452396944
184.206996932672 8.55151859574469e-05
222.29964825262 6.64666076118882e-05
268.269579527972 5.20693618495348e-05
323.745754281764 4.10873648536399e-05
390.693993705462 3.26358080369161e-05
471.486636345739 2.60764442349109e-05
568.98660290183 2.09451235294574e-05
686.6488450043 1.6901415606728e-05
828.642772854684 1.36934156754742e-05
1000 1.11330303939053e-05
};
\addlegendentry{$\varianceJackknife - \varianceOnXY$}
\addplot [semithick, color0]
table {%
0.1 -0.0290024042827147
0.120679264063933 -0.0323974889640208
0.145634847750124 -0.0352727433365506
0.175751062485479 -0.0369328762795746
0.212095088792019 -0.0362426823675641
0.255954792269954 -0.0313738636639486
0.308884359647748 -0.0193987072928155
0.372759372031494 0.00437062667394583
0.449843266896945 0.0474566715862226
0.542867543932386 0.121908744532397
0.655128556859551 0.247332619318359
0.79060432109077 0.479437543405759
0.954095476349994 1.6156759003407
1.15139539932645 7.86475660015954
1.38949549437314 9.08098320446128
1.67683293681101 4.16280228633364
2.02358964772516 1.9229919157932
2.44205309454865 1.00180430214172
2.94705170255181 0.573385744320883
3.55648030622313 0.351520483888848
4.29193426012878 0.226919837200018
5.17947467923121 0.152499115327351
6.25055192527397 0.105857089821527
7.54312006335462 0.0754681702708129
9.10298177991522 0.0550219503021483
10.9854114198756 0.0408855758075759
13.2571136559011 0.0308795468379956
15.9985871960606 0.0236503186470185
19.3069772888325 0.0183321409879502
23.2995181051537 0.0143568947170851
28.1176869797423 0.0113432273079023
33.9322177189533 0.00902988785006188
40.9491506238043 0.00723455062755208
49.4171336132384 0.00582777787096967
59.6362331659464 0.00471620958926378
71.9685673001152 0.00383150621141082
86.8511373751353 0.00312295077574621
104.811313415469 0.00255242131138134
126.48552168553 0.00209092349193818
152.641796717523 0.00171616580855849
184.206996932672 0.0014108410203656
222.29964825262 0.00116139236904311
268.269579527972 0.000957116693824479
323.745754281764 0.000789504528491026
390.693993705462 0.000651748874159531
471.486636345739 0.000538375429892546
568.98660290183 0.000444961285420154
686.6488450043 0.000367918770990073
828.642772854684 0.000304327828516083
1000 0.00025180490207688
};
\addlegendentry{$\varianceSubsampling(0.8) - \varianceOnXY$}
\addplot [semithick, blue]
table {%
0.1 0.108396205181957
0.120679264063933 0.13373315842474
0.145634847750124 0.165851981079586
0.175751062485479 0.207047368143192
0.212095088792019 0.260679581326429
0.255954792269954 0.331864961283365
0.308884359647748 0.428776312783043
0.372759372031494 0.565272033093733
0.449843266896945 0.7666868447964
0.542867543932386 1.08392325271859
0.655128556859551 1.63099555442867
0.79060432109077 2.6712019495304
0.954095476349994 4.19332761060624
1.15139539932645 3.18210385349377
1.38949549437314 1.56967828557395
1.67683293681101 0.821252306578271
2.02358964772516 0.464778250576536
2.44205309454865 0.277415477515238
2.94705170255181 0.171591554174453
3.55648030622313 0.108794798084853
4.29193426012878 0.0702180850577657
5.17947467923121 0.0459208305814696
6.25055192527397 0.0303319632536625
7.54312006335462 0.0201896051286029
9.10298177991522 0.0135195507397065
10.9854114198756 0.00909612860229914
13.2571136559011 0.00614317542424048
15.9985871960606 0.00416147656802179
19.3069772888325 0.00282596225726817
23.2995181051537 0.00192286447769707
28.1176869797423 0.00131049104759551
33.9322177189533 0.000894321568566947
40.9491506238043 0.000610975227979149
49.4171336132384 0.000417772045814391
59.6362331659464 0.000285872470730975
71.9685673001152 0.000195734048263008
86.8511373751353 0.000134083574391153
104.811313415469 9.18887316431194e-05
126.48552168553 6.29934576983349e-05
152.641796717523 4.31966100664649e-05
184.206996932672 2.96281077608596e-05
222.29964825262 2.03254831491195e-05
268.269579527972 1.39458907388867e-05
323.745754281764 9.56991874012658e-06
390.693993705462 6.56775729601655e-06
471.486636345739 4.50780091043601e-06
568.98660290183 3.09417213872898e-06
686.6488450043 2.12398187993124e-06
828.642772854684 1.45807265783393e-06
1000 1.00098084099276e-06
};
\addlegendentry{$\varianceOnY - \varianceResidualBootstrap$}
\end{axis}

\end{tikzpicture}
    \input{Figures/ridge/sigma=1lambda=1/ridge_regression_lambda=1.0_variance}
\begin{tikzpicture}

\tikzstyle{every node}=[font=\tiny]
\begin{axis}[
width=\figwidth,
height=\figheight,
legend cell align={left},
legend style={fill opacity=0.8, draw opacity=1, text opacity=1, draw=white!80!black},
log basis x={10},
log basis y={10},
tick align=outside,
tick pos=left,
x grid style={white!69.0196078431373!black},
xlabel={\(\displaystyle n / d\)},
xmajorgrids,
xmin=0.0630957344480193, xmax=1584.89319246111,
xmode=log,
xtick style={color=black},
y grid style={white!69.0196078431373!black},
ymajorgrids,
ymin=0.000708226796148293, ymax=1.34045581197096,
ymode=log,
ytick style={color=black}
]
\addplot [draw=blue, fill=blue, forget plot, mark=*, only marks, mark size=1pt]
table{%
x  y
0.1 0.0247295580853182
0.158489319246111 0.0319697826113499
0.251188643150958 0.0490534850111683
0.398107170553497 0.0797630227608136
0.630957344480193 0.111310562902696
1 0.148593018474783
1.58489319246111 0.169988397795071
2.51188643150958 0.170054360375917
3.98107170553497 0.154589837730002
6.30957344480193 0.108422756126142
10 0.0802493107886243
15.8489319246111 0.0520618847899596
25.1188643150958 0.0337899063568539
39.8107170553497 0.0225123640232767
63.0957344480193 0.0146402344129415
100 0.00877606092638864
};
\addplot [draw=blue, fill=blue, forget plot, mark=*, only marks, mark size=1pt]
table{%
x  y
0.1 0.00934093317346137
0.158489319246111 0.0128021810194445
0.251188643150958 0.0216466236220817
0.398107170553497 0.034962448168303
0.630957344480193 0.0740794578372404
1 0.0954805857355204
1.58489319246111 0.116220577588606
2.51188643150958 0.139551488272207
3.98107170553497 0.14043433169254
6.30957344480193 0.102211989967323
10 0.0737118488395581
15.8489319246111 0.0526544767497575
25.1188643150958 0.0311794790145994
39.8107170553497 0.0209237678654111
63.0957344480193 0.014111595538551
100 0.0088894983934606
};
\addplot [semithick, black]
table {%
0.1 0.951249219725039
0.120679264063933 0.941479149581631
0.145634847750124 0.929830259620891
0.175751062485479 0.915978098023596
0.212095088792019 0.899559775422429
0.255954792269954 0.880178451897206
0.308884359647748 0.857413732304305
0.372759372031494 0.830840736193387
0.449843266896945 0.800061166989237
0.542867543932386 0.764749750660854
0.655128556859551 0.724718172485006
0.79060432109077 0.679994913816788
0.954095476349994 0.630911893083236
1.15139539932645 0.578177441501334
1.38949549437314 0.522903447622853
1.67683293681101 0.466551797298133
2.02358964772516 0.410783386850318
2.44205309454865 0.35723383338175
2.94705170255181 0.307282514152105
3.55648030622313 0.26189171569943
4.29193426012878 0.22155793280544
5.17947467923121 0.18636415136042
6.25055192527397 0.156088059021772
7.54312006335462 0.130319652852411
9.10298177991522 0.108559466140178
10.9854114198756 0.0902877537860572
13.2571136559011 0.0750068229272322
15.9985871960606 0.0622632036977974
19.3069772888325 0.051656537798877
23.2995181051537 0.0428405721044067
28.1176869797423 0.0355199321742889
33.9322177189533 0.0294449659099199
40.9491506238043 0.0244059848050846
49.4171336132384 0.020227616829027
59.6362331659464 0.0167636171513071
71.9685673001152 0.0138922732833311
86.8511373751353 0.0115124279798814
104.811313415469 0.00954008641021109
126.48552168553 0.00790554910133878
152.641796717523 0.00655100441585887
184.206996932672 0.00542851546288103
222.29964825262 0.00449834164283569
268.269579527972 0.00372754192700753
323.745754281764 0.00308881412652284
390.693993705462 0.00255953115458984
471.486636345739 0.00212094134705143
568.98660290183 0.00175750519621309
686.6488450043 0.00145634538867079
828.642772854684 0.00120679088313369
1000 0.000999999000002028
};
\addplot [semithick, blue]
table {%
0.1 0.0243449779108685
0.120679264063933 0.0292073035025911
0.145634847750124 0.0349922218612459
0.175751062485479 0.0418496778448723
0.212095088792019 0.0499400822757064
0.255954792269954 0.0594261037798243
0.308884359647748 0.0704577924517934
0.372759372031494 0.0831477918223682
0.449843266896945 0.0975327746599168
0.542867543932386 0.113517655995298
0.655128556859551 0.130802251465699
0.79060432109077 0.148798454777266
0.954095476349994 0.166562073429343
1.15139539932645 0.182785183357421
1.38949549437314 0.195908546885555
1.67683293681101 0.204391254837999
2.02358964772516 0.20709462886206
2.44205309454865 0.203631217794067
2.94705170255181 0.194495235786194
3.55648030622313 0.180897199731198
4.29193426012878 0.16439996336321
5.17947467923121 0.146542886001174
6.25055192527397 0.12859163680225
7.54312006335462 0.111443767807055
9.10298177991522 0.0956470923853163
10.9854114198756 0.0814717276784314
13.2571136559011 0.0689926451993672
15.9985871960606 0.058161023941504
19.3069772888325 0.0488577679362787
23.2995181051537 0.0409301378059322
28.1176869797423 0.0342150985803708
33.9322177189533 0.0285532040927183
40.9491506238043 0.0237961584791442
49.4171336132384 0.019810354811352
59.6362331659464 0.0164779676716214
71.9685673001152 0.0136966346408302
86.8511373751353 0.0113783839364185
104.811313415469 0.00944821324891443
126.48552168553 0.00784256125265304
152.641796717523 0.00650780947007112
184.206996932672 0.00539888758411899
222.29964825262 0.00447801595229991
268.269579527972 0.00371359575937791
323.745754281764 0.00307924397544757
390.693993705462 0.00255296322988652
471.486636345739 0.0021164334343059
568.98660290183 0.00175441095263063
686.6488450043 0.00145422136246065
828.642772854684 0.00120533278351875
1000 0.000998998003005847
};
\addplot [semithick, blue, dashed]
table {%
0.1 0.0124766112215532
0.120679264063933 0.0150438409337978
0.145634847750124 0.0181322796068707
0.175751062485479 0.0218424615020609
0.212095088792019 0.026290345227278
0.255954792269954 0.0316066295302236
0.308884359647748 0.0379332785293794
0.372759372031494 0.0454150872758304
0.449843266896945 0.0541830335816027
0.542867543932386 0.0643250726924815
0.655128556859551 0.075839782726494
0.79060432109077 0.0885707769431335
0.954095476349994 0.102128186161214
1.15139539932645 0.115820425860058
1.38949549437314 0.128641475722807
1.67683293681101 0.139368589104425
2.02358964772516 0.146794065476214
2.44205309454865 0.150033997669134
2.94705170255181 0.148778273770718
3.55648030622313 0.143353979965652
4.29193426012878 0.134582207644992
5.17947467923121 0.123522685537262
6.25055192527397 0.11122996712816
7.54312006335462 0.0985949129751231
9.10298177991522 0.0862805246870011
10.9854114198756 0.0747249773241222
13.2571136559011 0.0641787974996295
15.9985871960606 0.0547519896566523
19.3069772888325 0.0464579129974684
23.2995181051537 0.0392487010007073
28.1176869797423 0.0330414678822538
33.9322177189533 0.0277365036871885
40.9491506238043 0.0232292263341978
49.4171336132384 0.0194175833750947
59.6362331659464 0.0162062916037339
71.9685673001152 0.013508964415397
86.8511373751353 0.0112488819926244
104.811313415469 0.00935892826456375
126.48552168553 0.00778104779723232
152.641796717523 0.00646545425072409
184.206996932672 0.00536973787901118
222.29964825262 0.00445796251388153
268.269579527972 0.00369980458267449
323.745754281764 0.00306976205105924
390.693993705462 0.00254644552323624
471.486636345739 0.00211195410352472
568.98660290183 0.00175133297582519
686.6488450043 0.00145210659371897
828.642772854684 0.0012038799521682
1000 0.000998000003999966
};
\end{axis}

\end{tikzpicture}
\begin{tikzpicture}

\definecolor{color0}{rgb}{1,0.647058823529412,0}

\tikzstyle{every node}=[font=\tiny]
\begin{axis}[
width=\figwidth,
height=\figheight,
legend cell align={left},
legend style={fill opacity=0.8, draw opacity=1, text opacity=1, draw=white!80!black},
log basis x={10},
log basis y={10},
tick align=outside,
tick pos=left,
x grid style={white!69.0196078431373!black},
xlabel={\(\displaystyle n / d\)},
xmajorgrids,
xmin=2, xmax=1000,
xmode=log,
xtick style={color=black},
y grid style={white!69.0196078431373!black},
ymajorgrids,
ymin=3.99925595964354e-10, ymax=0.283790178464387,
ymode=log,
ytick style={color=black}
]
\addplot [semithick, red]
table {%
0.1 0.0297878981934776
0.120679264063933 0.0351963250536859
0.145634847750124 0.0414005831554186
0.175751062485479 0.0484331443050164
0.212095088792019 0.0562839857915439
0.255954792269954 0.064877261990844
0.308884359647748 0.0740419087828203
0.372759372031494 0.0834776743794654
0.449843266896945 0.0927211862916768
0.542867543932386 0.101121764074719
0.655128556859551 0.107843724689384
0.79060432109077 0.111919148949241
0.954095476349994 0.112377238000616
1.15139539932645 0.10846350832338
1.38949549437314 0.099922534057052
1.67683293681101 0.0872516558156067
2.02358964772516 0.071773560690283
2.44205309454865 0.0553957181148736
2.94705170255181 0.0400798650431075
3.55648030622313 0.0272633920771417
4.29193426012878 0.0175480325234908
5.17947467923121 0.0107816981770218
6.25055192527397 0.00638456833636503
7.54312006335462 0.00367732145742194
9.10298177991522 0.0020764392351974
10.9854114198756 0.00115684668598792
13.2571136559011 0.000639086104141207
15.9985871960606 0.000351393052260285
19.3069772888325 0.000192826541867475
23.2995181051537 0.000105810327851508
28.1176869797423 5.81378457772397e-05
33.9322177189533 3.20137805981391e-05
40.9491506238043 1.76760109630303e-05
49.4171336132384 9.78829594122654e-06
59.6362331659464 5.43660521346023e-06
71.9685673001152 3.02838038612752e-06
86.8511373751353 1.6915347709201e-06
104.811313415469 9.47202274392822e-07
126.48552168553 5.31604074649472e-07
152.641796717523 2.9895611364239e-07
184.206996932672 1.68419709400958e-07
222.29964825262 9.50263434607024e-08
268.269579527972 5.36869335654444e-08
323.745754281764 3.03657462508156e-08
390.693993705462 1.71916276769934e-08
471.486636345739 9.74100800021915e-09
568.98660290183 5.52318324409384e-09
686.6488450043 3.13347547908904e-09
828.642772854684 1.77858261452002e-09
1000 1.00994612672878e-09
};
\addplot [semithick, blue]
table {%
0.1 0.0118683666893152
0.120679264063933 0.0141634625687933
0.145634847750124 0.0168599422543752
0.175751062485479 0.0200072163428114
0.212095088792019 0.0236497370484284
0.255954792269954 0.0278194742496007
0.308884359647748 0.0325245139224139
0.372759372031494 0.0377327045465378
0.449843266896945 0.0433497410783141
0.542867543932386 0.0491925833028165
0.655128556859551 0.0549624687392047
0.79060432109077 0.0602276778341321
0.954095476349994 0.0644338872681289
1.15139539932645 0.066964757497363
1.38949549437314 0.0672670711627474
1.67683293681101 0.0650226657335741
2.02358964772516 0.0603005633858463
2.44205309454865 0.0535972201249334
2.94705170255181 0.0457169620154763
3.55648030622313 0.0375432197655463
4.29193426012878 0.0298177557182173
5.17947467923121 0.0230202004639121
6.25055192527397 0.01736166967409
7.54312006335462 0.0128488548319318
9.10298177991522 0.00936656769831523
10.9854114198756 0.00674675035430916
13.2571136559011 0.00481384769973769
15.9985871960606 0.00340903428485173
19.3069772888325 0.00239985493881034
23.2995181051537 0.00168143680522492
28.1176869797423 0.00117363069811693
33.9322177189533 0.000816700405529835
40.9491506238043 0.000566932144946408
49.4171336132384 0.000392771436257267
59.6362331659464 0.000271676067887472
71.9685673001152 0.000187670225433201
86.8511373751353 0.00012950194379413
104.811313415469 8.92849843506793e-05
126.48552168553 6.15134554207142e-05
152.641796717523 4.23552193470345e-05
184.206996932672 2.91497051078116e-05
222.29964825262 2.0053438418377e-05
268.269579527972 1.37911767034149e-05
323.745754281764 9.48192438832152e-06
390.693993705462 6.51770665027751e-06
471.486636345739 4.4793307811819e-06
568.98660290183 3.07797680543231e-06
686.6488450043 2.11476874167893e-06
828.642772854684 1.45283135055685e-06
1000 9.97999005880601e-07
};
\addplot [semithick, green!50!black]
table {%
0.1 0.000667263779093479
0.120679264063933 0.00106176336680414
0.145634847750124 0.00163671345155358
0.175751062485479 0.00246167022069498
0.212095088792019 0.00362655165232958
0.255954792269954 0.00524311164778976
0.308884359647748 0.00744260509820499
0.372759372031494 0.0103661562125872
0.449843266896945 0.0141430529277234
0.542867543932386 0.0188518221121274
0.655128556859551 0.0244614625747654
0.79060432109077 0.0307586244117463
0.954095476349994 0.0372831343281025
1.15139539932645 0.0433155499619066
1.38949549437314 0.0479696761218671
1.67683293681101 0.0504118301779695
2.02358964772516 0.0501457188635733
2.44205309454865 0.0472173596850989
2.94705170255181 0.0422056647597251
3.55648030622313 0.0359979750181655
4.29193426012878 0.0294902663681694
5.17947467923121 0.0233683849079649
6.25055192527397 0.0180325737570647
7.54312006335462 0.0136325648599274
9.10298177991522 0.01014872734733
10.9854114198756 0.00747142469079623
13.2571136559011 0.00545840607805496
15.9985871960606 0.00396864159636762
19.3069772888325 0.0028784801388318
23.2995181051537 0.00208688728693092
28.1176869797423 0.00151493036270142
33.9322177189533 0.00110277778810927
40.9491506238043 0.000806018292902368
49.4171336132384 0.000592171182388062
59.6362331659464 0.000437728727998174
71.9685673001152 0.000325800923411489
86.8511373751353 0.000244311436036971
104.811313415469 0.000184651107545611
126.48552168553 0.000140690867318154
152.641796717523 0.000108067302711062
184.206996932672 8.36702835373907e-05
222.29964825262 6.52778891030393e-05
268.269579527972 5.12974692546929e-05
323.745754281764 4.05825328538808e-05
390.693993705462 3.23034772965448e-05
471.486636345739 2.58563835056052e-05
568.98660290183 2.07986427566127e-05
686.6488450043 1.68034635361623e-05
828.642772854684 1.36276516983214e-05
1000 1.10887236928195e-05
};
\addplot [semithick, color0]
table {%
0.1 -0.0081226903301539
0.120679264063933 -0.00936706922456696
0.145634847750124 -0.0106878377260883
0.175751062485479 -0.0120311381307703
0.212095088792019 -0.0133093775381444
0.255954792269954 -0.0143902015944984
0.308884359647748 -0.0150852661128407
0.372759372031494 -0.0151417440566391
0.449843266896945 -0.0142419686547022
0.542867543932386 -0.0120197734913406
0.655128556859551 -0.00810490817218371
0.79060432109077 -0.00220659391392039
0.954095476349994 0.00576132166786766
1.15139539932645 0.0155340404720871
1.38949549437314 0.0263857512893135
1.67683293681101 0.0371451730645818
2.02358964772516 0.0464096569004696
2.44205309454865 0.0529314293904296
2.94705170255181 0.0560100725516518
3.55648030622313 0.0556764018980835
4.29193426012878 0.0525765423556052
5.17947467923121 0.0476603908205819
6.25055192527397 0.0418660906069381
7.54312006335462 0.0359278171184521
9.10298177991522 0.0303214772877759
10.9854114198756 0.0252979662497221
13.2571136559011 0.0209482231964421
15.9985871960606 0.0172657305643239
19.3069772888325 0.0141933148616467
23.2995181051537 0.0116533362888602
28.1176869797423 0.0095649912178768
33.9322177189533 0.00785305109357182
40.9491506238043 0.00645147719616191
49.4171336132384 0.00530423623510579
59.6362331659464 0.00436473997371723
71.9685673001152 0.00359471639485476
86.8511373751353 0.00296293808013737
104.811313415469 0.00244401216575174
126.48552168553 0.00201731525640037
152.641796717523 0.00166609460660228
184.206996932672 0.00137672771294084
222.29964825262 0.00113812080555109
268.269579527972 0.000941223861426589
323.745754281764 0.000778640938951193
390.693993705462 0.000644317378437177
471.486636345739 0.000533288509307562
568.98660290183 0.000441477403475777
686.6488450043 0.000365531712334111
828.642772854684 0.000302691683710777
1000 0.000250683109625727
};
\end{axis}

\end{tikzpicture}
    \caption{Variances for ridge regression at $\lambda = 10^{-2}$ (Top) and $\lambda = 1$ (Bottom). Left: variance of pair resampling methods and of Bayes-posterior. Middle: variance of conditional resampling and residual bootstrap. Right: difference between the true variances $\varianceOnXY$, $\varianceOnY$ and their estimation. Dots are simulations done at $d = 200$, with $B = 10$ resamples for bootstrap and subsampling.}
    \label{fig:variance_ridge}
\end{figure*}

\paragraph{Variance --} \cref{fig:variance_ridge} shows the different variances for ridge regression. We consider two important choices of regularization:  $\lambda = 10^{-2}$ to approximate the behavior of unpenalized estimators, and $\lambda = \sigma^2 = 1$ which is the optimal value of $\lambda$: this regularization minimizes the generalization error of $\werm$ and its test error is the same as the Bayes-optimal estimator. As explained in~\cref{sec:resampling_estimates}, the variance of Jackknife is approximated by doing subsampling with $r = 0.99$. Note that the subsampling variances with ratio $r$ are rescaled by a factor $1 - r$. We compare our theoretical predictions with numerical experiments on Gaussian data and observe an excellent agreement.
For $\lambda = 10^{-2}$ in the regime where $n > d$, our results are qualitatively consistent with \cite{ElKaroui2018}, who showed that pair (respectively residual) bootstrap overestimates (resp. underestimates) the variance. On the other hand, our results allow us to study the variances at $d > n$. In this regime, we observe that both pair and residual bootstrap suffer from under-coverage: for residual bootstrap, it is easy to understand why, as without regularization $d > n$ the ERM interpolates the training data. Thus, the residual is exactly~$0$, and the residual bootstrap thus fatally underestimates the true level of noise in the data. On the other hand, subsampling and Jackknife are closer to $\varianceOnXY$ than pair bootstrap, and as is classically known \cite{efron1981jackknife}, the Jackknife estimate provides an upper bound of the true variance. On the right panel, we see that all variances converge to $0$ with rate $\sfrac{1}{\alpha}$, and pair bootstrap converges to $\varianceOnXY$ the fastest.
On the bottom row of Fig.~\cref{fig:variance_ridge}, we observe that optimal regularization greatly mitigates the under-coverage of bootstrapping, most notably for residual bootstrap. 
We thus conclude that for small values $\sfrac{n}{d}$, bootstrap fails to accurately capture the true variances, and appropriately regularizing partially mitigates this issue. 

Note that conditioned on $\dataset$ and if the data generating process is known, the Bayes-optimal posterior variance $\varianceBO$ is the best estimation of uncertainty on the weights. As in \cref{thm:pair_resampling} and \ref{thm:conditional_resampling}, this variance can be obtained by solving a corresponding set of self-consistent equations \citep{clarte2023theoretical}. We observe that at large $\alpha$, all variances agree with $\varianceBO$. However, at optimal $\lambda$ and small $\sfrac{n}{d}$, resampling will underestimate the actual posterior variance. 

\begin{figure*}[t]
    \centering
    \def\figwidth{0.32\linewidth}
    \def\figheight{0.32\linewidth}
    
\begin{tikzpicture}

\definecolor{color0}{rgb}{1,0.647058823529412,0}

\tikzstyle{every node}=[font=\tiny]
\begin{axis}[
width=\figwidth,
height=\figheight,
legend cell align={left},
legend style={fill opacity=0.8, text opacity=1, draw=none, at={(0.5,1.4)}, anchor=north},
legend columns=2, 
log basis x={10},
log basis y={10},
tick align=outside,
tick pos=left,
x grid style={white!69.0196078431373!black},
xlabel={\(\displaystyle n / d\)},
xmajorgrids,
xmin=0.1, xmax=100,
xmode=log,
xtick style={color=black},
y grid style={white!69.0196078431373!black},
ylabel={Bias},
ymajorgrids,
ymin=6.06341352069556e-17, ymax=641.628905093759,
ymode=log,
ytick style={color=black}
]
\addplot [semithick, red]
table {%
0.1 0.811976843778531
0.120679264063933 0.775589607761439
0.145634847750124 0.732816050440726
0.175751062485479 0.682854888245363
0.212095088792019 0.624976044748717
0.255954792269954 0.558643361910893
0.308884359647748 0.483711780852927
0.372759372031494 0.40073645225173
0.449843266896945 0.31144821565219
0.542867543932386 0.219471793852964
0.655128556859551 0.131377719357722
0.79060432109077 0.0580355551189304
0.954095476349994 0.0142924181916355
1.15139539932645 0.00228501108525259
1.38949549437314 0.000558540302706856
1.67683293681101 0.000203443882083265
2.02358964772516 9.18935493152695e-05
2.44205309454865 4.69834023231197e-05
2.94705170255181 2.59738546641852e-05
3.55648030622313 1.51359067606105e-05
4.29193426012878 9.15519461774927e-06
5.17947467923121 5.6909714636344e-06
6.25055192527397 3.6109682941543e-06
7.54312006335462 2.32756360052377e-06
9.10298177991522 1.51882265098102e-06
10.9854114198756 1.00071803066371e-06
13.2571136559011 6.64442658449005e-07
15.9985871960606 4.43896598589788e-07
19.3069772888325 2.98034471546416e-07
23.2995181051537 2.00910285075295e-07
28.1176869797423 1.35882230356543e-07
33.9322177189533 9.21480882709602e-08
40.9491506238043 6.26270655335048e-08
49.4171336132384 4.26401542963362e-08
59.6362331659464 2.90748332076873e-08
71.9685673001152 1.98492371428216e-08
86.8511373751353 1.35645548127172e-08
104.811313415469 9.2773972948379e-09
126.48552168553 6.34955066303178e-09
152.641796717523 4.34814984018317e-09
184.206996932672 2.97898439249877e-09
222.29964825262 2.04173389306561e-09
268.269579527972 1.39980738111944e-09
323.745754281764 9.59957446866611e-10
390.693993705462 6.58461285496514e-10
471.486636345739 4.5173820240052e-10
568.98660290183 3.09962056022073e-10
686.6488450043 2.12707851332539e-10
828.642772854684 1.45983447552567e-10
1000 1.0019807206163e-10
};
\addlegendentry{$\biasOnXY$}
\addplot [semithick, red, dashed]
table {%
0.1 0.0307444594423554
0.120679264063933 0.0384056801638483
0.145634847750124 0.0483771317860913
0.175751062485479 0.0615903401167459
0.212095088792019 0.0795020968657063
0.255954792269954 0.104500346150639
0.308884359647748 0.140726306532124
0.372759372031494 0.195869432850021
0.449843266896945 0.285453172950898
0.542867543932386 0.444187921403182
0.655128556859551 0.759181051621025
0.79060432109077 1.4588460161747
0.954095476349994 2.62963433962682
1.15139539932645 1.73770437452218
1.38949549437314 0.376475869589829
1.67683293681101 0.0254709836939861
2.02358964772516 0.00142403531851798
2.44205309454865 0.000156550885761497
2.94705170255181 2.74596882210432e-05
3.55648030622313 6.42908987424917e-06
4.29193426012878 1.81714600255845e-06
5.17947467923121 5.8409336611831e-07
6.25055192527397 2.0561612501524e-07
7.54312006335462 7.73221748850972e-08
9.10298177991522 3.05372780395885e-08
10.9854114198756 1.25151302654558e-08
13.2571136559011 5.27682608719715e-09
15.9985871960606 2.2745112460143e-09
19.3069772888325 9.97518512235729e-10
23.2995181051537 4.43512337966467e-10
28.1176869797423 1.9936008399668e-10
33.9322177189533 9.04027963599674e-11
40.9491506238043 4.12847533937111e-11
49.4171336132384 1.89634974390174e-11
59.6362331659464 8.75077788009548e-12
71.9685673001152 4.05320221830152e-12
86.8511373751353 1.88293824976427e-12
104.811313415469 8.77076189453874e-13
126.48552168553 4.10338429901458e-13
152.641796717523 1.91402449445377e-13
184.206996932672 9.05941988094128e-14
222.29964825262 4.21884749357559e-14
268.269579527972 1.90958360235527e-14
323.745754281764 9.32587340685131e-15
390.693993705462 3.99680288865056e-15
471.486636345739 1.33226762955019e-15
568.98660290183 8.88178419700125e-16
686.6488450043 0
828.642772854684 4.44089209850063e-16
1000 8.88178419700125e-16
};
\addlegendentry{ $\biasPairBootstrap$}
\addplot [semithick, green!50!black, dashed]
table {%
0.1 0.252966236484186
0.120679264063933 0.32330952825954
0.145634847750124 0.419117359242715
0.175751062485479 0.5533645620992
0.212095088792019 0.748353920086985
0.255954792269954 1.04487948643816
0.308884359647748 1.52343302501911
0.372759372031494 2.35851460297142
0.449843266896945 3.97625510224042
0.542867543932386 7.58749251116518
0.655128556859551 17.31432654962
0.79060432109077 48.3542724017382
0.954095476349994 87.6054020706361
1.15139539932645 7.93436795158085
1.38949549437314 0.240138372564757
1.67683293681101 0.0174825158971714
2.02358964772516 0.00256356326389095
2.44205309454865 0.000589974713527396
2.94705170255181 0.000182925621317053
3.55648030622313 6.95688351370903e-05
4.29193426012878 3.05526004495959e-05
5.17947467923121 1.48844847558393e-05
6.25055192527397 7.82546027977558e-06
7.54312006335462 4.35585789659853e-06
9.10298177991522 2.53185472587347e-06
10.9854114198756 1.52151180543569e-06
13.2571136559011 9.37840916037656e-07
15.9985871960606 5.90900661734394e-07
19.3069772888325 3.77857745093024e-07
23.2995181051537 2.45736764270532e-07
28.1176869797423 1.60187418885016e-07
33.9322177189533 1.0623946167243e-07
40.9491506238043 7.07345293449178e-08
49.4171336132384 4.73976413672971e-08
59.6362331659464 3.2236435743016e-08
71.9685673001152 2.1609380951304e-08
86.8511373751353 1.4637180356658e-08
104.811313415469 9.96536186903539e-09
126.48552168553 6.35047570085588e-09
152.641796717523 4.85389506366118e-09
184.206996932672 2.27373675443232e-09
222.29964825262 2.07389660999979e-09
268.269579527972 1.31450406115618e-09
323.745754281764 4.84057238736567e-10
390.693993705462 1.03916875104914e-09
471.486636345739 7.194245199571e-10
568.98660290183 4.88498130835068e-10
686.6488450043 -6.66133814775093e-11
828.642772854684 6.17284001691586e-10
1000 1.15463194561016e-10
};
\addlegendentry{$\biasJackknife$}
\addplot [semithick, color0, dashed]
table {%
0.1 0.239018054367252
0.120679264063933 0.301485415742153
0.145634847750124 0.38438251301015
0.175751062485479 0.496866639660318
0.212095088792019 0.653848152223477
0.255954792269954 0.880910171671096
0.308884359647748 1.22479515604736
0.372759372031494 1.77768360756085
0.449843266896945 2.73925457506241
0.542867543932386 4.59479266134042
0.655128556859551 8.68790311658253
0.79060432109077 18.9389136490345
0.954095476349994 36.4588960851021
1.15139539932645 14.9121513873249
1.38949549437314 0.905211559316999
1.67683293681101 0.0528255123630084
2.02358964772516 0.00629009648047685
2.44205309454865 0.00126621726826981
2.94705170255181 0.000359647273184827
3.55648030622313 0.000128801322230032
4.29193426012878 5.41929850328416e-05
5.17947467923121 2.55814450733638e-05
6.25055192527397 1.31337343001192e-05
7.54312006335462 7.17819172990631e-06
9.10298177991522 4.11351099849355e-06
10.9854114198756 2.44465159227048e-06
13.2571136559011 1.49458042608686e-06
15.9985871960606 9.34310340205969e-07
19.3069772888325 5.94474836024262e-07
23.2995181051537 3.83624054745013e-07
28.1176869797423 2.50384546429672e-07
33.9322177189533 1.64927649315416e-07
40.9491506238043 1.09449382890148e-07
49.4171336132384 7.30748461741371e-08
59.6362331659464 4.90314344681053e-08
71.9685673001152 3.30326876962772e-08
86.8511373751353 2.23285168132748e-08
104.811313415469 1.51343382270852e-08
126.48552168553 1.02813202396135e-08
152.641796717523 6.99740265730498e-09
184.206996932672 4.76970685170386e-09
222.29964825262 3.25539595280589e-09
268.269579527972 2.22416529638281e-09
323.745754281764 1.52092782812474e-09
390.693993705462 1.04072306328362e-09
471.486636345739 7.12552239434672e-10
568.98660290183 4.88209472848667e-10
686.6488450043 3.34565708470791e-10
828.642772854684 2.29372076887557e-10
1000 1.57229784747415e-10
};
\addlegendentry{$\biasSubsampling(0.8)$}
\end{axis}

\end{tikzpicture}
\begin{tikzpicture}

\tikzstyle{every node}=[font=\tiny]
\begin{axis}[
width=\figwidth,
height=\figheight,
legend cell align={left},
legend style={fill opacity=0.8, text opacity=1, draw=none, at={(0.5,1.4)}, anchor=north},
log basis x={10},
log basis y={10},
tick align=outside,
tick pos=left,
x grid style={white!69.0196078431373!black},
xlabel={\(\displaystyle n / d\)},
xmajorgrids,
xmin=0.0630957344480193, xmax=1584.89319246111,
xmode=log,
xtick style={color=black},
y grid style={white!69.0196078431373!black},
ymajorgrids,
ymin=3.18894054542737e-11, ymax=2.83071855964696,
ymode=log,
ytick style={color=black}
]
\addplot [semithick, blue]
table {%
0.1 0.900013352897804
0.120679264063933 0.879337983086125
0.145634847750124 0.854387790491297
0.175751062485479 0.824279267167048
0.212095088792019 0.787946634877396
0.255954792269954 0.744104634749766
0.308884359647748 0.691204321061515
0.372759372031494 0.627381761119008
0.449843266896945 0.550403078345099
0.542867543932386 0.457626883365172
0.655128556859551 0.346115571436841
0.79060432109077 0.214027083562278
0.954095476349994 0.0762211343157839
1.15139539932645 0.0107511282633943
1.38949549437314 0.00177918925042087
1.67683293681101 0.000483781296923169
2.02358964772516 0.000178344863539737
2.44205309454865 7.88176003581054e-05
2.94705170255181 3.91097300462828e-05
3.55648030622313 2.09927406287225e-05
4.29193426012878 1.19144245156466e-05
5.17947467923121 7.04458673794761e-06
6.25055192527397 4.29559315140438e-06
7.54312006335462 2.68204069819866e-06
9.10298177991522 1.70574407643009e-06
10.9854114198756 1.1007156039966e-06
13.2571136559011 7.18555863032933e-07
15.9985871960606 4.73450444182788e-07
19.3069772888325 3.14295561354427e-07
23.2995181051537 2.09911473092461e-07
28.1176869797423 1.40889234501174e-07
33.9322177189533 9.49444520781384e-08
40.9491506238043 6.41939308376749e-08
49.4171336132384 4.35204663418176e-08
59.6362331659464 2.95705120390011e-08
71.9685673001152 2.01288476997519e-08
86.8511373751353 1.37225184548839e-08
104.811313415469 9.36674782181512e-09
126.48552168553 6.40014219399632e-09
152.641796717523 4.37682001752648e-09
184.206996932672 2.99524272051599e-09
222.29964825262 2.05095895822183e-09
268.269579527972 1.405044081082e-09
323.745754281764 9.62931512304976e-10
390.693993705462 6.60150600850784e-10
471.486636345739 4.52698323272216e-10
568.98660290183 3.10507397571769e-10
686.6488450043 2.13018269690224e-10
828.642772854684 1.46159750968877e-10
1000 1.00298436223056e-10
};
\addlegendentry{$\biasOnY$}
\addplot [semithick, blue, dashed]
table {%
0.1 0.18561932674549
0.120679264063933 0.221376659671811
0.145634847750124 0.263314578707232
0.175751062485479 0.312143098856221
0.212095088792019 0.368444670013626
0.255954792269954 0.432503946454224
0.308884359647748 0.504009281453433
0.372759372031494 0.581506841056509
0.449843266896945 0.661303808433467
0.542867543932386 0.734856794746348
0.655128556859551 0.780733469946821
0.79060432109077 0.731989124838402
0.954095476349994 0.394070589874126
1.15139539932645 0.0496725098663653
1.38949549437314 0.00558654585282525
1.67683293681101 0.00113684563051208
2.02358964772516 0.000342740102881045
2.44205309454865 0.000131147172163448
2.94705170255181 5.84917588968281e-05
3.55648030622313 2.89529709598391e-05
4.29193426012878 1.54332840636151e-05
5.17947467923121 8.68660245378905e-06
6.25055192527397 5.09371309709294e-06
7.54312006335462 3.08232656820451e-06
9.10298177991522 1.91146848660395e-06
10.9854114198756 1.20850450713306e-06
13.2571136559011 7.75905206307215e-07
15.9985871960606 5.04341278961817e-07
19.3069772888325 3.31100809436435e-07
23.2995181051537 2.19127798573027e-07
28.1176869797423 1.45976887910848e-07
33.9322177189533 9.77680416625049e-08
40.9491506238043 6.57678724813593e-08
49.4171336132384 4.44009806699341e-08
59.6362331659464 3.00645583983794e-08
71.9685673001152 2.04067256426299e-08
86.8511373751353 1.38791258486037e-08
104.811313415469 9.45515443717682e-09
126.48552168553 6.45011732913758e-09
152.641796717523 4.40510206090039e-09
184.206996932672 3.01126279467212e-09
222.29964825262 2.06004102665247e-09
268.269579527972 1.41019684818389e-09
323.745754281764 9.65854951573419e-10
390.693993705462 6.61810606317204e-10
471.486636345739 4.53641124664728e-10
568.98660290183 3.11043635292663e-10
686.6488450043 2.13322692843576e-10
828.642772854684 1.46333167805324e-10
1000 1.00397024027643e-10
};
\addlegendentry{ $\biasResidualBootstrap$}
\end{axis}

\end{tikzpicture}
\begin{tikzpicture}

\definecolor{color0}{rgb}{1,0.647058823529412,0}

\tikzstyle{every node}=[font=\tiny]
\begin{axis}[
width=\figwidth,
height=\figheight,
legend cell align={left},
legend style={fill opacity=0.8, text opacity=1, draw=none, at={(0.5,1.7)}, anchor=north},
log basis x={10},
log basis y={10},
tick align=outside,
tick pos=left,
x grid style={white!69.0196078431373!black},
xlabel={\(\displaystyle n / d\)},
xmajorgrids,
xmin=1, xmax=100,
xmode=log,
xtick style={color=black},
y grid style={white!69.0196078431373!black},
ymajorgrids,
ymin=1e-10, ymax=489.656890018438,
ymode=log,
ytick style={color=black},
extra y ticks = {1e-4, 1e-8},
]
\addplot [semithick, red]
table {%
3.55648030622313 8.70681688636132e-06
4.29193426012878 7.33804861519083e-06
5.17947467923121 5.10687809751609e-06
6.25055192527397 3.40535216913906e-06
7.54312006335462 2.25024142563868e-06
9.10298177991522 1.48828537294143e-06
10.9854114198756 9.8820290039825e-07
13.2571136559011 6.59165832361808e-07
15.9985871960606 4.41622087343774e-07
19.3069772888325 2.9703695303418e-07
23.2995181051537 2.00466772737329e-07
28.1176869797423 1.35682870272547e-07
33.9322177189533 9.20576854746002e-08
40.9491506238043 6.25857807801111e-08
49.4171336132384 4.26211907988971e-08
59.6362331659464 2.90660824298072e-08
71.9685673001152 1.98451839406033e-08
86.8511373751353 1.35626718744675e-08
104.811313415469 9.27652021864844e-09
126.48552168553 6.34914032460188e-09
152.641796717523 4.34795843773372e-09
184.206996932672 2.97889379829996e-09
222.29964825262 2.04169170459068e-09
268.269579527972 1.39978828528342e-09
323.745754281764 9.59948120993204e-10
390.693993705462 6.58457288693626e-10
471.486636345739 4.51736870132891e-10
568.98660290183 3.09961167843653e-10
686.6488450043 2.12707851332539e-10
828.642772854684 1.45983003463357e-10
1000 1.00197183883211e-10
};
\addlegendentry{ $\biasOnXY - \biasPairBootstrap $}
\addplot [semithick, green!50!black]
table {%
0.1 -0.559010607294345
0.120679264063933 -0.452280079501898
0.145634847750124 -0.313698691198011
0.175751062485479 -0.129490326146163
0.212095088792019 0.123377875338267
0.255954792269954 0.486236124527271
0.308884359647748 1.03972124416619
0.372759372031494 1.95777815071969
0.449843266896945 3.66480688658822
0.542867543932386 7.36802071731222
0.655128556859551 17.1829488302623
0.79060432109077 48.2962368466193
0.954095476349994 87.5911096524445
1.15139539932645 7.9320829404956
1.38949549437314 0.23957983226205
1.67683293681101 0.0172790720150881
2.02358964772516 0.00247166971457568
2.44205309454865 0.000542991311204276
2.94705170255181 0.000156951766652868
3.55648030622313 5.44329283764798e-05
4.29193426012878 2.13974058318466e-05
5.17947467923121 9.19351329220494e-06
6.25055192527397 4.21449198562128e-06
7.54312006335462 2.02829429607475e-06
9.10298177991522 1.01303207489245e-06
10.9854114198756 5.20793774771985e-07
13.2571136559011 2.73398257588651e-07
15.9985871960606 1.47004063144606e-07
19.3069772888325 7.9823273546608e-08
23.2995181051537 4.48264791952364e-08
28.1176869797423 2.43051885284726e-08
33.9322177189533 1.40913734014701e-08
40.9491506238043 8.10746381141301e-09
49.4171336132384 4.75748707096093e-09
59.6362331659464 3.16160253532865e-09
71.9685673001152 1.7601438084824e-09
86.8511373751353 1.0726255439408e-09
104.811313415469 6.8796457419749e-10
126.48552168553 9.250378241061e-13
152.641796717523 5.05745223478007e-10
184.206996932672 -7.05247638066457e-10
222.29964825262 3.21627169341772e-11
268.269579527972 -8.53033199632616e-11
323.745754281764 -4.75900208130043e-10
390.693993705462 3.8070746555263e-10
471.486636345739 2.6768631755658e-10
568.98660290183 1.78536074812995e-10
686.6488450043 -2.79321232810048e-10
828.642772854684 4.71300554139019e-10
1000 1.52651224993858e-11
};
\addlegendentry{ $\biasJackknife - \biasOnXY$}
\addplot [semithick, color0]
table {%
0.1 -0.572958789411279
0.120679264063933 -0.474104192019286
0.145634847750124 -0.348433537430577
0.175751062485479 -0.185988248585045
0.212095088792019 0.0288721074747598
0.255954792269954 0.322266809760203
0.308884359647748 0.741083375194432
0.372759372031494 1.37694715530912
0.449843266896945 2.42780635941022
0.542867543932386 4.37532086748746
0.655128556859551 8.55652539722481
0.79060432109077 18.8808780939156
0.954095476349994 36.4446036669104
1.15139539932645 14.9098663762397
1.38949549437314 0.904653019014292
1.67683293681101 0.0526220684809251
2.02358964772516 0.00619820293116158
2.44205309454865 0.00121923386594669
2.94705170255181 0.000333673418520642
3.55648030622313 0.000113665415469422
4.29193426012878 4.50377904150923e-05
5.17947467923121 1.98904736097294e-05
6.25055192527397 9.52276600596492e-06
7.54312006335462 4.85062812938253e-06
9.10298177991522 2.59468834751253e-06
10.9854114198756 1.44393356160677e-06
13.2571136559011 8.30137767637852e-07
15.9985871960606 4.90413741616181e-07
19.3069772888325 2.96440364477846e-07
23.2995181051537 1.82713769669718e-07
28.1176869797423 1.14502316073129e-07
33.9322177189533 7.2779561044456e-08
40.9491506238043 4.68223173566429e-08
49.4171336132384 3.04346918778009e-08
59.6362331659464 1.9956601260418e-08
71.9685673001152 1.31834505534556e-08
86.8511373751353 8.76396200055752e-09
104.811313415469 5.85694093224732e-09
126.48552168553 3.93176957658171e-09
152.641796717523 2.64925281712181e-09
184.206996932672 1.79072245920509e-09
222.29964825262 1.21366205974027e-09
268.269579527972 8.24357915263364e-10
323.745754281764 5.60970381258131e-10
390.693993705462 3.82261777787108e-10
471.486636345739 2.60814037034152e-10
568.98660290183 1.78247416826594e-10
686.6488450043 1.21857857138252e-10
828.642772854684 8.33886293349907e-11
1000 5.70317126857845e-11
};
\addlegendentry{$\biasSubsampling(0.8) - \biasOnXY$}
\addplot [semithick, blue]
table {%
0.1 -0.714394026152314
0.120679264063933 -0.657961323414314
0.145634847750124 -0.591073211784065
0.175751062485479 -0.512136168310828
0.212095088792019 -0.41950196486377
0.255954792269954 -0.311600688295541
0.308884359647748 -0.187195039608083
0.372759372031494 -0.0458749200624994
0.449843266896945 0.110900730088368
0.542867543932386 0.277229911381176
0.655128556859551 0.43461789850998
0.79060432109077 0.517962041276125
0.954095476349994 0.317849455558342
1.15139539932645 0.038921381602971
1.38949549437314 0.00380735660240439
1.67683293681101 0.000653064333588915
2.02358964772516 0.000164395239341308
2.44205309454865 5.23295718053429e-05
2.94705170255181 1.93820288505453e-05
3.55648030622313 7.96023033111659e-06
4.29193426012878 3.51885954796849e-06
5.17947467923121 1.64201571584144e-06
6.25055192527397 7.98119945688569e-07
7.54312006335462 4.00285870005845e-07
9.10298177991522 2.05724410173858e-07
10.9854114198756 1.07788903136452e-07
13.2571136559011 5.73493432742822e-08
15.9985871960606 3.08908347790293e-08
19.3069772888325 1.6805248082008e-08
23.2995181051537 9.2163254805655e-09
28.1176869797423 5.08765340967443e-09
33.9322177189533 2.82358958436646e-09
40.9491506238043 1.57394164368441e-09
49.4171336132384 8.80514328116533e-10
59.6362331659464 4.94046359378331e-10
71.9685673001152 2.77877942878035e-10
86.8511373751353 1.5660739371981e-10
104.811313415469 8.84066153616914e-11
126.48552168553 4.99751351412669e-11
152.641796717523 2.82820433739062e-11
184.206996932672 1.60200741561312e-11
222.29964825262 9.08206843064363e-12
268.269579527972 5.15276710189028e-12
323.745754281764 2.92343926844296e-12
390.693993705462 1.66000546641953e-12
471.486636345739 9.42801392511683e-13
568.98660290183 5.36237720893951e-13
686.6488450043 3.04423153352218e-13
828.642772854684 1.73416836446449e-13
1000 9.85878045867139e-14
};
\addlegendentry{ $| \biasResidualBootstrap - \biasOnY |$}
\end{axis}

\end{tikzpicture}
\begin{tikzpicture}

\definecolor{color0}{rgb}{1,0.647058823529412,0}

\tikzstyle{every node}=[font=\tiny]
\begin{axis}[
width=\figwidth,
height=\figheight,
legend cell align={left},
legend style={fill opacity=0.8, draw opacity=1, text opacity=1, draw=white!80!black},
log basis x={10},
log basis y={10},
tick align=outside,
tick pos=left,
x grid style={white!69.0196078431373!black},
xlabel={\(\displaystyle n / d\)},
xmajorgrids,
xmin=0.1, xmax=100,
xmode=log,
xtick style={color=black},
y grid style={white!69.0196078431373!black},
ylabel={Bias},
ymajorgrids,
ymin=2.52529628869592e-13, ymax=3.58435740245558,
ymode=log,
ytick style={color=black}
]
\addplot [semithick, red]
table {%
0.1 0.904875078027496
0.120679264063933 0.886382989096951
0.145634847750124 0.864584311706653
0.175751062485479 0.839015876058925
0.212095088792019 0.809207789558051
0.255954792269954 0.774714107184161
0.308884359647748 0.735158308343999
0.372759372031494 0.690296328918369
0.449843266896945 0.64009787092418
0.542867543932386 0.584842181135838
0.655128556859551 0.525216429530007
0.79060432109077 0.462393082816701
0.954095476349994 0.398049816833873
1.15139539932645 0.334289153861028
1.38949549437314 0.273428015535866
1.67683293681101 0.217670579562118
2.02358964772516 0.168742990912218
2.44205309454865 0.12761601171262
2.94705170255181 0.0944225435036385
3.55648030622313 0.068587270751991
4.29193426012878 0.04908791758902
5.17947467923121 0.0347315969122894
6.25055192527397 0.0243634821691843
7.54312006335462 0.0169832119195725
9.10298177991522 0.0117851576886403
10.9854114198756 0.00815187848373156
13.2571136559011 0.00562602348563712
15.9985871960606 0.00387670653471339
19.3069772888325 0.00266839789736673
23.2995181051537 0.00183531461823305
28.1176869797423 0.00126166558166618
33.9322177189533 0.000867006017436722
40.9491506238043 0.00059565209430601
49.4171336132384 0.000409156482582107
59.6362331659464 0.000281018859995763
71.9685673001152 0.000192995256978801
86.8511373751353 0.000132535997992012
104.811313415469 9.1013248714189e-05
126.48552168553 6.24977065935006e-05
152.641796717523 4.29156588566126e-05
184.206996932672 2.94687801307703e-05
222.29964825262 2.02350775355686e-05
268.269579527972 1.38945688177294e-05
323.745754281764 9.54077270831633e-06
390.693993705462 6.55119973158591e-06
471.486636345739 4.4983921978492e-06
568.98660290183 3.08882451482972e-06
686.6488450043 2.12094189122602e-06
828.642772854684 1.45634423565077e-06
1000 9.99998000184021e-07
};
\addplot [semithick, red, dashed]
table {%
0.1 0.00325187891261912
0.120679264063933 0.00386489934714088
0.145634847750124 0.00457780442488614
0.175751062485479 0.0053992498928416
0.212095088792019 0.00633449934400884
0.255954792269954 0.00738267516426652
0.308884359647748 0.00853283990575324
0.372759372031494 0.00975872403672684
0.449843266896945 0.011012200431021
0.542867543932386 0.0122162950452238
0.655128556859551 0.0132598713076866
0.79060432109077 0.0139982689606545
0.954095476349994 0.0142666035596756
1.15139539932645 0.0139129934928983
1.38949549437314 0.0128532306148287
1.67683293681101 0.0111321065657732
2.02358964772516 0.00895474309351418
2.44205309454865 0.00664720008617214
2.94705170255181 0.00454518628391387
3.55648030622313 0.00287332900543036
4.29193426012878 0.00169370355808374
5.17947467923121 0.000941686832444777
6.25055192527397 0.000499952951259264
7.54312006335462 0.000256343203223874
9.10298177991522 0.000128138727021332
10.9854114198756 6.29053584462191e-05
13.2571136559011 3.04931007666376e-05
15.9985871960606 1.46525901578709e-05
19.3069772888325 6.99866469511257e-06
23.2995181051537 3.32911219214083e-06
28.1176869797423 1.57914167742135e-06
33.9322177189533 7.47620421792661e-07
40.9491506238043 3.53486964588257e-07
49.4171336132384 1.66985336003123e-07
59.6362331659464 7.88349783231723e-08
71.9685673001152 3.720309527111e-08
86.8511373751353 1.75515666445136e-08
104.811313415469 8.27881829756905e-09
126.48552168553 3.90448051668102e-09
152.641796717523 1.84127468827455e-09
184.206996932672 8.68253469121782e-10
222.29964825262 4.09408285051427e-10
268.269579527972 1.93043359075773e-10
323.745754281764 9.10214126292885e-11
390.693993705462 4.29165591953051e-11
471.486636345739 2.0235146891423e-11
568.98660290183 9.5403684952089e-12
686.6488450043 4.49795756196636e-12
828.642772854684 2.12119211084882e-12
1000 1.00031094518727e-12
};
\addplot [semithick, green!50!black, dashed]
table {%
0.1 0.047416725702104
0.120679264063933 0.0565552956441339
0.145634847750124 0.0672691028391624
0.175751062485479 0.0797344483588479
0.212095088792019 0.0940935102741734
0.255954792269954 0.110415593571888
0.308884359647748 0.128638763630118
0.372759372031494 0.148487062493485
0.449843266896945 0.169362392314909
0.542867543932386 0.190221208032914
0.655128556859551 0.209469278901197
0.79060432109077 0.224945022644362
0.954095476349994 0.234101430310884
1.15139539932645 0.234495704469761
1.38949549437314 0.224581846299809
1.67683293681101 0.204543487853925
2.02358964772516 0.176655396850744
2.44205309454865 0.14476037086597
2.94705170255181 0.113018399863218
3.55648030622313 0.0846488299144353
4.29193426012878 0.0613296560114661
5.17947467923121 0.0433426543189341
6.25055192527397 0.0300995826751915
7.54312006335462 0.0206630417753928
9.10298177991522 0.0140852525420953
10.9854114198756 0.00956433518828346
13.2571136559011 0.00648333058350657
15.9985871960606 0.00439332629387223
19.3069772888325 0.0029785320521114
23.2995181051537 0.00202127214965486
28.1176869797423 0.00137326884530608
33.9322177189533 0.000934161161758594
40.9491506238043 0.000636225279126988
49.4171336132384 0.000433798228449688
59.6362331659464 0.000296079007977567
71.9685673001152 0.000202264116566652
86.8511373751353 0.000138284128592403
104.811313415469 9.46063338780332e-05
126.48552168553 6.47623976668398e-05
152.641796717523 4.43548020534478e-05
184.206996932672 3.03903036069641e-05
222.29964825262 2.08302797233273e-05
268.269579527972 1.42817224713099e-05
323.745754281764 9.79419212399078e-06
390.693993705462 6.71807498520137e-06
471.486636345739 4.60891547149344e-06
568.98660290183 3.16269455069573e-06
686.6488450043 2.17062368079723e-06
828.642772854684 1.48894452323134e-06
1000 1.02265751422692e-06
};
\addplot [semithick, color0, dashed]
table {%
0.1 0.0475042564306787
0.120679264063933 0.0566931852021065
0.145634847750124 0.0674875722657013
0.175751062485479 0.0800822377079673
0.212095088792019 0.0946490310787037
0.255954792269954 0.111304208737242
0.308884359647748 0.130058669405114
0.372759372031494 0.150746033788914
0.449843266896945 0.17292550045399
0.542867543932386 0.1957632008873
0.655128556859551 0.217911500676214
0.79060432109077 0.237434210304471
0.954095476349994 0.251864406288444
1.15139539932645 0.258508511733288
1.38949549437314 0.255071990871789
1.67683293681101 0.240515541356712
2.02358964772516 0.215780619382106
2.44205309454865 0.183871637606531
2.94705170255181 0.149044775955415
3.55648030622313 0.115464853370317
4.29193426012878 0.0860843959419045
5.17947467923121 0.0622554560794908
6.25055192527397 0.0440105523264678
7.54312006335462 0.0306180877908346
9.10298177991522 0.0210753496751648
10.9854114198756 0.014411076769838
13.2571136559011 0.00981716968151414
15.9985871960606 0.00667555708241397
19.3069772888325 0.00453676193211683
23.2995181051537 0.00308387563407453
28.1176869797423 0.00209764134426749
33.9322177189533 0.00142805566918924
40.9491506238043 0.000973136893611671
49.4171336132384 0.000663767489583745
59.6362331659464 0.000453156627067131
71.9685673001152 0.000309625769590039
86.8511373751353 0.00021171121660224
104.811313415469 0.000144853745315254
126.48552168553 9.91648039572191e-05
152.641796717523 6.79193058961137e-05
184.206996932672 4.65378376279802e-05
222.29964825262 3.18984458680394e-05
268.269579527972 2.18705560217636e-05
323.745754281764 1.49988158337777e-05
390.693993705462 1.02883037922386e-05
471.486636345739 7.05839003356524e-06
568.98660290183 4.84317603799411e-06
686.6488450043 3.32358811583333e-06
828.642772854684 2.28101301180672e-06
1000 1.56561384412335e-06
};
\end{axis}

\end{tikzpicture}
\begin{tikzpicture}

\tikzstyle{every node}=[font=\tiny]
\begin{axis}[
width=\figwidth,
height=\figheight,
legend cell align={left},
legend style={fill opacity=0.8, draw opacity=1, text opacity=1, draw=white!80!black},
log basis x={10},
log basis y={10},
tick align=outside,
tick pos=left,
x grid style={white!69.0196078431373!black},
xlabel={\(\displaystyle n / d\)},
xmajorgrids,
xmin=0.0630957344480193, xmax=1584.89319246111,
xmode=log,
xtick style={color=black},
y grid style={white!69.0196078431373!black},
ymajorgrids,
ymin=5.03090866154614e-07, ymax=1.84241177243699,
ymode=log,
ytick style={color=black}
]
\addplot [semithick, blue]
table {%
0.1 0.926904241814171
0.120679264063933 0.91227184607904
0.145634847750124 0.894838037759645
0.175751062485479 0.874128420178724
0.212095088792019 0.849619693146723
0.255954792269954 0.820752348117381
0.308884359647748 0.786955939852512
0.372759372031494 0.747692944371019
0.449843266896945 0.70252839232932
0.542867543932386 0.651232094665556
0.655128556859551 0.593915921019307
0.79060432109077 0.531196459039523
0.954095476349994 0.464349819653893
1.15139539932645 0.395392258143913
1.38949549437314 0.326994900737298
1.67683293681101 0.262160542460134
2.02358964772516 0.203688757988258
2.44205309454865 0.153602615587683
2.94705170255181 0.112787278365911
3.55648030622313 0.0809945159682317
4.29193426012878 0.0571579694422311
5.17947467923121 0.0398212653592458
6.25055192527397 0.0274964222195222
7.54312006335462 0.0188758850453556
9.10298177991522 0.0129123737548611
10.9854114198756 0.008816026107626
13.2571136559011 0.00601417772786506
15.9985871960606 0.00410217975629323
19.3069772888325 0.00279876986259775
23.2995181051537 0.00191043429847459
28.1176869797423 0.00130483359391809
33.9322177189533 0.000891761817201964
40.9491506238043 0.000609826325940199
49.4171336132384 0.000417262017675002
59.6362331659464 0.000285649479685901
71.9685673001152 0.000195638642501272
86.8511373751353 0.000134044043462511
104.811313415469 9.18731612964407e-05
126.48552168553 6.29878486857471e-05
152.641796717523 4.31949457879721e-05
184.206996932672 2.96278787617066e-05
222.29964825262 2.03256905357829e-05
268.269579527972 1.39461676296193e-05
323.745754281764 9.57015107516668e-06
390.693993705462 6.56792470365986e-06
471.486636345739 4.5079127453107e-06
568.98660290183 3.09424358246879e-06
686.6488450043 2.12402621047048e-06
828.642772854684 1.45809961504817e-06
1000 1.00099699618106e-06
};
\addplot [semithick, blue, dashed]
table {%
0.1 0.0451873050286117
0.120679264063933 0.0533869242452807
0.145634847750124 0.0627905527909459
0.175751062485479 0.0734459324350456
0.212095088792019 0.0853359927851837
0.255954792269954 0.0983438169604283
0.308884359647748 0.112209110304527
0.372759372031494 0.1264791880232
0.449843266896945 0.140462706919252
0.542867543932386 0.153202512647726
0.655128556859551 0.163494260128446
0.79060432109077 0.169985568655159
0.954095476349994 0.171385995883196
1.15139539932645 0.16678537394083
1.38949549437314 0.156008139786672
1.67683293681101 0.139849070194705
2.02358964772516 0.120016800118507
2.44205309454865 0.0987305644038315
2.94705170255181 0.0781297199052603
3.55648030622313 0.0597827232190666
4.29193426012878 0.0444941678892541
5.17947467923121 0.0324000090344723
6.25055192527397 0.0232045590452337
7.54312006335462 0.016415986258963
9.10298177991522 0.0115106133534311
10.9854114198756 0.00802004691304914
13.2571136559011 0.00556307336397821
15.9985871960606 0.00384676490252178
19.3069772888325 0.0026541951014003
23.2995181051537 0.00182859020016002
28.1176869797423 0.00125848599316347
33.9322177189533 0.000865503920894684
40.9491506238043 0.000594942913895835
49.4171336132384 0.000408821801464621
59.6362331659464 0.000280860961168861
71.9685673001152 0.000192920777014827
86.8511373751353 0.000132500871065977
104.811313415469 9.09966833990694e-05
126.48552168553 6.24898951553376e-05
152.641796717523 4.29119755069873e-05
184.206996932672 2.94670433638178e-05
222.29964825262 2.02342586352877e-05
268.269579527972 1.38941827050321e-05
323.745754281764 9.54059065727542e-06
390.693993705462 6.55111389558094e-06
471.486636345739 4.49835172666724e-06
568.98660290183 3.0888054334266e-06
686.6488450043 2.12093289442272e-06
828.642772854684 1.4563399934886e-06
1000 9.99996000228265e-07
};
\end{axis}

\end{tikzpicture}
\begin{tikzpicture}

\definecolor{color0}{rgb}{1,0.647058823529412,0}

\tikzstyle{every node}=[font=\tiny]
\begin{axis}[
width=\figwidth,
height=\figheight,
legend cell align={left},
legend style={fill opacity=0.8, draw opacity=1, text opacity=1, draw=white!80!black},
log basis x={10},
log basis y={10},
tick align=outside,
tick pos=left,
x grid style={white!69.0196078431373!black},
xlabel={\(\displaystyle n / d\)},
xmajorgrids,
xmin=1, xmax=100,
xmode=log,
xtick style={color=black},
y grid style={white!69.0196078431373!black},
ymajorgrids,
ymin=1e-7, ymax=2.52786988185562,
ymode=log,
ytick style={color=black}
]
\addplot [semithick, red]
table {%
0.1 0.901623199114877
0.120679264063933 0.882518089749811
0.145634847750124 0.860006507281767
0.175751062485479 0.833616626166083
0.212095088792019 0.802873290214042
0.255954792269954 0.767331432019895
0.308884359647748 0.726625468438246
0.372759372031494 0.680537604881642
0.449843266896945 0.629085670493159
0.542867543932386 0.572625886090614
0.655128556859551 0.51195655822232
0.79060432109077 0.448394813856047
0.954095476349994 0.383783213274198
1.15139539932645 0.32037616036813
1.38949549437314 0.260574784921037
1.67683293681101 0.206538472996345
2.02358964772516 0.159788247818704
2.44205309454865 0.120968811626448
2.94705170255181 0.0898773572197247
3.55648030622313 0.0657139417465606
4.29193426012878 0.0473942140309362
5.17947467923121 0.0337899100798447
6.25055192527397 0.023863529217925
7.54312006335462 0.0167268687163487
9.10298177991522 0.011657018961619
10.9854114198756 0.00808897312528534
13.2571136559011 0.00559553038487048
15.9985871960606 0.00386205394455552
19.3069772888325 0.00266139923267161
23.2995181051537 0.00183198550604091
28.1176869797423 0.00126008643998876
33.9322177189533 0.00086625839701493
40.9491506238043 0.000595298607341421
49.4171336132384 0.000408989497246104
59.6362331659464 0.00028094002501744
71.9685673001152 0.00019295805388353
86.8511373751353 0.000132518446425367
104.811313415469 9.10049698958915e-05
126.48552168553 6.24938021129839e-05
152.641796717523 4.29138175819244e-05
184.206996932672 2.94679118773011e-05
222.29964825262 2.02346681272836e-05
268.269579527972 1.38943757743704e-05
323.745754281764 9.5406816869037e-06
390.693993705462 6.55115681502672e-06
471.486636345739 4.49837196270231e-06
568.98660290183 3.08881497446123e-06
686.6488450043 2.12093739326846e-06
828.642772854684 1.45634211445866e-06
1000 9.99996999873076e-07
};
\addplot [semithick, blue]
table {%
0.1 0.881716936785559
0.120679264063933 0.858884921833759
0.145634847750124 0.832047484968699
0.175751062485479 0.800682487743678
0.212095088792019 0.764283700361539
0.255954792269954 0.722408531156953
0.308884359647748 0.674746829547985
0.372759372031494 0.621213756347818
0.449843266896945 0.562065685410069
0.542867543932386 0.498029582017829
0.655128556859551 0.430421660890862
0.79060432109077 0.361210890384363
0.954095476349994 0.292963823770697
1.15139539932645 0.228606884203083
1.38949549437314 0.170986760950626
1.67683293681101 0.122311472265429
2.02358964772516 0.0836719578697515
2.44205309454865 0.0548720511838511
2.94705170255181 0.0346575584606507
3.55648030622313 0.0212117927491651
4.29193426012878 0.012663801552977
5.17947467923121 0.00742125632477353
6.25055192527397 0.00429186317428853
7.54312006335462 0.00245989878639263
9.10298177991522 0.00140176040143003
10.9854114198756 0.000795979194576857
13.2571136559011 0.000451104363886845
15.9985871960606 0.000255414853771452
19.3069772888325 0.000144574761197447
23.2995181051537 8.18440983145763e-05
28.1176869797423 4.63476007546149e-05
33.9322177189533 2.62578963072801e-05
40.9491506238043 1.48834120443642e-05
49.4171336132384 8.44021621038138e-06
59.6362331659464 4.78851851704043e-06
71.9685673001152 2.7178654864457e-06
86.8511373751353 1.54317239653423e-06
104.811313415469 8.76477897371331e-07
126.48552168553 4.9795353040949e-07
152.641796717523 2.82970280984784e-07
184.206996932672 1.60835397888803e-07
222.29964825262 9.14319004952091e-08
268.269579527972 5.1984924587245e-08
323.745754281764 2.95604178912612e-08
390.693993705462 1.68108080789153e-08
471.486636345739 9.56101864346692e-09
568.98660290183 5.43814904219175e-09
686.6488450043 3.09331604775309e-09
828.642772854684 1.75962155957166e-09
1000 1.00099595279346e-09
};
\addplot [semithick, green!50!black]
table {%
0.1 -0.857458352325392
0.120679264063933 -0.829827693452818
0.145634847750124 -0.797315208867491
0.175751062485479 -0.759281427700077
0.212095088792019 -0.715114279283878
0.255954792269954 -0.664298513612273
0.308884359647748 -0.606519544713881
0.372759372031494 -0.541809266424884
0.449843266896945 -0.470735478609271
0.542867543932386 -0.394620973102924
0.655128556859551 -0.31574715062881
0.79060432109077 -0.237448060172339
0.954095476349994 -0.163948386522989
1.15139539932645 -0.0997934493912671
1.38949549437314 -0.0488461692360566
1.67683293681101 -0.0131270917081931
2.02358964772516 0.00791240593852593
2.44205309454865 0.0171443591533504
2.94705170255181 0.0185958563595796
3.55648030622313 0.0160615591624443
4.29193426012878 0.0122417384224461
5.17947467923121 0.00861105740664462
6.25055192527397 0.0057361005060072
7.54312006335462 0.00367982985582023
9.10298177991522 0.00230009485345503
10.9854114198756 0.00141245670455191
13.2571136559011 0.000857307097869452
15.9985871960606 0.000516619759158835
19.3069772888325 0.000310134154744674
23.2995181051537 0.000185957531421806
28.1176869797423 0.000111603263639901
33.9322177189533 6.71551443218715e-05
40.9491506238043 4.05731848209789e-05
49.4171336132384 2.46417458675811e-05
59.6362331659464 1.50601479818041e-05
71.9685673001152 9.26885958785168e-06
86.8511373751353 5.74813060039123e-06
104.811313415469 3.5930851638442e-06
126.48552168553 2.26469107333914e-06
152.641796717523 1.43914319683517e-06
184.206996932672 9.21523476193843e-07
222.29964825262 5.95202187758622e-07
268.269579527972 3.87153653580416e-07
323.745754281764 2.53419415674455e-07
390.693993705462 1.66875253615461e-07
471.486636345739 1.10523273644239e-07
568.98660290183 7.38700358660023e-08
686.6488450043 4.9681789571206e-08
828.642772854684 3.26002875805615e-08
1000 2.26595140428979e-08
};
\addplot [semithick, color0]
table {%
0.1 -0.857370821596817
0.120679264063933 -0.829689803894845
0.145634847750124 -0.797096739440952
0.175751062485479 -0.758933638350957
0.212095088792019 -0.714558758479347
0.255954792269954 -0.663409898446919
0.308884359647748 -0.605099638938885
0.372759372031494 -0.539550295129455
0.449843266896945 -0.46717237047019
0.542867543932386 -0.389078980248538
0.655128556859551 -0.307304928853793
0.79060432109077 -0.22495887251223
0.954095476349994 -0.14618541054543
1.15139539932645 -0.07578064212774
1.38949549437314 -0.018356024664077
1.67683293681101 0.0228449617945937
2.02358964772516 0.0470376284698873
2.44205309454865 0.0562556258939109
2.94705170255181 0.0546222324517763
3.55648030622313 0.0468775826183264
4.29193426012878 0.0369964783528845
5.17947467923121 0.0275238591672014
6.25055192527397 0.0196470701572835
7.54312006335462 0.0136348758712621
9.10298177991522 0.00929019198652449
10.9854114198756 0.00625919828610644
13.2571136559011 0.00419114619587702
15.9985871960606 0.00279885054770057
19.3069772888325 0.00186836403475011
23.2995181051537 0.00124856101584148
28.1176869797423 0.000835975762601305
33.9322177189533 0.000561049651752522
40.9491506238043 0.000377484799305661
49.4171336132384 0.000254611007001638
59.6362331659464 0.000172137767071368
71.9685673001152 0.000116630512611238
86.8511373751353 7.9175218610228e-05
104.811313415469 5.38404966010653e-05
126.48552168553 3.66670973637185e-05
152.641796717523 2.5003647039501e-05
184.206996932672 1.70690574972099e-05
222.29964825262 1.16633683324707e-05
268.269579527972 7.97598720403415e-06
323.745754281764 5.45804312546139e-06
390.693993705462 3.73710406065265e-06
471.486636345739 2.55999783571604e-06
568.98660290183 1.75435152316439e-06
686.6488450043 1.20264622460731e-06
828.642772854684 8.24668776155947e-07
1000 5.65615843939327e-07
};
\end{axis}

\end{tikzpicture}
    
    \caption{Bias of ridge regression and its estimation using pair bootstrap and subsampling at $\lambda = 10^{-2}$ (Top) and $\lambda = 1$ (Bottom). Left: bias of pair resampling methods. Middle: conditional bias and bias of residual bootstrap. Right: difference between the various biases.}
    \label{fig:bias_ridge}
\end{figure*}

\paragraph{Bias --} In~\cref{fig:bias_ridge}, we plot the bias of the different resampling methods for ridge regression with regularization $\lambda \in \{ 10^{-2}, 1 \}$. For the Jackknife and subsampling, the estimation of the squared bias is rescaled by a factor $(1 - r)^2$. We observe that as $\alpha \to \infty$, $\biasOnXY$ and $\biasPairBootstrap$ converge to zero, as expected by the consistency of the MLE estimator \eqref{eq:consistency}. However, $\biasPairBootstrap$ converges as $\sfrac{1}{\alpha^4}$, while $\biasOnXY \sim \sfrac{1}{\alpha^2}$, and pair bootstrap underestimates the true bias. We deduce that in our model, subsampling or Jackknife should thus be preferred to estimate $\biasOnXY$.

\begin{figure}[t!]
    \centering
    \def\figwidth{0.49\columnwidth}
    \def\figheight{0.49\columnwidth}
    
\begin{tikzpicture}

\definecolor{color0}{rgb}{1,0.647058823529412,0}

\tikzstyle{every node}=[font=\tiny]
\begin{axis}[
width=\figwidth,
height=\figheight,
legend cell align={left},
legend style={fill opacity=0.8, draw opacity=1, text opacity=1, draw=white!80!black},
log basis x={10},
log basis y={10},
tick align=outside,
tick pos=left,
x grid style={white!69.0196078431373!black},
xmajorgrids,
xmin=0.0630957344480193, xmax=1584.89319246111,
xmode=log,
xtick style={color=black},
y grid style={white!69.0196078431373!black},
ylabel={Variance},
ymajorgrids,
ymin=0.00305391681162586, ymax=70.464833100408,
ymode=log,
ytick style={color=black}
]
\addplot [semithick, red]
table {%
0.1 1.1411799974142
0.120679264063933 1.38071010639783
0.145634847750124 1.67169466869731
0.175751062485479 2.02458026093157
0.212095088792019 2.45521454053853
0.255954792269954 2.98240434320841
0.308884359647748 3.62245905798376
0.372759372031494 4.40888833105308
0.449843266896945 5.39326788597018
0.542867543932386 6.61575987872015
0.655128556859551 8.15616034929923
0.79060432109077 10.0958319981829
0.954095476349994 12.5214415331626
1.15139539932645 15.5213868009453
1.38949549437314 18.809131048677
1.67683293681101 20.906475127785
2.02358964772516 17.8210044121495
2.44205309454865 11.2204155477898
2.94705170255181 6.51170000460974
3.55648030622313 4.00277425494676
4.29193426012878 2.6325022389731
5.17947467923121 1.82859408531125
6.25055192527397 1.32059009879714
7.54312006335462 0.998331579952156
9.10298177991522 0.761254225912081
10.9854114198756 0.590917484500754
13.2571136559011 0.464064859077931
15.9985871960606 0.368507890887867
19.3069772888325 0.294649244386373
23.2995181051537 0.237386604780994
28.1176869797423 0.192162404684661
33.9322177189533 0.156268007801502
40.9491506238043 0.127499258276423
49.4171336132384 0.10431889290948
59.6362331659464 0.0855506836188558
71.9685673001152 0.0702150433580291
86.8511373751353 0.057706460623133
104.811313415469 0.0475774629027041
126.48552168553 0.0392389389348002
152.641796717523 0.032390068074543
184.206996932672 0.0267600013726845
222.29964825262 0.0221194056320817
268.269579527972 0.018292935455692
323.745754281764 0.0151349537795114
390.693993705462 0.0125267394325874
471.486636345739 0.0103712440990176
568.98660290183 0.00858896975421097
686.6488450043 0.00711465122637933
828.642772854684 0.00589461549505277
1000 0.0048846976970276
};
\addplot [semithick, red, dashed]
table {%
0.1 0.327336617156071
0.120679264063933 0.397170507437325
0.145634847750124 0.483340332914615
0.175751062485479 0.589521990824059
0.212095088792019 0.720780504111012
0.255954792269954 0.88161206010175
0.308884359647748 1.08358985019618
0.372759372031494 1.33757630565884
0.449843266896945 1.65434041943184
0.542867543932386 2.05919641899901
0.655128556859551 2.57765171551764
0.79060432109077 3.25591030434592
0.954095476349994 4.14855622425983
1.15139539932645 5.34564864880342
1.38949549437314 6.9618278160293
1.67683293681101 9.16167940233001
2.02358964772516 12.0418716151737
2.44205309454865 15.2825880226993
2.94705170255181 16.1612202782455
3.55648030622313 11.4197696806998
4.29193426012878 6.19784004139283
5.17947467923121 3.624320450113
6.25055192527397 2.22703119184664
7.54312006335462 1.47130397785656
9.10298177991522 1.02719095193537
10.9854114198756 0.744210447216872
13.2571136559011 0.559588674338466
15.9985871960606 0.428674465591868
19.3069772888325 0.333322638910958
23.2995181051537 0.262156230961528
28.1176869797423 0.20846728962642
33.9322177189533 0.167056279338968
40.9491506238043 0.134563141099679
49.4171336132384 0.109043627167845
59.6362331659464 0.0887023893759487
71.9685673001152 0.0723339273368555
86.8511373751353 0.0592102370304273
104.811313415469 0.0485738226111803
126.48552168553 0.0399257541351206
152.641796717523 0.032860887324955
184.206996932672 0.0270794514462807
222.29964825262 0.0223409068936486
268.269579527972 0.0184272876099112
323.745754281764 0.0152255755705146
390.693993705462 0.0125870943692985
471.486636345739 0.010410602344836
568.98660290183 0.00862414647338294
686.6488450043 0.00714072454775083
828.642772854684 0.00591451379534069
1000 0.00489953957672484
};
\addplot [semithick, color0, dashed]
table {%
0.1 0.944838010125148
0.120679264063933 1.1511118411493
0.145634847750124 1.40520762009983
0.175751062485479 1.71953702675849
0.212095088792019 2.11033177365825
0.255954792269954 2.59906210889653
0.308884359647748 3.21452889980048
0.372759372031494 3.99590725339023
0.449843266896945 4.99724281569915
0.542867543932386 6.29412529391632
0.655128556859551 7.99353169459525
0.79060432109077 10.2480089941896
0.954095476349994 13.2717448045843
1.15139539932645 17.3492881049239
1.38949549437314 22.7712778209658
1.67683293681101 29.4297802350969
2.02358964772516 35.0666341113252
2.44205309454865 32.1147967467352
2.94705170255181 19.862977511274
3.55648030622313 10.5778143049769
4.29193426012878 5.94110288155885
5.17947467923121 3.63939425702671
6.25055192527397 2.39342664775321
7.54312006335462 1.65916488735831
9.10298177991522 1.19582725471957
10.9854114198756 0.887384801349676
13.2571136559011 0.673325401633077
15.9985871960606 0.519848961083457
19.3069772888325 0.406791629678848
23.2995181051537 0.321752408402233
28.1176869797423 0.256685675943367
33.9322177189533 0.206193973416077
40.9491506238043 0.16656464962177
49.4171336132384 0.135164483445808
59.6362331659464 0.110091021497339
71.9685673001152 0.0899410756330421
86.8511373751353 0.0736620986339365
104.811313415469 0.0604522596995338
126.48552168553 0.0496953721485294
152.641796717523 0.0409091649076387
184.206996932672 0.0337148500772355
222.29964825262 0.0278119322005799
268.269579527972 0.0229603870277806
323.745754281764 0.0189673605412855
390.693993705462 0.0156770922785343
471.486636345739 0.0129632854361184
568.98660290183 0.0107231621674397
686.6488450043 0.00887282498737263
828.642772854684 0.00734359997892731
1000 0.00607920663033723
};
\addplot [semithick, green!50!black, dashed]
table {%
0.1 1.19550919945874
0.120679264063933 1.46325971124632
0.145634847750124 1.79641736653953
0.175751062485479 2.21345077907609
0.212095088792019 2.7392711172963
0.255954792269954 3.408019082182
0.308884359647748 4.26718840295996
0.372759372031494 5.38416360417413
0.449843266896945 6.85621553476831
0.542867543932386 8.82676022819186
0.655128556859551 11.5089344179408
0.79060432109077 15.2224607489593
0.954095476349994 20.4278788565068
1.15139539932645 27.6684087041843
1.38949549437314 36.9464269106824
1.67683293681101 44.632273356272
2.02358964772516 38.7980099479776
2.44205309454865 22.0399763489084
2.94705170255181 11.1497474373205
3.55648030622313 6.10528513891113
4.29193426012878 3.68779273273984
5.17947467923121 2.40426197423158
6.25055192527397 1.65693514858014
7.54312006335462 1.18627139987635
9.10298177991522 0.877262054805649
10.9854114198756 0.663861092498341
13.2571136559011 0.51141259727887
15.9985871960606 0.39952639082732
19.3069772888325 0.315592850763479
23.2995181051537 0.25150692819269
28.1176869797423 0.201862287496301
33.9322177189533 0.162953242824204
40.9491506238043 0.132159986336888
49.4171336132384 0.107594882091577
59.6362331659464 0.0878691694422117
71.9685673001152 0.0719433831368076
86.8511373751353 0.0590272311456807
104.811313415469 0.0485140013867679
126.48552168553 0.0399300926434254
152.641796717523 0.0329035593891635
184.206996932672 0.0271397532513484
222.29964825262 0.022403518799563
268.269579527972 0.0185061588143354
323.745754281764 0.0152949369391475
390.693993705462 0.0126468854086958
471.486636345739 0.0104610972613051
568.98660290183 0.00865555833369224
686.6488450043 0.0071638475626079
828.642772854684 0.00593007672653467
1000 0.00491039483190292
};
\addplot [semithick, black]
table {%
0.1 0.983162229006957
0.120679264063933 0.979739168871433
0.145634847750124 0.975634669875599
0.175751062485479 0.97071954633219
0.212095088792019 0.964843007841849
0.255954792269954 0.957830344789851
0.308884359647748 0.949480988750977
0.372759372031494 0.939567324025174
0.449843266896945 0.927834779489269
0.542867543932386 0.914003910578825
0.655128556859551 0.897775373271491
0.79060432109077 0.878838857968365
0.954095476349994 0.856887123244038
1.15139539932645 0.831636139324378
1.38949549437314 0.802851872004329
1.67683293681101 0.770383247125935
2.02358964772516 0.734199227582066
2.44205309454865 0.69442577580459
2.94705170255181 0.651376170536173
3.55648030622313 0.605566535498387
4.29193426012878 0.557708670731715
5.17947467923121 0.508675333546975
6.25055192527397 0.459439003661934
7.54312006335462 0.410992359977918
9.10298177991522 0.364264408562407
10.9854114198756 0.320047784619594
13.2571136559011 0.278949355603
15.9985871960606 0.241369444027752
19.3069772888325 0.20750774621329
23.2995181051537 0.177388953691215
28.1176869797423 0.150899228995837
33.9322177189533 0.127825534072808
40.9491506238043 0.107892104666337
49.4171336132384 0.0907909103499174
59.6362331659464 0.0762050194485786
71.9685673001152 0.0638251548762098
86.8511373751353 0.0533604416354901
104.811313415469 0.044544590636133
126.48552168553 0.0371387268783343
152.641796717523 0.0309318963916211
184.206996932672 0.0257400702291545
222.29964825262 0.0214042558102376
268.269579527972 0.0177881489444466
323.745754281764 0.0147756201513168
390.693993705462 0.0122682242268112
471.486636345739 0.0101828468074734
568.98660290183 0.00844954957380373
686.6488450043 0.00700964095732259
828.642772854684 0.0058139769141976
1000 0.0048214828027241
};
\end{axis}

\end{tikzpicture}
\begin{tikzpicture}

\tikzstyle{every node}=[font=\tiny]
\begin{axis}[
width=\figwidth,
height=\figheight,
legend cell align={left},
legend style={fill opacity=0.8, draw opacity=1, text opacity=1, draw=white!80!black},
log basis x={10},
log basis y={10},
tick align=outside,
tick pos=left,
x grid style={white!69.0196078431373!black},
xmajorgrids,
xmin=0.0630957344480193, xmax=1584.89319246111,
xmode=log,
xtick style={color=black},
y grid style={white!69.0196078431373!black},
ymajorgrids,
ymin=0.00317292502391139, ymax=31.5731329379902,
ymode=log,
ytick style={color=black}
]
\addplot [semithick, blue]
table {%
0.1 0.966557766977209
0.120679264063933 1.17521351554781
0.145634847750124 1.43125761616471
0.175751062485479 1.74731150273775
0.212095088792019 2.13310758717616
0.255954792269954 2.61439482183268
0.308884359647748 3.21166481258717
0.372759372031494 3.96031853342076
0.449843266896945 4.90144217157462
0.542867543932386 6.10212628241815
0.655128556859551 7.63407210417236
0.79060432109077 9.58175655692805
0.954095476349994 12.0689498160962
1.15139539932645 15.1431077079495
1.38949549437314 18.5527212979592
1.67683293681101 20.7776710363105
2.02358964772516 17.8164299304324
2.44205309454865 11.1870986404611
2.94705170255181 6.5158483057983
3.55648030622313 4.0142088632698
4.29193426012878 2.64853785491123
5.17947467923121 1.84461861825727
6.25055192527397 1.33757601137859
7.54312006335462 0.998172249836798
9.10298177991522 0.761229019942763
10.9854114198756 0.590759565221016
13.2571136559011 0.463931815613217
15.9985871960606 0.368365773783483
19.3069772888325 0.294516246956954
23.2995181051537 0.237306199519375
28.1176869797423 0.192157014836074
33.9322177189533 0.156241979710755
40.9491506238043 0.127469256790124
49.4171336132384 0.104283825641993
59.6362331659464 0.0855104527920114
71.9685673001152 0.0702485143144611
86.8511373751353 0.057739970989835
104.811313415469 0.0475707333009276
126.48552168553 0.0392338346561185
152.641796717523 0.0323863529838944
184.206996932672 0.0267535225375655
222.29964825262 0.0221139069278777
268.269579527972 0.0182883480648146
323.745754281764 0.0151311146301298
390.693993705462 0.0125235589111616
471.486636345739 0.0103686484897207
568.98660290183 0.00858705251057557
686.6488450043 0.0071131827054558
828.642772854684 0.00589356838391675
1000 0.00488401484976642
};
\addplot [semithick, blue, dashed]
table {%
0.1 0.944061631654746
0.120679264063933 1.10933191129846
0.145634847750124 1.29997289202629
0.175751062485479 1.52044689460985
0.212095088792019 1.76905675513549
0.255954792269954 2.0523794083922
0.308884359647748 2.37587833912414
0.372759372031494 2.73855314365498
0.449843266896945 3.14035285405844
0.542867543932386 3.57923100107771
0.655128556859551 4.04467321489259
0.79060432109077 4.51612973510611
0.954095476349994 4.96846688720795
1.15139539932645 5.36483812035461
1.38949549437314 5.71870453173202
1.67683293681101 6.12806746184432
2.02358964772516 7.20615998759594
2.44205309454865 9.50036390251717
2.94705170255181 10.5980647194924
3.55648030622313 7.33660214329336
4.29193426012878 4.28935134630652
5.17947467923121 2.62899957156173
6.25055192527397 1.73776547731381
7.54312006335462 1.21760801640616
9.10298177991522 0.887246687413533
10.9854114198756 0.665466521101142
13.2571136559011 0.509722448044279
15.9985871960606 0.396759576677214
19.3069772888325 0.312979038673751
23.2995181051537 0.2492485108783
28.1176869797423 0.199819332658247
33.9322177189533 0.161304610943046
40.9491506238043 0.130830757491676
49.4171336132384 0.106427890907729
59.6362331659464 0.086982253154303
71.9685673001152 0.0712386977908746
86.8511373751353 0.0584587048255747
104.811313415469 0.0480620412305939
126.48552168553 0.0395696739874496
152.641796717523 0.0326162724177734
184.206996932672 0.0269111812634195
222.29964825262 0.0222217894824472
268.269579527972 0.0183619122538772
323.745754281764 0.0151816367255595
390.693993705462 0.0125648140237229
471.486636345739 0.0103886414047054
568.98660290183 0.00860036845354561
686.6488450043 0.00712225314603421
828.642772854684 0.00589967035202266
1000 0.0048880982881887
};
\addplot [semithick, black]
table {%
0.1 0.983162229006957
0.120679264063933 0.979739168871433
0.145634847750124 0.975634669875599
0.175751062485479 0.97071954633219
0.212095088792019 0.964843007841849
0.255954792269954 0.957830344789851
0.308884359647748 0.949480988750977
0.372759372031494 0.939567324025174
0.449843266896945 0.927834779489269
0.542867543932386 0.914003910578825
0.655128556859551 0.897775373271491
0.79060432109077 0.878838857968365
0.954095476349994 0.856887123244038
1.15139539932645 0.831636139324378
1.38949549437314 0.802851872004329
1.67683293681101 0.770383247125935
2.02358964772516 0.734199227582066
2.44205309454865 0.69442577580459
2.94705170255181 0.651376170536173
3.55648030622313 0.605566535498387
4.29193426012878 0.557708670731715
5.17947467923121 0.508675333546975
6.25055192527397 0.459439003661934
7.54312006335462 0.410992359977918
9.10298177991522 0.364264408562407
10.9854114198756 0.320047784619594
13.2571136559011 0.278949355603
15.9985871960606 0.241369444027752
19.3069772888325 0.20750774621329
23.2995181051537 0.177388953691215
28.1176869797423 0.150899228995837
33.9322177189533 0.127825534072808
40.9491506238043 0.107892104666337
49.4171336132384 0.0907909103499174
59.6362331659464 0.0762050194485786
71.9685673001152 0.0638251548762098
86.8511373751353 0.0533604416354901
104.811313415469 0.044544590636133
126.48552168553 0.0371387268783343
152.641796717523 0.0309318963916211
184.206996932672 0.0257400702291545
222.29964825262 0.0214042558102376
268.269579527972 0.0177881489444466
323.745754281764 0.0147756201513168
390.693993705462 0.0122682242268112
471.486636345739 0.0101828468074734
568.98660290183 0.00844954957380373
686.6488450043 0.00700964095732259
828.642772854684 0.0058139769141976
1000 0.0048214828027241
};
\end{axis}

\end{tikzpicture}
\begin{tikzpicture}

\definecolor{color0}{rgb}{1,0.647058823529412,0}

\tikzstyle{every node}=[font=\tiny]
\begin{axis}[
width=\figwidth,
height=\figheight,
legend cell align={left},
legend style={fill opacity=0.8, draw opacity=1, text opacity=1, draw=white!80!black},
log basis x={10},
log basis y={10},
tick align=outside,
tick pos=left,
x grid style={white!69.0196078431373!black},
xlabel={\(\displaystyle n / d\)},
xmajorgrids,
xmin=0.0630957344480193, xmax=1584.89319246111,
xmode=log,
xtick style={color=black},
y grid style={white!69.0196078431373!black},
ylabel={Variance},
ymajorgrids,
ymin=0.00369579942775541, ymax=1.28261824595928,
ymode=log,
ytick style={color=black}
]
\addplot [semithick, red]
table {%
0.1 0.0155638581356856
0.120679264063933 0.0186559152932359
0.145634847750124 0.0223309956927142
0.175751062485479 0.0266866248823507
0.212095088792019 0.0318288271858709
0.255954792269954 0.0378712676824971
0.308884359647748 0.0449349086520908
0.372759372031494 0.053137023920021
0.449843266896945 0.0625855319892355
0.542867543932386 0.0733634783279273
0.655128556859551 0.0855224466277829
0.79060432109077 0.0990437695978249
0.954095476349994 0.113865747331795
1.15139539932645 0.129717775105339
1.38949549437314 0.146278586233567
1.67683293681101 0.163074767920841
2.02358964772516 0.179460081881536
2.44205309454865 0.194685957331992
2.94705170255181 0.20793612885655
3.55648030622313 0.218396325095059
4.29193426012878 0.225367437629855
5.17947467923121 0.228362958968941
6.25055192527397 0.227168467707192
7.54312006335462 0.221880440142387
9.10298177991522 0.212891079187607
10.9854114198756 0.200821439425996
13.2571136559011 0.186605503959001
15.9985871960606 0.170849412452091
19.3069772888325 0.154379410889666
23.2995181051537 0.137823604743969
28.1176869797423 0.121888245728416
33.9322177189533 0.106692963258957
40.9491506238043 0.0926310483970932
49.4171336132384 0.0798513644755569
59.6362331659464 0.0684138832555966
71.9685673001152 0.0583041119154657
86.8511373751353 0.0494662290644552
104.811313415469 0.0418072893410969
126.48552168553 0.035219218803146
152.641796717523 0.0296736036000639
184.206996932672 0.0248826688155026
222.29964825262 0.0208257072253317
268.269579527972 0.0174025104266701
323.745754281764 0.0145230150997252
390.693993705462 0.0121067993093823
471.486636345739 0.0100836799884002
568.98660290183 0.00839235403366279
686.6488450043 0.0069805399547257
828.642772854684 0.00580341322353395
1000 0.00482292444115262
};
\addplot [semithick, red, dashed]
table {%
0.1 0.00909686135593953
0.120679264063933 0.0109372289694956
0.145634847750124 0.0131399204347034
0.175751062485479 0.0157713509724816
0.212095088792019 0.0189088024980041
0.255954792269954 0.0226399858025803
0.308884359647748 0.0270627775221041
0.372759372031494 0.0322866606076644
0.449843266896945 0.0384280285749114
0.542867543932386 0.0456074852268559
0.655128556859551 0.0539422197624675
0.79060432109077 0.0635488191008923
0.954095476349994 0.0744862158302118
1.15139539932645 0.086618734525503
1.38949549437314 0.100147935382736
1.67683293681101 0.114800829650218
2.02358964772516 0.130272399060326
2.44205309454865 0.146092382131443
2.94705170255181 0.161610673532529
3.55648030622313 0.176026471526084
4.29193426012878 0.188425074137458
5.17947467923121 0.19789644712621
6.25055192527397 0.203645462651308
7.54312006335462 0.205324546602285
9.10298177991522 0.20226111652285
10.9854114198756 0.195221890685964
13.2571136559011 0.184624004223208
15.9985871960606 0.171323971706797
19.3069772888325 0.156491558126068
23.2995181051537 0.140624824050149
28.1176869797423 0.124690481887159
33.9322177189533 0.109315771163764
40.9491506238043 0.0949244050581419
49.4171336132384 0.081749666581024
59.6362331659464 0.0699240818678678
71.9685673001152 0.0594743119704647
86.8511373751353 0.0503595070902931
104.811313415469 0.0424781738795933
126.48552168553 0.035760598746991
152.641796717523 0.0299965470737881
184.206996932672 0.0251168496416314
222.29964825262 0.0209895631706902
268.269579527972 0.0175227843130966
323.745754281764 0.0146065636609806
390.693993705462 0.0121637631101733
471.486636345739 0.0101219324110974
568.98660290183 0.00842017641989967
686.6488450043 0.00700016368641054
828.642772854684 0.00581668307659344
1000 0.00483165215132042
};
\addplot [semithick, color0, dashed]
table {%
0.1 0.0125625301523949
0.120679264063933 0.015085525353611
0.145634847750124 0.0180968331492457
0.175751062485479 0.0216827258936291
0.212095088792019 0.025941008971133
0.255954792269954 0.0309808119150132
0.308884359647748 0.0369213954199017
0.372759372031494 0.0438860913849563
0.449843266896945 0.0520085166709946
0.542867543932386 0.0614100159181005
0.655128556859551 0.0721966985326226
0.79060432109077 0.0844410930601771
0.954095476349994 0.098160714340956
1.15139539932645 0.113291758664125
1.38949549437314 0.129659706219566
1.67683293681101 0.146950336544342
2.02358964772516 0.164684046798321
2.44205309454865 0.182220521684513
2.94705170255181 0.198762274306093
3.55648030622313 0.213414651706051
4.29193426012878 0.22526916172708
5.17947467923121 0.233514185276623
6.25055192527397 0.237550256088523
7.54312006335462 0.237083408650114
9.10298177991522 0.232170769386788
10.9854114198756 0.223207090833675
13.2571136559011 0.210856723321013
15.9985871960606 0.195957236421866
19.3069772888325 0.179390130568423
23.2995181051537 0.162006680147028
28.1176869797423 0.144537669103603
33.9322177189533 0.127575303426407
40.9491506238043 0.111547983317792
49.4171336132384 0.09673989517135
59.6362331659464 0.0833076230110363
71.9685673001152 0.0713080977736558
86.8511373751353 0.0607217740738203
104.811313415469 0.0514803821218912
126.48552168553 0.0434833556448283
152.641796717523 0.0366134177401856
184.206996932672 0.0307474686239445
222.29964825262 0.0257640665589848
268.269579527972 0.0215482246377718
323.745754281764 0.017994222158087
390.693993705462 0.0150068732135894
471.486636345739 0.0125019369778295
568.98660290183 0.0104057156876852
686.6488450043 0.00865446898234756
828.642772854684 0.00719351051001627
1000 0.00597603001596892
};
\addplot [semithick, green!50!black, dashed]
table {%
0.1 0.0154699375991631
0.120679264063933 0.0185584052256285
0.145634847750124 0.0222364985368924
0.175751062485479 0.0266048287481349
0.212095088792019 0.0317757950302325
0.255954792269954 0.0378724980253431
0.308884359647748 0.0449917361770684
0.372759372031494 0.0533144778631818
0.449843266896945 0.0629451810706532
0.542867543932386 0.0739906660427988
0.655128556859551 0.086522336856043
0.79060432109077 0.100553769408942
0.954095476349994 0.116013381049287
1.15139539932645 0.132713785357533
1.38949549437314 0.150321846107862
1.67683293681101 0.168336082640527
2.02358964772516 0.186004795692643
2.44205309454865 0.2026219211616
2.94705170255181 0.217211082960067
3.55648030622313 0.228844217046475
4.29193426012878 0.236706651431606
5.17947467923121 0.240215816121814
6.25055192527397 0.239114905409771
7.54312006335462 0.233512879526087
9.10298177991522 0.223860494194294
10.9854114198756 0.21087455924641
13.2571136559011 0.195430247372186
15.9985871960606 0.178439103467576
19.3069772888325 0.160749098174972
23.2995181051537 0.143090730590889
28.1176869797423 0.126038957918473
33.9322177189533 0.110002289219269
40.9491506238043 0.0952445336158236
49.4171336132384 0.0819033557665105
59.6362331659464 0.0700196957672649
71.9685673001152 0.0595613441695341
86.8511373751353 0.0504506942735072
104.811313415469 0.0425806975382969
126.48552168553 0.0358299310190379
152.641796717523 0.0300729362845309
184.206996932672 0.0251872457814439
222.29964825262 0.0210576515337291
268.269579527972 0.017578889815173
323.745754281764 0.0146566040576501
390.693993705462 0.0122074009832884
471.486636345739 0.010158843711372
568.98660290183 0.00844782628343087
686.6488450043 0.00702099420004653
828.642772854684 0.00583233735214072
1000 0.00484253032227988
};
\addplot [semithick, black]
table {%
0.1 0.983162229006957
0.120679264063933 0.979739168871433
0.145634847750124 0.975634669875599
0.175751062485479 0.97071954633219
0.212095088792019 0.964843007841849
0.255954792269954 0.957830344789851
0.308884359647748 0.949480988750977
0.372759372031494 0.939567324025174
0.449843266896945 0.927834779489269
0.542867543932386 0.914003910578825
0.655128556859551 0.897775373271491
0.79060432109077 0.878838857968365
0.954095476349994 0.856887123244038
1.15139539932645 0.831636139324378
1.38949549437314 0.802851872004329
1.67683293681101 0.770383247125935
2.02358964772516 0.734199227582066
2.44205309454865 0.69442577580459
2.94705170255181 0.651376170536173
3.55648030622313 0.605566535498387
4.29193426012878 0.557708670731715
5.17947467923121 0.508675333546975
6.25055192527397 0.459439003661934
7.54312006335462 0.410992359977918
9.10298177991522 0.364264408562407
10.9854114198756 0.320047784619594
13.2571136559011 0.278949355603
15.9985871960606 0.241369444027752
19.3069772888325 0.20750774621329
23.2995181051537 0.177388953691215
28.1176869797423 0.150899228995837
33.9322177189533 0.127825534072808
40.9491506238043 0.107892104666337
49.4171336132384 0.0907909103499174
59.6362331659464 0.0762050194485786
71.9685673001152 0.0638251548762098
86.8511373751353 0.0533604416354901
104.811313415469 0.044544590636133
126.48552168553 0.0371387268783343
152.641796717523 0.0309318963916211
184.206996932672 0.0257400702291545
222.29964825262 0.0214042558102376
268.269579527972 0.0177881489444466
323.745754281764 0.0147756201513168
390.693993705462 0.0122682242268112
471.486636345739 0.0101828468074734
568.98660290183 0.00844954957380373
686.6488450043 0.00700964095732259
828.642772854684 0.0058139769141976
1000 0.0048214828027241
};
\end{axis}

\end{tikzpicture}
\begin{tikzpicture}

\tikzstyle{every node}=[font=\tiny]
\begin{axis}[
width=\figwidth,
height=\figheight,
legend cell align={left},
legend style={fill opacity=0.8, draw opacity=1, text opacity=1, draw=white!80!black},
log basis x={10},
log basis y={10},
tick align=outside,
tick pos=left,
x grid style={white!69.0196078431373!black},
xlabel={\(\displaystyle n / d\)},
xmajorgrids,
xmin=0.0630957344480193, xmax=1584.89319246111,
xmode=log,
xtick style={color=black},
y grid style={white!69.0196078431373!black},
ymajorgrids,
ymin=0.00369021829594777, ymax=1.28271055319768,
ymode=log,
ytick style={color=black}
]
\addplot [semithick, blue]
table {%
0.1 0.0129463508123306
0.120679264063933 0.0155385887869398
0.145634847750124 0.0186284725019873
0.175751062485479 0.022303162747177
0.212095088792019 0.0266592824391986
0.255954792269954 0.0318036078717115
0.308884359647748 0.0378513129984255
0.372759372031494 0.044921696759485
0.449843266896945 0.0531324600382492
0.542867543932386 0.0625908685781179
0.655128556859551 0.0733977019730803
0.79060432109077 0.0855568915125542
0.954095476349994 0.0990540527228056
1.15139539932645 0.11376063829533
1.38949549437314 0.129427607915932
1.67683293681101 0.145683016540204
2.02358964772516 0.161966929287979
2.44205309454865 0.17757753487539
2.94705170255181 0.191702315054297
3.55648030622313 0.203508818181127
4.29193426012878 0.212167465424139
5.17947467923121 0.217063365474673
6.25055192527397 0.217822769296982
7.54312006335462 0.214408456374783
9.10298177991522 0.207105504882003
10.9854114198756 0.196466386672242
13.2571136559011 0.183637910999421
15.9985871960606 0.168795836595702
19.3069772888325 0.152962259300189
23.2995181051537 0.136880076311368
28.1176869797423 0.121299609447017
33.9322177189533 0.106313475008277
40.9491506238043 0.0923951994922144
49.4171336132384 0.0797098038288585
59.6362331659464 0.0683323837349226
71.9685673001152 0.0582634128306743
86.8511373751353 0.0494506842662814
104.811313415469 0.0418183607156287
126.48552168553 0.0352445798903555
152.641796717523 0.0296517525075296
184.206996932672 0.0248645973622315
222.29964825262 0.0208105589325955
268.269579527972 0.0173894616941606
323.745754281764 0.0145151579397602
390.693993705462 0.0121010008330292
471.486636345739 0.0100784318355116
568.98660290183 0.00838797229219079
686.6488450043 0.00697661963073659
828.642772854684 0.00579989414790105
1000 0.00481974951387765
};
\addplot [semithick, blue, dashed]
table {%
0.1 0.0156029813439024
0.120679264063933 0.0187116666996585
0.145634847750124 0.0224115192277514
0.175751062485479 0.0268016138066082
0.212095088792019 0.0319915684696598
0.255954792269954 0.0381026826991911
0.308884359647748 0.0452603781995377
0.372759372031494 0.05359165336604
0.449843266896945 0.0632157994654353
0.542867543932386 0.0742297389455237
0.655128556859551 0.0866936960452798
0.79060432109077 0.100604221560436
0.954095476349994 0.115874256347277
1.15139539932645 0.13229680372307
1.38949549437314 0.14950553027498
1.67683293681101 0.166990378808604
2.02358964772516 0.184104734826234
2.44205309454865 0.199912610659277
2.94705170255181 0.213544767021951
3.55648030622313 0.224119619934797
4.29193426012878 0.23089170102404
5.17947467923121 0.233314046676161
6.25055192527397 0.231290569155314
7.54312006335462 0.225007569641793
9.10298177991522 0.214978492651075
10.9854114198756 0.201948439440803
13.2571136559011 0.186755634810266
15.9985871960606 0.170580971027873
19.3069772888325 0.153682903753089
23.2995181051537 0.136908780781964
28.1176869797423 0.120778076022652
33.9322177189533 0.105612507771179
40.9491506238043 0.0917709836026084
49.4171336132384 0.0791265313356452
59.6362331659464 0.0678280580014566
71.9685673001152 0.0578494902845373
86.8511373751353 0.0491209933125482
104.811313415469 0.0415447007720138
126.48552168553 0.0350312657975479
152.641796717523 0.0294615489270297
184.206996932672 0.024716763628201
222.29964825262 0.020729093969671
268.269579527972 0.0173313393181082
323.745754281764 0.0144698146730505
390.693993705462 0.012066655621024
471.486636345739 0.010053246941265
568.98660290183 0.00837287213590454
686.6488450043 0.00696590033775557
828.642772854684 0.00579227622914202
1000 0.00481454821204474
};
\addplot [semithick, black]
table {%
0.1 0.983162229006957
0.120679264063933 0.979739168871433
0.145634847750124 0.975634669875599
0.175751062485479 0.97071954633219
0.212095088792019 0.964843007841849
0.255954792269954 0.957830344789851
0.308884359647748 0.949480988750977
0.372759372031494 0.939567324025174
0.449843266896945 0.927834779489269
0.542867543932386 0.914003910578825
0.655128556859551 0.897775373271491
0.79060432109077 0.878838857968365
0.954095476349994 0.856887123244038
1.15139539932645 0.831636139324378
1.38949549437314 0.802851872004329
1.67683293681101 0.770383247125935
2.02358964772516 0.734199227582066
2.44205309454865 0.69442577580459
2.94705170255181 0.651376170536173
3.55648030622313 0.605566535498387
4.29193426012878 0.557708670731715
5.17947467923121 0.508675333546975
6.25055192527397 0.459439003661934
7.54312006335462 0.410992359977918
9.10298177991522 0.364264408562407
10.9854114198756 0.320047784619594
13.2571136559011 0.278949355603
15.9985871960606 0.241369444027752
19.3069772888325 0.20750774621329
23.2995181051537 0.177388953691215
28.1176869797423 0.150899228995837
33.9322177189533 0.127825534072808
40.9491506238043 0.107892104666337
49.4171336132384 0.0907909103499174
59.6362331659464 0.0762050194485786
71.9685673001152 0.0638251548762098
86.8511373751353 0.0533604416354901
104.811313415469 0.044544590636133
126.48552168553 0.0371387268783343
152.641796717523 0.0309318963916211
184.206996932672 0.0257400702291545
222.29964825262 0.0214042558102376
268.269579527972 0.0177881489444466
323.745754281764 0.0147756201513168
390.693993705462 0.0122682242268112
471.486636345739 0.0101828468074734
568.98660290183 0.00844954957380373
686.6488450043 0.00700964095732259
828.642772854684 0.0058139769141976
1000 0.0048214828027241
};
\end{axis}

\end{tikzpicture}

    \caption{Variance for logistic regression at $\lambda = 10^{-2}$ (Top) and $\lambda = 1$ (Bottom). Left: variance of full resampling, pair bootstrap, subsampling. Right: variance of label resampling, residual bootstrap. See \cref{fig:variance_ridge} for the legend.
    }
    \label{fig:variance_logistic}
\end{figure}
\subsection{Logistic regression}
\label{sec:logistic_numerical_results}

Our results extend beyond ridge regression, and the quantities of interest can be computed for any convex loss. \Cref{fig:variance_logistic} displays the true variances and their estimation for regularized logistic regression with $\lambda \in \{ 10^{-2}, 1 \}$, similarly to~\cref{fig:variance_ridge}. However, contrary to the ridge case, $\lambda = 1$ yields the maximum-a-posteriori estimator but does not minimize the misclassification error. 

Qualitatively, we observe similar results as for ridge regression : at large $\alpha$, all methods consistently estimate the true variance and the Jackknife provides an upper bound of $\varianceOnXY$. Moreover, at low $\alpha$, regularization improves the estimation of the variance, even though $\lambda$ is not optimal.

Finally, at $\lambda = 0.01$ for both ridge and logistic regression, we observe a local maximum in the true and resampled bias and variance around $d = n$. This behavior is reminiscent of the double-descent behavior observed e.g. in random features models or neural networks : the test error achieves a local maximum at the interpolation threshold where the model can perfectly fit the training data, then decreases with the number of parameters. Moreover, we see that regularization can mitigate this ``double-descent'' phenomenon.

\section{Conclusion \& Perspectives}
In this work, we have provided an exact asymptotic comparison of the uncertainty estimations provided by different resampling methods, in the context of high-dimensional regularized maximum likelihood with generalized linear models.

Our results highlight the limitations of these methods in the high-dimensional regime relevant to modern machine learning practice and discuss how cross-validation can, to some extent, mitigate some of these limitations. 

Avenues for future work are manifold. For instance, how would our results change in a misspecified scenario? Can structure in the data help or hinder resampling methods? These interesting questions are left for future investigation. 

\section*{Acknowledgements}
This research was supported by the Swiss National Science Foundation grant SNFS OperaGOST, $200021\_200390$ and the NCCR MARVEL, a National Centre of Competence in Research, funded by the Swiss National Science Foundation (grant number 205602).


\bibliography{bibliography}

\clearpage
\appendix
\onecolumn

\section{Derivation of the results for pair resampling}
\label{appendix:se_pair_resampling}

In this appendix we show how the self-consistent equations~\eqref{eq:thm_se_overlaps} and~\eqref{eq:thm_se_hat_overlaps} can be derived from the state-evolution equation of GAMP (Generalized Approximate Message Passing), and how to extend them to generic log-concave losses. 

As stated in \cref{sec:technical}, the key observation is that in order to asymptotically characterize the biases and variances associated with any of the resampling methods in \cref{sec:setting}, it is sufficient to characterize only the correlation $\werm(\dataset^{\star}_b)^{\top} \werm(\dataset^{\star}_{b'})$ between two resampled datasets $\dataset^{\star}_b, \dataset^{\star}_{b'}$. Indeed, the resampling variances can be written 
\begin{equation}
    \widehat{\Var} = \frac1d\left(\frac{1}{B}\sum\limits_{b=1}^{B}\lVert\hat{\vec{\theta}}_{b}\lVert^{2} - \frac{1}{B^2}\sum\limits_{b,b'=1}^{B}\hat{\vec{\theta}}_{b}^{\top}\hat{\vec{\theta}}_{b'}\right).
\end{equation}
It is natural to study these variances in the limit $B \to \infty$. In that limit, $\widehat{\Var}$ converges to
\begin{equation*}
    \widehat{\Var} = \frac{1}{d} \mathbb{E}_{\dataset^{\star}}\left[ \| \hatw(\dataset^{\star}) \|^2 \right] - \frac{1}{d} \mathbb{E}_{\dataset^{\star}, \dataset^{\star '}} \left[ \hatw(\dataset^{\star}) \hatw(\dataset^{\star '}) \right]
\end{equation*}
where the expectations are over resampled dataset conditioned on $\dataset$ and where the resampling depends on the method considered. In a similar way for the bias 
\begin{align*}
    \widehat{\Bias^2} &= \frac{1}{d} \left\| \frac{1}{B} \sum_{b = 1}^B \hat{\vec{\theta}}_{b} - \hatw \right\|^2 \\
    &\stackrel{B\to\infty}{\rightarrow} \frac{1}{d} \left( \| \hatw \|^2 + \left\| \mathbb{E}_{\dataset^{\star}} \left[ \hatw(\dataset^{\star}) \right] \right\|^2 \right) 
\end{align*}

To do so, we observe that computing the ERM estimator on a resampled dataset $\dataset^{\star}$ is equivalent to solving an wERM problem \cref{eq:def_weighted_erm}, where for each sample $(\vec{x}_{i},y_{i})\in\mathcal{D}$, we introduce a sample weight $p_{i}$. The distribution on the sample weights depends on the way $\dataset$ is resampled: for example, with $p_{i}=1$ for all $i\in[n]$, this reduces to standard MLE \eqref{eq:def_erm}. On the other hand, by choosing $p_{i}\in\{0,1\}$ at random from a Bernoulli distribution with probability $r\in(0,1]$, the wERM \eqref{eq:def_weighted_erm} asymptotically corresponds to doing subsampling. Also, pair bootstrap is asymptotically equivalent to taking $p_{i}\sim \Pois(1)$ independently.
The problem is thus to compute the correlation between estimators $\werm(\dataset, \Vec{p})$ trained with different, possibly correlated vectors $\Vec{p}$.

The use of GAMP for deriving high-dimensional asymptotics characterization is now a classic rigorous tool, that has been used in many situations \citep{bayati2011lasso,JMLR:v15:javanmard14a,sur2019likelihood,emami2020generalization,loureiro2021learning,Loureiro2022_ensembling,gerbelot2022asymptotic}. The idea is to proceed in two steps: i) to propose a GAMP algorithm that solves the optimisation problem asymptotically, and ii) to use the fact that GAMP performance can be tracked with a rigorous state evolution \cite{bayati2011dynamics,gerbelot2023graph}
. This was, to the best of our knwoldege, introduced first in \citep{bayati2011lasso} for studying the LASSO risk. We shall not repeat the proof technique, and refer the reader to \citep{loureiro2021learning,Loureiro2022_ensembling} for details with our current notation. Our results directly uses Thm. 1 in \citep{loureiro2021learning} or Thm 2.1 in \cite{Loureiro2022_ensembling}.

The novelty of our approach consists in adapting these results to the bootstrap situation by introducing sample weights $\vec{p}$ and studying the performance of GAMP for several estimators. The properties of the estimators are given by the distribution on the weights $\Vec{p}$. All previous proof still trivially apply: indeed the state evolution theorems  generalize to vector estimations \cite{javanmard2013state}, and, since GAMP is applied to two problems in parallel, the convergence guarantees still independently apply to each of them. A similar strategy was used in \cite{Loureiro2022_ensembling}.

\begin{algorithm}
    \caption{GAMP with sample weights}
    \begin{algorithmic}
        \STATE \textbf{Input:} $\mat{X} \in \mathbb{R}^{n \times d}$, $\Vec{y} \in \mathbb{R}^n$, and $\Vec{p}_{\mu} \in \mathbb{R}^{B} \text{ for } 1 \leq \mu \leq n$
        \STATE \textbf{Initialize:} ${\channel}_{\mu}^{(0)} = \Vec{0} \text{ for } 1 \leq \mu \leq n$, $\quad\mat{A}_i^{(0)} = \mat{I}_{B} \text{ for } 1 \leq i \leq d$
        \STATE \textbf{Initialize:} ${\hat{\vec{\theta}}}_{i}^{(1)} \in\reals^B \text{ and } \hat{\mat{C}}_i^{(1)} \in\reals^{B\times B} \text{ for } 1 \leq i \leq d$
        \STATE \textbf{Repeat for $t=1, 2, \dots$:}
            \STATE \quad // Update of the means $\vec{\omega}_{\mu} \in \RR^{B}$ and covariances $\mat{V}_{\mu} \in \mathcal{S}_B^{+}\text{ for } 1 \leq \mu \leq n$: 
            \STATE \quad $\Vec{\omega}_{\mu}^{(t)} = \sum_{i=1}^d X_{\mu, i} \hat{\vec{\theta}}^{(t)}_i - X_{\mu, i}^2 \left(\mat{A}_i^{(t-1)}\right)^{-1} \hat{\mat{C}}_i^{(t)} \mat{A}_i^{(t-1)} {\channel}_{\mu}^{(t-1)}$ $|$ $\mat{V}_{\mu}^{(t)} = \sum_{i = 1}^d X_{\mu, i}^2 \hat{\mat{C}}_i^{(t)}$
            \STATE \quad // Update of ${\channel}_{\mu}$ and ${\partial_{\omega}\channel}_{\mu}\text{ for } 1 \leq \mu \leq n:$
            \STATE \quad ${\channel}_{\mu}^{(t)} = \channel \left( \Vec{\omega}_{\mu}^{(t)}, y_{\mu}, \mat{V}_{\mu}^{(t)}, \Vec{p}_{\mu} \right)$ $|$ $\partial_{\vec{\omega}} {\channel}_{\mu}^{(t)} = \partial_{\vec{\omega}} {\channel} \left( \Vec{\omega}_{\mu}^{(t)}, y_{\mu}, \mat{V}_{\mu}^{(t)}, \Vec{p}_{\mu} \right)$
            \STATE \quad // Update of means $\Vec{b}_i \in \mathbb{R}^{B}$ and covariances $\mat{A}_i \in \RR^{B \times B}\text{ for } 1 \leq i \leq d:$ 
            \STATE \quad $\mat{A}_i^{(t)} = -\sum_{\mu = 1}^n X_{\mu, i}^2 \partial_{\omega} {\channel}_{\mu}^{(t)}$ $|$ $\vec{b}_i^{(t)} = \mat{A}_i^{(t)}\hat{\vec{\theta}}_i^{(t)} + \sum_{\mu = 1}^n X_{\mu, i} {\channel}_{\mu}^{(t)}$
            \STATE \quad // Update of the estimated marginals $\hat{\vec{\theta}}_i \in \RR^{B}$ and $\hat{\mat{C}}_i \in \RR^{B \times B}\text{ for } 1 \leq i \leq d:$
            \STATE \quad $\hat{\vec{\theta}}_i^{(t+1)} = \denoiser(\Vec{b}_i^{(t)}, \mat{A}_i^{(t)})$ $|$ $\hat{\mat{C}}_i^{(t+1)} = \partial_{\vec{b}}\Vec{f}_a(\Vec{b}_i^{(t)}, \mat{A}_i^{(t)})$
        \STATE \textbf{Until convergence}
        \STATE \textbf{Output:} $\hat{\vec{\theta}}_1, \dots, \hat{\vec{\theta}}_d$ and $\hat{\mat{C}}_1, \dots, \hat{\mat{C}}_d$
    \end{algorithmic}
    \label{alg:gamp_sample_weights}
\end{algorithm}

Consider a convex loss function $\ell$ and regularizer $r$, and the following empirical risk minimization problem 
\begin{align}
    (\hatw_1, \dots, \hatw_B) &= \arg\min_{\Vec{\theta}_1, \dots, \Vec{\theta}_B\in\reals^d} \mathcal{L} \left( \Vec{\theta}_1, \dots, \Vec{\theta}_B \right)
    \label{eq:def:weighted_erm}
\end{align}
where
\begin{equation}
    \mathcal{L} \left( \vec{\theta}_1, \dots, \vec{\theta}_B \right) \vcentcolon= \sum_{\mu = 1}^n \ell_{\Vec{p}} (y_{\mu}, \Vec{\theta}_1^{\top} \Vec{x}_{\mu}, \dots, \Vec{\theta}_B^{\top} \Vec{x}_{\mu} ) + \sum_{b = 1}^B r( \Vec{\theta}_b )
\end{equation}
and
\begin{equation}
    \ell_{\vec{p}}(y, z_1, \dots, z_B) \vcentcolon= \sum_{b = 1}^B p_b \ell(y, z_b)
\end{equation}

We define a \textit{channel function} associated to the function $\ell$ : 
\begin{align}
    \channel(y, \Vec{\omega}, \mat{V}, \Vec{p}) &= {\mat{V}}^{-1} \left( {\rm prox}_{\mat{V}, \ell_{\vec{p}}(y, \cdot)}(\Vec{\omega}) - \Vec{\omega} \right),
\end{align}
where the proximal operator is 
\begin{equation}
    {\rm prox}_{\mat{V}, \ell_{\vec{p}}(y, \cdot)}(\Vec{\omega}) = \arg\min_{\vec{z}\in\reals^B} \left(\frac{1}{2}(\Vec{z} - \Vec{\omega})^\top \mat{V}^{-1} (\Vec{z} - \Vec{\omega}) + \ell_{\vec{p}}(y, \vec{z})\right).
\end{equation}

Let us also define the \textit{denoising function} associated to the regularizer $r$:
\begin{align}
    \vec{f}_a(\vec{b}, \mat{A}) &= {\rm prox}_{\mat{A}^{-1}, r}(\mat{A}^{-1}\Vec{b}) = \arg\min_{\vec{z}\in\reals^B} \left(\frac{1}{2}(\Vec{z} - \mat{A}^{-1}\Vec{b})^\top \mat{A} (\Vec{z} - \mat{A}^{-1}\Vec{b}) + r(\vec{z})\right).
\end{align}

Using~\cref{alg:gamp_sample_weights} with this choice of channel and denoising functions returns a set of vectors $\hat{\vec{\theta}}_1, \cdots, \hat{\vec{\theta}}_d \in\reals^B$, where $\hat{\vec{\theta}}_i$ contains the $B$ estimates for $\theta_{\star i}$. Hence, these vectors allow to solve the minimization problem~\eqref{eq:def:weighted_erm}.

\paragraph{Intuition of GAMP algorithm} We are interested in solving the minimization problem~\eqref{eq:def:weighted_erm}, which is equivalent to sampling from the distribution
\begin{align}
    p(\Vec{\theta}_1, \dots, \Vec{\theta}_B) &\propto \exp \left( - \beta \mathcal{L}\left( \Vec{\theta}_1, \dots, \Vec{\theta}_B \right) \right) = \exp \left( - \beta \left( \sum_{\mu = 1}^n \ell_{\Vec{p}} (y_{\mu}, \Vec{\theta}_1^{\top} \Vec{x}_{\mu}, \dots, \Vec{\theta}_B^{\top} \Vec{x}_{\mu} ) + \sum_{b = 1}^B r( \Vec{\theta}_b ) \right) \right) 
    \label{eq:def:gibbs}
\end{align}
in the limit $\beta \to \infty$. Sampling the distribution on a graphical model can be used with Belief Propagation, which iterates messages between different nodes (here the coordinates $\Vec{\theta}_{ij}$ for $i \leq B, j \leq d$). However in high dimensions, Belief Propagation is intractable as it involves computing $d$-dimensional integrals. To alleviate this issue, GAMP only computes the first two moments of the different messages. In the high-dimensional limit, the output of GAMP coincides with the true minimizer of~\eqref{eq:def:weighted_erm}.

Similarly to our work, in \cite{Aubin2019Committee}, the authors introduce a GAMP algorithm for for a generic coupled system of estimates. They provide a detailed analysis of GAMP and its state evolution to track its behaviour in the asymptotic limit.

\subsection{State evolution equations}
In this section, we inspect the behavior of~\cref{alg:gamp_sample_weights} in the $n, d\to\infty$ limit and derive the asymptotic distribution of $\hat{\vec{\theta}}_1, \dots, \hat{\vec{\theta}}_d$. To do so, we start from the more convenient relaxed Belief Propagation (rBP) equations, which are very close to GAMP. In the high-dimensional limit, rBP and GAMP are equivalent. The rBP equations are written,

\begin{align}
    \begin{cases}
        \Vec{\omega}^{(t)}_{\mu \to i} &= \sum_{j \neq i} X_{\mu, j} \hat{\vec{\theta}}^{(t)}_{j \to \mu} \\
        \Vec{V}^{(t)}_{\mu \to i} &= \sum_{j \neq i} X_{\mu, j}^2 \hat{\mat{C}}^{(t)}_{j \to \mu}
    \end{cases},\quad
    \begin{cases}
        \channel{}_{\mu \to i}^{(t)} &= \channel(y_{\mu}, \Vec{\omega}_{\mu \to i}^{(t)}, \Vec{V}^{(t)}_{\mu \to i}, \Vec{p}_{\mu}) \\
        \partial\channel{}_{\mu \to i}^{(t)} &= \partial_{\vec{\omega}}\channel(y_{\mu}, \Vec{\omega}_{\mu \to i}^{(t)}, \Vec{V}^{(t)}_{\mu \to i}, \Vec{p}_{\mu})
    \end{cases}
\end{align} 

\begin{align}
    \begin{cases}
        \Vec{b}_{\mu \to i}^{(t)} &= \sum_{\nu \neq \mu} X_{\nu, i} \channel^{(t)}{}_{\nu \to i} \\
        \Vec{A}_{\mu \to i}^{(t)} &= - \sum_{\nu \neq \mu} X_{\nu, i}^2 \partial \channel^{(t)}{}_{\nu \to i} 
    \end{cases},\quad
    \begin{cases}
        \hat{\Vec{\theta}}^{(t)}_{i \to \mu} &= \Vec{f}_a(\Vec{b}^{(t)}_{i \to \mu}, \Vec{A}_{i \to \mu}^{(t)}) \\
        \hat{\mat{C}}^{(t)}_{i \to \mu} &= \partial_{\vec{b}}\Vec{f}_a(\Vec{b}^{(t)}_{i \to \mu}, \Vec{A}_{i \to \mu}^{(t)}).
    \end{cases}
\end{align}

It turns out that the average asymptotic behavior of these equations can be tracked with some overlap parameters defined as follows:
\begin{align}\label{eq:rbp_se_overlaps}
    \vec{m}^{(t)} &\equiv \lim_{d\to\infty}\frac1d\sum_{i=1}^d \hat{\vec{\theta}}^{(t)}_i\wstar^\top, \quad &\mat{Q}^{(t)} &\equiv \lim_{d\to\infty}\frac1d\sum_{i=1}^d \hat{\vec{\theta}}^{(t)}_i\hat{\vec{\theta}}^{(t)\top}_i\\
    \mat{V}^{(t)} &\equiv \lim_{d\to\infty}\frac1d\sum_{i=1}^d \hat{\mat{C}}^{(t)}_i, \quad &\rho&= \lim_{d\to\infty} \frac{\|\wstar\|^2}{d}.
\end{align}
To derive the asymptotic behavior of these overlap parameters, we compute the overlap distributions starting from the rBP equations above.

\subsubsection{Messages Distribution}
For convenience, let us define $z_{\mu} \equiv \sum_{i=1}^d X_{\mu, i}\theta_{\star i}=\vec{X}_\mu^\top\wstar$ and $z_{\mu\to i} \equiv \frac1d \sum_{j\neq i}X_{\mu, i}\theta_{\star j}$.

\paragraph{Distribution of $(z_{\mu}, \Vec{\omega}^{(t)}_{\mu \to i})$}
By the Central Limit Theorem, since $(z_{\mu}, \Vec{\omega}^{(t)}_{\mu \to i})$ are the sum of independent variables, they follow Gaussian distributions in the $d\to\infty$ limit. Therefore, we only need to compute their means, variances, and cross-correlation. Recall that from our assumptions, the random variables $X_{\mu, j}$ are i.i.d. zero-mean Gaussian with variance $\sfrac1d$. Hence, the first and second-order statistics read

\begin{align}
    \mathbb{E} \left[ z_{\mu} \right] &= \wstar^\top\mathbb{E}[\vec{X}_\mu] =  0 \\
    \mathbb{E} \left[ z_{\mu}^2 \right] &= \sum_{i, j=1}^d \mathbb{E}[X_{\mu, i}X_{\mu, j}]\theta_{\star i}\theta_{\star j} = \sum_{i, j=1}^d \frac1d\delta_{ij}\theta_{\star i}\theta_{\star j} =  \frac{\| \wstar \|^2}{d} \stackrel{d\to\infty}{\longrightarrow} \rho \\
    \mathbb{E} \left[ \Vec{\omega}^{(t)}_{\mu \to i} \right] &= \sum_{j \neq i} \mathbb{E}[X_{\mu, j}]\hat{\vec{\theta}}^{(t)}_{j \to \mu} = \Vec{0} \\
    \mathbb{E} \left[ \Vec{\omega}^{(t)}_{\mu \to i}(\Vec{\omega}^{(t)}_{\mu \to i})^\top \right] &= \sum_{j \neq i}^{d}\sum_{k \neq i}^{d}\mathbb{E}[X_{\mu, j}X_{\mu, k}]\hat{\Vec{\theta}}^{(t)}_{j \to \mu}\hat{\Vec{\theta}}^{(t) \top}_{k \to \mu} = \frac1d\sum_{j \neq i}^{d}\hat{\Vec{\theta}}^{(t)}_{j \to \mu}\hat{\Vec{\theta}}^{(t) \top}_{k \to \mu} \\
    &= \frac{1}{d} \sum_{j=1}^{d} \hat{\Vec{\theta}}^{(t)}_{j \to \mu} \hat{\Vec{\theta}}^{(t) \top}_{j \to \mu} - \frac1d\hat{\Vec{\theta}}^{(t)}_{i \to \mu} \hat{\Vec{\theta}}^{(t) \top}_{i \to \mu}  \stackrel{d\to\infty}{\longrightarrow} \mat{Q}^{(t)}\\
    \mathbb{E} \left[ z_{\mu} \Vec{\omega}^{(t)}_{\mu \to i} \right] &= \sum_{j=1}^d \sum_{k\neq i}^d\mathbb{E}[X_{\mu, j}X_{\mu, k}] \hat{\Vec{\theta}}^{(t)}_{k \to \mu} \wstar{}_j = \frac{1}{d} \sum_{j \neq i} \hat{\Vec{\theta}}^{(t)}_{j \to \mu} \wstar\\
    &= \frac{1}{d} \sum_{j=1}^d \hat{\Vec{\theta}}^{(t)}_{j \to \mu} \wstar - \frac{1}{d}\hat{\Vec{\theta}}^{(t)}_{i \to \mu} \wstar\stackrel{d\to\infty}{\longrightarrow} \Vec{m}^{(t)}
\end{align}

In summary, in the $d \to \infty$ limit : 
\begin{equation}\label{eq:joint_distribution_z_omega}
    \left( z_{\mu}, \Vec{\omega}^{(t)}_{\mu \to i} \right) \sim \mathcal{N}\left( 0, \begin{bmatrix}
        \rho & \Vec{m}^{(t) \top} \\
        \Vec{m}^{(t)} & \mat{Q}^{(t)}
    \end{bmatrix}
    \right)
\end{equation}

\paragraph{Concentration of $\Vec{V}^{(t)}_{\mu \to i}$}

In the asymptotic limit, the variances $\Vec{V}^{(t)}_{\mu \to i}$ concentrate around their means, which equates  
\begin{equation}
    \mathbb{E} \left[ \Vec{V}^{(t)}_{\mu \to i} \right] = \sum_{j \neq i}^d \mathbb{E} \left[ X_{\mu, j}^2 \right] \hat{\mat{C}}^{(t)} = \frac{1}{d} \sum_{j \neq i} \hat{\mat{C}}^{(t)}_j = \frac{1}{d} \sum_{j=1}^d \hat{\mat{C}}^{(t)}_j - \frac1d \hat{\mat{C}}^{(t)}_i \stackrel{d\to\infty}{\longrightarrow} \mat{V}^{(t)}
\end{equation}

\paragraph{Distribution of $\Vec{b}^{(t)}_{\mu \to i}$}
Recall from our setting that for a given input $\vec{x}_\mu$, the corresponding label is distributed as $y_\mu\sim p(\cdot|z_\mu)$. In fact, one can equivalently write $y^\mu=\varphi_0(z_\mu)$ for some (random) function $\varphi_0$. For example, the choice $\varphi_0(x)=x+\sqrt{\Delta}\xi$ corresponds to the linear regression, where $\xi\sim\mathcal{N}(0, 1)$ is Gaussian noise scaled by a variance $\Delta\geq 0$.
With this representation for $y_\mu$, we have
\begin{align}
    \Vec{b}^{(t)}_{\mu \to i} &= \sum_{\nu \neq \mu} X_{\nu, i} \channel( \varphi_0 \left( z_\nu \right) , \Vec{\omega}^{(t)}_{\nu \to i}, \Vec{V}^{(t)}_{\nu \to i}, \Vec{p}_{\nu}) \\
    &= \sum_{\nu \neq \mu} X_{\nu, i} \channel( \varphi_0 \left( z_{\nu \to i} + \theta_{\star i} X_{\nu, i} \right) , \Vec{\omega}^{(t)}_{\nu \to i}, \Vec{V}^{(t)}_{\nu \to i}, \Vec{p}_{\nu}) \\
    &= \sum_{\nu \neq \mu} X_{\nu, i} \channel( \varphi_0 \left( z_{\nu \to i} \right) , \Vec{\omega}^{(t)}_{\nu \to i}, \Vec{V}^{(t)}_{\nu \to i}, \Vec{p}_{\nu}) + X_{\nu, i}^2 \theta_{\star i} \partial_z \channel( \varphi_0 \left( z_{\nu \to i} \right) , \Vec{\omega}^{(t)}_{\nu \to i}, \Vec{V}^{(t)}_{\nu \to i}, \Vec{p}_{\nu}) + O(d^{-3/2}),
\end{align}
where in the last equality we have expanded the denoising function at leading order. Taking expectation on both sides yields
\begin{align}
    \mathbb{E}[\Vec{b}^{(t)}_{\mu \to i}] &= \frac{\theta_{\star i}}{d}  \sum_{\nu \neq \mu}\partial_z \channel( \varphi_0 \left( z_{\nu \to i} \right) , \Vec{\omega}^{(t)}_{\nu \to i}, \Vec{V}^{(t)}_{\nu \to i}, \Vec{p}_{\nu}) + O(d^{-3/2})\\
    &= \frac{\theta_{\star i}}{d}  \sum_{\nu =1}^n \partial_z \channel( \varphi_0 \left( z_{\nu \to i} \right) , \Vec{\omega}^{(t)}_{\nu \to i}, \Vec{V}^{(t)}_{\nu \to i}, \Vec{p}_{\nu}) - \frac{\theta_{\star i}}{d}\partial_z \channel( \varphi_0 \left( z_{\mu \to i} \right) , \Vec{\omega}^{(t)}_{\mu \to i}, \Vec{V}^{(t)}_{\mu \to i}, \Vec{p}_{\mu}) + O(d^{-3/2}),
\end{align}
Note that as $d\to\infty$, it follows from our computations above that for all $\nu$, $(z_{\nu\to i}, \vec{\omega}^{(t)}_{\nu\to i})$ are identically distributed according to~\cref{eq:joint_distribution_z_omega}. Consequently, by the Law of Large Numbers,
\begin{equation}
    \frac{n}{d}\cdot \frac1n\sum_{\nu =1}^n \partial_z \channel( \varphi_0 \left( z_{\nu \to i} \right) , \Vec{\omega}^{(t)}_{\nu \to i}, \Vec{V}^{(t)}_{\nu \to i}, \Vec{p}_{\nu}) \stackrel{n, d\to\infty}{\longrightarrow} \alpha\mean{(z, \omega), \vec{p}}{\partial_z \channel( \varphi_0 \left( z \right) , \Vec{\omega}, \Vec{V}^{(t)}, \Vec{p})} \equiv \hat{\vec{m}}^{(t)},
\end{equation}
from which we find that
\begin{equation}
    \mathbb{E}[\Vec{b}^{(t)}_{\mu \to i}] \stackrel{n, d\to\infty}{\longrightarrow} \theta_{\star i}\hat{\vec{m}}^{(t)}.
\end{equation}
The second moment can be computed in a similar fashion:
\begin{align}
    \mathbb{E}[\Vec{b}^{(t)}_{\mu \to i}\Vec{b}^{(t)\top}_{\mu \to i}] &= \sum_{\nu \neq \mu}\sum_{\kappa \neq \mu} \mathbb{E}[X_{\nu, i}X_{\kappa, i}] \channel( \varphi_0 \left( z_\nu \right) , \Vec{\omega}^{(t)}_{\nu \to i}, \Vec{V}^{(t)}_{\nu \to i}, \Vec{p}_{\nu})\channel( \varphi_0 \left( z_\kappa \right) , \Vec{\omega}^{(t)}_{\kappa \to i}, \Vec{V}^{(t)}_{\kappa \to i}, \Vec{p}_{\kappa})^\top\\
    &= \frac1d \sum_{\nu \neq \mu} \channel( \varphi_0 \left( z_{\nu \to i} \right) , \Vec{\omega}^{(t)}_{\nu \to i}, \Vec{V}^{(t)}_{\nu \to i}, \Vec{p}_{\nu}) \channel( \varphi_0 \left( z_{\nu \to i} \right) , \Vec{\omega}^{(t)}_{\nu \to i}, \Vec{V}^{(t)}_{\nu \to i}, \Vec{p}_{\nu})^\top + O(d^{-2})\\
    &= \frac1d \sum_{\nu = 1}^n \channel( \varphi_0 \left( z_{\nu \to i} \right) , \Vec{\omega}^{(t)}_{\nu \to i}, \Vec{V}^{(t)}_{\nu \to i}, \Vec{p}_{\nu}) \channel( \varphi_0 \left( z_{\nu \to i} \right) , \Vec{\omega}^{(t)}_{\nu \to i}, \Vec{V}^{(t)}_{\nu \to i}, \Vec{p}_{\nu})^\top + O(d^{-2})\\
    &\stackrel{n, d\to\infty}{\longrightarrow} \alpha\mean{(z, \vec{\omega}^{(t)}), \vec{p}}{\channel( \varphi_0 \left( z \right) , \Vec{\omega}^{(t)}, \Vec{V}^{(t)}, \Vec{p})\channel( \varphi_0 \left( z \right) , \Vec{\omega}^{(t)}, \Vec{V}^{(t)}, \Vec{p})^\top} \equiv \hat{\mat{Q}}^{(t)}.
\end{align}
Hence, $\Vec{b}^{(t)}_{\mu \to i} = \theta_{\star i}\hat{\vec{m}}^{(t)} + \left(\hat{\mat{Q}}^{(t)}\right)^{\sfrac12}\vec{\xi}$ with $\vec{\xi}\sim\mathcal{N}(\vec{0}, \mat{I}_{B})$.

\paragraph{Concentration of $\mat{A}^{(t)}_{\mu \to i}$}
It remains to show that the covariances $\mat{A}^{(t)}_{\mu \to i}$ concentrate. We have
\begin{align}
    \mat{A}^{(t)}_{\mu \to i} &= - \sum_{\nu \neq \mu} X_{\nu, i}^2 \partial_{\vec{\omega}} \channel(y_{\nu}, \Vec{\omega}_{\nu \to i}^{(t)}, \Vec{V}^{(t)}_{\nu \to i}, \Vec{p}_{\nu})\\
    &= - \sum_{\nu \neq \mu} X_{\nu, i}^2 \partial_{\vec{\omega}} \channel(\varphi_0(z_\nu), \Vec{\omega}_{\nu \to i}^{(t)}, \Vec{V}^{(t)}_{\nu \to i}, \Vec{p}_{\nu})\\
    &= - \sum_{\nu \neq \mu} X_{\nu, i}^2 \partial_{\vec{\omega}} \channel(\varphi_0(z_{\nu\to i}), \Vec{\omega}_{\nu \to i}^{(t)}, \Vec{V}^{(t)}_{\nu \to i}, \Vec{p}_{\nu}) + O(d^{-3/2}).
\end{align}
Taking the expectation gives
\begin{align}
    \mathbb{E}[\mat{A}^{(t)}_{\mu \to i}] &= -\frac1d \sum_{\nu \neq \mu} \partial_{\vec{\omega}} \channel(\varphi_0(z_{\nu\to i}), \Vec{\omega}_{\nu \to i}^{(t)}, \Vec{V}^{(t)}_{\nu \to i}, \Vec{p}_{\nu}) + O(d^{-3/2}) \\
    &= -\frac1d \sum_{\nu = 1}^n \partial_{\vec{\omega}} \channel(\varphi_0(z_{\nu\to i}), \Vec{\omega}_{\nu \to i}^{(t)}, \Vec{V}^{(t)}_{\nu \to i}, \Vec{p}_{\nu})-\frac1d\partial_{\vec{\omega}}\channel(\varphi_0(z_{\mu\to i}), \Vec{\omega}_{\mu \to i}^{(t)}, \Vec{V}^{(t)}_{\mu \to i}, \Vec{p}_{\mu}) + O(d^{-3/2}) \\
    &\stackrel{n, d\to\infty}{\longrightarrow} -\alpha\mean{(z, \vec{\omega}^{(t)}), \vec{p}}{\partial_{\vec{\omega}}\channel( \varphi_0 \left( z \right) , \Vec{\omega}^{(t)}, \Vec{V}^{(t)}, \Vec{p})} \equiv \hat{\mat{V}}^{(t)}
\end{align}

\subsubsection{Summary}
Having shown the distribution of messages and concentration, we are ready to characterize the asymptotic distribution of the estimator:
\begin{equation}
    \hatw_i\sim \vec{f}_a\left(\theta_{\star i}\hat{\vec{m}}^{(t)} + \left(\hat{\mat{Q}}^{(t)}\right)^{\sfrac12}\vec{\xi}, \hat{\mat{V}}^{(t)}\right)\quad \forall i\in\{1, \dots, d\},
\end{equation}
where $\vec{\xi}\sim\mathcal{N}(\vec{0}, \vec{I}_B)$.

From that, the definitions of overlaps in~\cref{eq:rbp_se_overlaps} at time $t+1$, and the message distributions, we obtain the state-evolution equations of the GAMP algorithm described in~\cref{alg:gamp_sample_weights}:
\begin{align}
    \begin{cases}
        \Vec{m}^{(t+1)} &= \mathbb{E}_{\theta_{\star}, \Vec{\xi}} \left[ \denoiser \left(\hat{\Vec{m}} \theta_{\star} + \sqrt{\hat{\mat{Q}}^{(t)}} \Vec{\xi}, \hat{\mat{V}}^{(t)}\right) \theta_{\star} \right] \\
        \mat{Q}^{(t+1)} &= \mathbb{E}_{\theta_{\star}, \Vec{\xi}} \left[ \denoiser\left(\hat{\Vec{m}} \theta_{\star} + \sqrt{\hat{\mat{Q}}^{(t)}} \Vec{\xi}, \hat{\mat{V}}^{(t)}\right) \denoiser\left(\hat{\Vec{m}} \theta_{\star} + \sqrt{\hat{\mat{Q}}^{(t)}} \Vec{\xi}, \hat{\mat{V}}^{(t)}\right)^\top \right] \\
        \mat{V}^{(t+1)} &= \mathbb{E}_{\theta_{\star}, \Vec{\xi}} \left[ \partial_{\vec{b}}\vec{f}_a\left(\hat{\Vec{m}} \theta_{\star} + \sqrt{\hat{\mat{Q}}^{(t)}} \Vec{\xi}, \hat{\mat{V}}^{(t)}\right)\right] \\
    \end{cases}
\end{align}
where $\vec{\xi}\sim\mathcal{N}(0, \mat{I}_B)$, and
\begin{align}
    \begin{cases}
        \hat{\Vec{m}}^{(t)} &= \alpha \mean{(z, \vec{\omega}^{(t)}), \vec{p}}{ \partial_z\channel(\varphi_0(z), \Vec{\omega}^{(t)}, \mat{V}^{(t)}, \vec{p})} \\
        \hat{\mat{Q}}^{(t)} &= \alpha \mean{(z, \vec{\omega}^{(t)}), \vec{p}}{ \channel(\varphi_0(z), \Vec{\omega}^{(t)}, \mat{V}^{(t)}, \vec{p}) \channel(\varphi_0(z), \Vec{\omega}^{(t)}, \mat{V}^{(t)}, \vec{p})^\top} \\
        \hat{\mat{V}}^{(t)} &= - \alpha \mean{(z, \vec{\omega}^{(t)}), \vec{p}}{ \partial_{\vec{\omega}} \channel(\varphi_0(z), \Vec{\omega}^{(t)}, \mat{V}^{(t)}, \vec{p})}
    \end{cases},
\end{align}
where $\left( z, \Vec{\omega}^{(t)} \right) \sim \mathcal{N}\left( 0, \begin{bmatrix}
        \rho & \Vec{m}^{(t) \top} \\
        \Vec{m}^{(t)} & \mat{Q}^{(t)}
    \end{bmatrix}
    \right)$.

Let us note that the overlaps $\hat{\Vec{m}}^{(t)}, \hat{\mat{Q}}^{(t)}, \hat{\mat{V}}^{(t)}$ can be written slightly differently. For that, first notice that since $\left( z, \Vec{\omega}^{(t)} \right)$ is Gaussian, so is $z$ conditioned on $\vec{\omega}^{(t)}$, and in particular $z|\vec{\omega}^{(t)}\sim\mathcal{N}(\mu_\star(\vec{\omega}^{(t)}), v_\star)$ with $\mu_{\star}(\Vec{\omega}^{(t)}) = (\Vec{m}^{(t)})^\top (\mat{Q}^{(t)})^{-1}\vec{\omega}^{(t)}$, $v_{\star} = \rho - (\Vec{m}^{(t)})^\top (\mat{Q}^{(t)})^{-1} \Vec{m}^{(t)}$. Moreover, using that $p(y|z)=\delta(y-\varphi_0(z))$, we have for an arbitrary function $\vec{f}:\reals\times\reals^B\to\reals^B$ that
\begin{align}
    \mean{(z, \vec{\omega}^{(t)})}{f(\varphi_0(z), \vec{\omega}^{(t)})} &= \mean{\vec{\omega}^{(t)}}{\mean{z|\vec{\omega}^{(t)}}{\vec{f}(\varphi_0(z), \vec{\omega}^{(t)})}} \\
    &= \mean{\vec{\omega}^{(t)}}{\int \dd z \mathcal{N}(z|\mu_\star(\vec{\omega}^{(t)}), v_\star)\vec{f}(\varphi_0(z), \vec{\omega}^{(t)})} \\
    &= \mean{\vec{\omega}^{(t)}}{\int \dd z \mathcal{N}(z|\mu_\star(\vec{\omega}^{(t)}), v_\star)\int \dd y p(y|z)\vec{f}(y, \vec{\omega}^{(t)})} \\
    &= \mean{\vec{\omega}^{(t)}}{\int \dd y \mathcal{Z}_0(y, \mu_\star(\vec{\omega}^{(t)}), v_\star)\vec{f}(y, \vec{\omega}^{(t)})},
\end{align}
where we have defined $\mathcal{Z}_0(y, \mu, v)\equiv \int \dd z \mathcal{N}(z|\mu, v) p(y|z)$. Consequently, we can rewrite
\begin{align}
    \begin{cases}
        \hat{\Vec{m}}^{(t)} &= \alpha \mean{\vec{\omega}^{(t)}, \vec{p}}{\int \dd y \partial_{\mu}\mathcal{Z}_0(y, \mu_\star(\vec{\omega}^{(t)}), v_\star)\cdot \channel(y, \vec{\omega}^{(t)}, \mat{V}^{(t)}, \vec{p})} \\
        \hat{\mat{Q}}^{(t)} &= \alpha \mean{\vec{\omega}^{(t)}, \vec{p}}{\int \dd y \mathcal{Z}_0(y, \mu_\star(\vec{\omega}^{(t)}), v_\star) \cdot \channel(y, \vec{\omega}^{(t)}, \mat{V}^{(t)}, \vec{p}) \channel(y, \vec{\omega}^{(t)}, \mat{V}^{(t)}, \vec{p})^\top} \\
        \hat{\mat{V}}^{(t)} &= - \alpha \mean{\vec{\omega}^{(t)}, \vec{p}}{\int \dd y \mathcal{Z}_0(y, \mu_\star(\vec{\omega}^{(t)}), v_\star)\cdot\partial_{\vec{\omega}} \channel(\varphi_0(z), \vec{\omega}^{(t)}, \mat{V}^{(t)}, \vec{p})}
    \end{cases},
\end{align}
where $\vec{\omega}^{(t)}\sim\mathcal{N}(\vec{0}, \mat{Q}^{(t)})$.

\subsubsection{Self-Consistent Equations}
In the limit $t\to\infty$, the state-evolution equations derived above yield a set of self-consistent equations:

\begin{align}
    \begin{cases}
        \Vec{m} &= \mathbb{E}_{\theta_{\star}, \Vec{\xi}} \left[ \denoiser (\hat{\Vec{m}} \theta_{\star} + \sqrt{\hat{\mat{Q}}} \Vec{\xi}, \hat{\mat{V}}) \theta_{\star} \right] \\
        \mat{Q} &= \mathbb{E}_{\theta_{\star}, \Vec{\xi}} \left[\left[\denoiser\denoiser^\top\right](\hat{\Vec{m}} \theta_{\star} + \sqrt{\hat{\mat{Q}}} \Vec{\xi}, \hat{\mat{V}}) \right] \\
        \mat{V} &= \mathbb{E}_{\theta_{\star}, \Vec{\xi}} \left[ \partial_{\vec{b}}\vec{f}_a(\hat{\Vec{m}} \theta_{\star} + \sqrt{\hat{\mat{Q}}} \Vec{\xi}, \hat{\mat{V}})\right] \\
    \end{cases}
    ,
    \begin{cases}
        \hat{\Vec{m}} &= \alpha \mean{\vec{\omega}, \vec{p}}{\int \dd y \partial_{\mu} \mathcal{Z}_0(y, \mu_{\star}(\Vec{\omega}), v_{\star}) \cdot \channel(y, \Vec{\omega}, \mat{V}, \vec{p})} \\
        \hat{\mat{Q}} &= \alpha \mean{\vec{\omega}, \vec{p}}{\int \dd y \mathcal{Z}_0(y, \mu_{\star}(\Vec{\omega}), v_{\star}) \cdot \left[ \channel \channel^\top \right](y, \Vec{\omega}, \mat{V}, \vec{p})} \\
        \hat{\mat{V}} &= - \alpha \mean{\vec{\omega}, \vec{p}}{\int \dd y \mathcal{Z}_0(y, \mu_{\star}(\Vec{\omega}), v_{\star}) \cdot \partial_{\vec{\omega}} \channel(y, \Vec{\omega}, \mat{V}, \vec{p})} \\
    \end{cases}
    \label{eq:se_gamp_bootstrap}
\end{align}

where $\vec{\xi}\sim\mathcal{N}(0, \mat{I}_B)$, $\vec{\omega}\sim\mathcal{N}(\vec{0}, \mat{Q})$, and $\mu_{\star}(\Vec{\omega}) = \Vec{m}^\top \mat{Q}^{-1}\vec{\omega}$ and $v_{\star} = \rho - \Vec{m}^\top \mat{Q}^{-1} \Vec{m}$ with $\rho=\sfrac1d\|\wstar\|_2^2$.

\subsubsection{Channels}

\paragraph{Channel for square loss} When the loss is the square loss $\ell(y, \omega) = \frac{1}{2 \Delta} (y - \omega)^2$, we can conveniently write the proximal in a matrix form 
\begin{equation}
    {\rm prox}(y, \Vec{\omega}, \mat{V}, \Vec{p}) = \arg\min_{\vec z\in\reals^B} \frac{1}{2}(\Vec{z} - \Vec{\omega})^\top \mat{V}^{-1} (\Vec{z} - \Vec{\omega}) + \frac{1}{2 \Delta}(\vec{z} - \Vec{1}_B y)^\top \mat{P} (\vec{z} - \Vec{1}_B y),
    \label{eq:matrix_proximal}
\end{equation}
where we have defined $\mat{P}=\mathrm{Diag}(\Vec{p})$. In that case, the vector $\Vec{z}$ that cancels the derivative of the function to minimize is 
\begin{equation}
    \Vec{z}_* = \left( \mat{V}^{-1} + \frac{\mat{P}}{\Delta} \right)^{-1} \left( \mat{V}^{-1} \Vec{\omega} + \frac{\mat{P}}{\Delta} \Vec{1}_B y \right) 
\end{equation}
such that 
\begin{align}
    \channel(y, \Vec{\omega}, \mat{V}, \Vec{p}) &= \left( \mat{I}_B + \frac{\mat{PV}}{\Delta} \right)^{-1} \frac{\mat{P}}{\Delta} (\Vec{1}_B y - \vec{\omega}) \\
    \partial_{\vec{\omega}} \channel(y, \Vec{\omega}, \mat{V}, \Vec{p}) &= - \left( \mat{I}_B + \frac{\mat{PV}}{\Delta} \right)^{-1} \frac{\mat{P}}{\Delta}
\end{align}

\paragraph{Channel for logistic loss} In classification tasks one usually uses the logistic loss $\ell(y, z) = \log \left( 1 + e^{-z} \right)$. We thus aim to compute the proximal 
\begin{equation}
    \prox_{\ell(y, \cdot), \mat{V}}(\Vec{\omega}) = \arg\min_{\vec{z}\in\reals^B} \sum_{b=1}^B p_b \ell(y, z_b) + \frac{1}{2} (\Vec{z} - \Vec{\omega}) \mat{V}^{-1} (\Vec{z} - \Vec{\omega})
\end{equation}
We deduce the channel from it. On the other hand, to compute $\partial_{\Vec{\omega}} \channel$, one needs to compute the Hessian of the loss function: 
\begin{align}
    \nabla^2 \ell(y, \Vec{z}, \Vec{p}) = {\rm Diag} \left( p_1 \sigma'(y z_1), \dots, p_B \sigma'(y z_B) \right)
\end{align}

\subsubsection{Denoiser for $\ell_2$ regularization}
In a similar way, the denoiser is written 
\begin{align}
    \vec{f}_a(\vec{b}, \mat{A}) &= \left( \lambda \mat{I}_B + \mat{A} \right)^{-1} \vec{b} \\
    \partial_{\vec{b}}f_a(\vec{b}, \mat{A}) &= \left( \lambda \mat{I}_B + \mat{A} \right)^{-1}
\end{align}

\subsection{Ridge regression}\label{appendix:gamp_ridge_regression}
Using the channel for square loss and the denoiser for $\ell_2$ regularization, we can compute the various overlaps for the ridge regression. First, defining $\mat{R}(\lambda)\equiv(\lambda \mat{I}_B+\hat{\mat{V}})^{-1}$, we find that
\begin{align}
    \vec{m} &= \mean{\theta_\star, \vec{\xi}}{\mat{R}(\lambda)\left(\hat{\vec{m}}\theta_\star+\sqrt{\hat{\mat{Q}}}\vec{\xi}\right)\theta_\star} = \mat{R}(\lambda)\hat{\vec{m}}\mean{\theta_\star}{\theta_\star^2}= \mat{R}(\lambda)\hat{\vec{m}}\rho\\
    \mat{Q} &= \mean{\theta_\star, \vec{\xi}}{\mat{R}(\lambda)\left(\hat{\vec{m}}\theta_\star+\sqrt{\hat{\mat{Q}}}\vec{\xi}\right)\left(\hat{\vec{m}}\theta_\star+\sqrt{\hat{\mat{Q}}}\vec{\xi}\right)^\top\mat{R}(\lambda)^\top} = \mat{R}(\lambda) \left( \rho\hat{\vec{m}} \hat{\vec{m}}^\top + \hat{\mat{Q}} \right) \mat{R}(\lambda)^{\top}\\
    \mat{V} &= \mean{\theta_\star, \vec{\xi}}{\mat{R}(\lambda)} = \mat{R}(\lambda).
\end{align}
In order to compute the other overlaps, we must first evaluate $\mathcal{Z}_0(y, \mu, v)\equiv \int \dd z \mathcal{N}(z|\mu, v) p(y|z)$. Since $p(y|z)=\mathcal{N}(y|z, \Delta)$ for ridge regression, $\mathcal{Z}_0(y, \mu, v)$ is simply the convolution of $\mathcal{N}(y|0, \Delta)$ and $\mathcal{N}(y|\mu, v)$, from which we can conclude $\mathcal{Z}_0(y, \mu, v)$ is equal to the density of $\mathcal{N}(0, \Delta) + \mathcal{N}(\mu, v) = \mathcal{N}(\mu_\star(\vec{\omega}), v_\star +\Delta)$. Hence, $\mathcal{Z}_0(y, \mu, v)=\mathcal{N}(y|\mu, v +\Delta)$, and we also find that $\partial_\mu \mathcal{Z}_0(y, \mu, v) = \frac{y-\mu}{v+\Delta}\mathcal{N}(y|\mu, v +\Delta)$. Defining $\mat{G}(\Vec{p}) \equiv (\mat{I}_2 + \mat{P V})^{-1} \mat{P}$ with $\mat{P} = \mathrm{Diag}(\Vec{p})$, the overlaps are given by
\begin{align}
    \hat{\vec{m}} &= \alpha\mean{\vec{\omega, \vec{p}}}{\int\dd y \mathcal{N}(y|\mu_\star(\vec{\omega}), v_\star +\Delta)\frac{y-\mu_\star(\vec{\omega})}{v_\star+\Delta}G(\vec{p})(\vec{1}_By-\vec{\omega})} \\
    &= \alpha\mean{\vec{p}}{G(\vec{p})}\mean{\vec{\omega}}{\int\dd y \mathcal{N}(y|\mu_\star(\vec{\omega}), v_\star +\Delta)\left(\vec{1}_B\frac{y^2}{v_\star+\Delta}-\vec{1}_B\frac{y\mu_\star(\vec{\omega})}{v_\star+\Delta}-\frac{y-\mu_\star(\vec{\omega})}{v_\star+\Delta}\vec{\omega}\right)}\\
    &= \alpha\mean{\vec{p}}{G(\vec{p})}\mean{\vec{\omega}}{\left(\vec{1}_B\frac{v_\star+\Delta + \mu_\star(\vec{\omega})^2}{v_\star+\Delta}-\vec{1}_B\frac{\mu_\star(\vec{\omega})^2}{v_\star+\Delta}\right)}\\
    &= \alpha\mean{\vec{p}}{G(\vec{p})}\vec{1}_B\\
    \hat{\mat{Q}} &= \alpha\mean{\vec{\omega, \vec{p}}}{\int\dd y \mathcal{N}(y|\mu_\star(\vec{\omega}), v_\star +\Delta)G(\vec{p})(\vec{1}_By-\vec{\omega})(\vec{1}_By-\vec{\omega})^\top G(\vec{p})^\top}\\
    &= \alpha\mean{\vec{p}}{G(\vec{p})\mean{\vec{\omega}}{\mat{1}_{B\times B}(v_\star+\Delta+\mu_\star(\vec{\omega})^2)-\vec{1}_B \mu_\star(\vec{\omega})\vec{\omega}^\top-\vec{\omega}\vec{1}_B^\top\mu_\star(\vec{\omega})+\vec{\omega}\vec{\omega}^\top}G(\vec{p})^\top}\\
    &= \alpha\mean{\vec{p}}{G(\vec{p})\left(\mat{1}_{B\times B}(v_\star+\Delta+\vec{m}^\top Q^{-1}\vec{m})-\vec{m}\vec{1}_B^\top-\vec{1}_B\vec{m}^\top+\mat{Q}\right)G(\vec{p})^\top}\\
    &= \alpha\mean{\vec{p}}{G(\vec{p})\left(\mat{1}_{B\times B}(v_\star+\Delta)+\mat{B}\mat{Q}\mat{B}^\top\right)G(\vec{p})^\top}\label{eq:definition_matrix_b}\\
    \hat{\mat{V}} &=  -\alpha\mean{\vec{\omega, \vec{p}}}{\int\dd y \mathcal{N}(y|\mu_\star(\vec{\omega}), v_\star +\Delta)(-G(\vec{p}))} = \alpha\mean{\vec{p}}{G(\vec{p})},
\end{align}
where $\mat{B} = \vec{1}_B\vec{m}^\top\mat{Q}^{-1}-\mat{I}_B$ in~\cref{eq:definition_matrix_b}.

\subsubsection{Summary}
Overall, the closed-form expressions for the state-evolution for ridge regression are
\begin{align}\label{eq:system_equations_ridge}
    \begin{cases}
        \hat{\vec{m}} &= \alpha \mathbb{E}_{\Vec{p}} \left[ \mat{G}(\Vec{p}) \right] \mathbf{1}_B \\
        \hat{\mat{Q}}       &= \alpha \mathbb{E}_{\Vec{p}} \left[ \mat{G}(\Vec{p}) \left( \left(v_{\star} + \Delta \right) \mathbf{1}_{B \times B} + \mat{B Q B}^{\top} \right) \mat{G}(\Vec{p})^{\top} \right]\\
        \hat{\mat{V}}       &= \alpha \mathbb{E}_{\Vec{p}} \left[ \mat{G}(\Vec{p}) \right]\\
    \end{cases}, 
    \begin{cases}
        \vec{m}       &= \rho\mat{R}(\lambda) \hat{\vec{m}}\\
        \mat{Q}             &= \mat{R}(\lambda) \left( \rho\hat{\vec{m}} \hat{\vec{m}}^\top + \hat{\mat{Q}} \right) \mat{R}(\lambda)^{\top}\\
        \mat{V}             &= \mat{R}(\lambda) \\
    \end{cases}
\end{align}
with $\mat{G}(\Vec{p}) = (\mat{I}_B + \mat{P V})^{-1} \mat{P}, \mat{P} = \mathrm{Diag}(\Vec{p})$, $\mat{B} = \vec{1}_B\vec{m}^\top \mat{Q}^{-1} - \mat{I}_B$, and $\mat{R}(\lambda) = \left( \lambda \mat{I}_B + \hat{\mat{V}} \right)^{-1}$, and $v_{\star} = \rho - \Vec{m}^\top \mat{Q}^{-1} \Vec{m}$.

\clearpage

\section{Derivation of the results for residual resampling}\label{appendix:residual_resampling}
As for pair resampling, one can consider the state-evolution equations of a well-chosen AMP algorithm to compute the conditional bias / variance and the bias and variance of residual bootstrap. 
Indeed, as for pair resampling, we leverage the fact that the conditional bias and variance, together with the estimates by residual bootstrap, can be written in terms of correlations between estimators trained on different resampled datasets $\dataset^{\star}_b$ with same covariates $\mat{X}$ but resampled labels $y^{\star}$. Introducing an augmented dataset $\tilde{\dataset} = (\Vec{x}_i, \Vec{y}^{\star}_i = (y^{\star}_{b, i})_{b = 1}^B)_{i = 1}^n$ where the labels are now $B$-dimensional vectors comprised of the resampled labels, we see that \cref{eq:def_erm_residual} is mathematically equivalent to the following minimization problem
\begin{equation}
    (\hatw)_{b = 1}^B = \arg\min_{\Vec{\theta}_1, \dots, \Vec{\theta}_B\in\reals^d} \sum_{b=1}^B \sum_{i = 1}^n - \log p(y^{\star}_{b,i} | \Vec{\theta}_b^{\top} \Vec{x}_i) + \frac{\lambda}{2} \| \Vec{\theta}_b \|^2  
    \label{eq:erm_residual_joint}
\end{equation}
While \cref{eq:erm_residual_joint} is equivalent \cref{eq:def_erm_residual}, formulating it as a joint minimization over $B$ estimators allow us to solve it using a specific AMP algorithm. As for pair resampling, the state-evolution equations of AMP will yield the correlation between two estimators $\mathbb{E}_{\dataset^{\star}_b, \dataset^{\star}_{b'}} \left[ \hatw( \dataset^{\star}_b )^\top \hatw( \dataset^{\star}_{b'} ) \right]$ in the high-dimensional limit. These correlations are sufficient to compute the true variance and its estimation with the residual bootstrap, depending on the resampling process $\dataset^{\star}$.

For residual bootstrap, the AMP algorithm is similar to \cref{alg:gamp_sample_weights} to compute the estimators $\hatw_i$. The main difference with \cref{alg:gamp_sample_weights} is the absence of sample weights $p_i$, as all the covariates $\Vec{x}_i$ are resampled only once. Equivalently, we can consider constant sample weights $p_i = 1 \;\forall i$. Moreover, the labels are now $B$-dimensional.

The overlaps can be computed using the state evolution equations~\eqref{eq:se_gamp_bootstrap} of \cref{alg:gamp_sample_weights}, where the 2-dimensional channel function is 
\begin{equation}
    \channel(\Vec{y}, \Vec{\omega}, \mat{V}) = \arg\min_{\vec{z}\in\reals^B} \frac{1}{2} (\Vec{z} - \Vec{\omega})^\top \mat{V}^{-1} (\Vec{z} - \Vec{\omega}) + \sum_{b=1}^B\ell(y_b, z_b)
\end{equation}
Note that here the channel function takes a vector label as input instead of scalar label. Moreover, the channel function does not depend on any sample weight $\vec{p}$. This yields the following equations:

\begin{align}
    \begin{cases}
        \Vec{m} &= \mathbb{E}_{\theta_{\star}, \Vec{\xi}} \left[ \denoiser (\hat{\Vec{m}} \theta_{\star} + \sqrt{\hat{\mat{Q}}} \Vec{\xi}, \hat{\mat{V}}) \theta_{\star} \right] \\
        \mat{Q} &= \mathbb{E}_{\theta_{\star}, \Vec{\xi}} \left[ \denoiser(\hat{\Vec{m}} \theta_{\star} + \sqrt{\hat{\mat{Q}}} \Vec{\xi}, \hat{\mat{V}}) \denoiser(\hat{\Vec{m}} \theta_{\star} + \sqrt{\hat{\mat{Q}}} \Vec{\xi}, \hat{\mat{V}})^\top \right] \\
        \mat{V} &= \mathbb{E}_{\theta_{\star}, \Vec{\xi}} \left[ \partial_{\vec{b}}\vec{f}_a(\hat{\Vec{m}} \theta_{\star} + \sqrt{\hat{\mat{Q}}} \Vec{\xi}, \hat{\mat{V}})\right] \\
    \end{cases}
    \label{eq:se_y_resampling_overlaps}
\end{align}
with $\vec{\xi}\sim\mathcal{N}(\Vec{0}, \mat{I}_B)$ and
\begin{align}
    \begin{cases}
        \hat{\Vec{m}} &= \alpha \mean{\Vec{\omega}}{\int \dd \Vec{y} \partial_{\mu} \mathcal{Z}_0(\Vec{y}, \mu_{\star}(\Vec{\omega}), v_{\star}) \cdot \channel(\Vec{y}, \Vec{\omega}, \mat{V})} \\
        \hat{\mat{Q}} &= \alpha \mean{\Vec{\omega}}{\int \dd \Vec{y} \mathcal{Z}_0(\Vec{y}, \mu_{\star}(\Vec{\omega}), v_{\star}) \cdot \channel(\vec{y}, \Vec{\omega}, \mat{V}) \channel(\vec{y}, \Vec{\omega}, \mat{V})^\top} \\
        \hat{\mat{V}} &= - \alpha \mean{\Vec{\omega}}{\int \dd \Vec{y} \mathcal{Z}_0(\Vec{y}, \mu_{\star}(\Vec{\omega}), v_{\star}) \cdot \partial_{\omega} \channel(\vec{y}, \Vec{\omega}, \mat{V})}
    \end{cases},
    \label{eq:se_y_resampling_hat_overlaps}
\end{align}
where $\vec{\omega}\sim\mathcal{N}(\Vec{0}, \mat{Q})$. Now the integrals in~\cref{eq:se_y_resampling_hat_overlaps} carry over vector labels $\Vec{y}$ and the teacher partition $\mathcal{Z}_0$ is
\begin{equation}
    \mathcal{Z}_0(\Vec{y}, \mu, v) = \int \dd z \mathcal{N}(z | \mu, v)\prod_{i=1}^B p(y_i | z)
\end{equation}

In Equations~\eqref{eq:se_y_resampling_overlaps} and \eqref{eq:se_y_resampling_hat_overlaps}, $\rho$ is the squared norm $\sfrac{1}{d} \| \wstar \|^2$ of the label-generating vector $\wstar$. In the case of conditional resampling, $\wstar = 1$ as for pair resampling. However, in the case of residual bootstrap, $\wstar$ is replaced by the ERM estimator $\werm$, and $\rho = \sfrac{1}{d} \| \werm \|^2$. In the high-dimensional limit, $\sfrac{1}{d} \| \werm \|^2$ is obtained by running the equations~\eqref{eq:se_gamp_bootstrap} for full resampling, and we have $\rho = Q_{11}^{\fr}$. 

\paragraph{Ridge regression}\label{appendix:gamp_ridge_regression_resampling_labels}
In the Ridge regression case, the state-evolution equations are given by
\begin{align}\label{eq:system_equations_residual_ridge}
    \begin{cases}
        \hat{\vec{m}} &= \alpha \mat{G} \mathbf{1}_B \\
        \hat{\mat{Q}} &= \alpha \mat{G} \left( v_{\star} \mathbf{1}_{B \times B} + \Delta \mat{I}_B + \mat{B Q B}^{\top} \right) \mat{G}^{\top} \\
        \hat{\mat{V}} &= \alpha  \mat{G}
    \end{cases}, 
    \begin{cases}
        \vec{m}             &= \rho \mat{R}(\lambda) \hat{\vec{m}}\\
        \mat{Q}             &= \mat{R}(\lambda) \left(\rho \hat{\vec{m}} \hat{\vec{m}}^\top + \hat{\mat{Q}} \right) \mat{R}(\lambda)^{\top}\\
        \mat{V}             &= \mat{R}(\lambda)
    \end{cases}
\end{align}
with $\mat{G} = (\mat{I}_B + \mat{V})^{-1}$, $\mat{B} = \vec{1}_B\vec{m}^\top \mat{Q}^{-1} - \mat{I}_B$, and $\mat{R}(\lambda) = \left( \lambda \mat{I}_B + \hat{\mat{V}} \right)^{-1}$, and $v_{\star} = \rho - \Vec{m}^\top \mat{Q}^{-1} \Vec{m}$. Note that $\Delta$ is the variance of the Gaussian noise, which will be $1$ for conditional resampling but not for residual bootstrap. 

\subsection{Residual bootstrap}\label{appendix:residual_bootstrap}

In residual bootstrap, one uses the ERM estimator trained on the whole dataset $\dataset$ to sample new labels with fixed input data $X$. Then, to compute the asymptotic behaviour of residual bootstrap, the idea is to solve Equations~\eqref{eq:se_y_resampling_overlaps} and \eqref{eq:se_y_resampling_hat_overlaps} where $\wstar$ is replaced by $\werm$. Its squared norm $\|\wstar\|_2^2$ will be replaced by $\| \werm \|^2$ and, in the case of ridge regression, the noise variance is generally replaced by the training square-loss
\begin{equation}
    \hat{\Delta} = \frac{1}{n} \sum_{i = 1}^n \left(y_{i} - \werm^{\top} \Vec{x}_i\right)^2
\end{equation}
Note that $\hat{\Delta}$ will typically underestimate $\Delta$ as $\werm$ is correlated to $\Vec{x}_i$. In practice, to compute the asymptotics of residual bootstrap, we first run the state-evolution equations to compute the (scalar) overlaps $\Vec{m}^{\fr}, \mat{Q}^{\fr}, \vec{V}^{\fr}$ for the ERM estimator. We then plug these overlaps in Equations~\eqref{eq:se_y_resampling_overlaps} and \eqref{eq:se_y_resampling_hat_overlaps}, yielding new update equations for $\hat{\Vec{m}}, \hat{\mat{Q}}, \hat{\mat{V}}$: 
\begin{align}
    \begin{cases}
        \hat{\Vec{m}} &= \alpha \mean{\vec{\omega}}{\int \dd \Vec{y} \partial_{\omega} \mathcal{Z}_0(\Vec{y}, \mu_{\star}(\Vec{\omega}), \Tilde{v}_{\star}) \cdot \channel(\Vec{y}, \Vec{\omega}, \mat{V}) )} \\
        \hat{\mat{Q}} &= \alpha \mean{\vec{\omega}}{\int \dd \Vec{y} \mathcal{Z}_0(\Vec{y}, \mu_{\star}(\Vec{\omega}), \Tilde{v}_{\star}) \cdot \channel(y, \Vec{\omega}, \mat{V}) \channel(y, \Vec{\omega}, \mat{V})^\top} \\
        \hat{\mat{V}} &= - \alpha \mean{\vec{\omega}}{\int \dd \Vec{y}\mathcal{Z}_0(\Vec{y}, \mu_{\star}(\Vec{\omega}), \Tilde{v}_{\star}) \cdot \partial_{\vec{\omega}} \channel(y, \Vec{\omega}, \mat{V})}
    \end{cases},
    \label{eq:se_residual_bootstrap_hat_overlaps}
\end{align}
where $\vec{\omega}\sim\mathcal{N}(\Vec{0}, \mat{Q})$.
Also note that here, $\Tilde{v}_{\star} = Q_{11}^{\fr} - \Vec{m}^\top \mat{Q}^{-1} \Vec{m}$ as we replaced $\rho$ by $Q_{11}^{\fr}$, and for ridge regression, 
\begin{equation}
    \mathcal{Z}_0(y, \mu, v) = \int \dd z \mathcal{N}(y | z, \tilde{\Delta}) \mathcal{N}(z |\mu, v) = \mathcal{N}(y | \mu, \tilde{\Delta} + v)
\end{equation}
wherein high-dimensions, the $\ell_2$ loss of $\werm$ on the training set $\dataset$ is $\Tilde{\Delta} = \frac{1 + \Delta - 2 m_1^{\fr} + Q_{11}^{\fr}}{(1 + V_1^{\fr})^2}$, see \citep{loureiro2021learning} for a proof.

\clearpage

\section{Overlaps and Rates in Ridge Regression}
\label{appendix:overlaps_rates}
This section is devoted to the simplification of the system of equations in~\cref{eq:system_equations_ridge}. Indeed, while the GAMP algorithm can be run with general $B \geq 1$, we can in fact restrict ourselves to the case $B = 2$ without loss of generality. Since our main goal is to compute the correlation between various independent bootstrap resamples and the resamples are i.i.d, the overlaps will have a simple structure that does not depend on $B$. Once analytical expressions for the overlaps of interest are obtained, the rates of various quantitie like bias and variance are computed in the regime $\alpha\to\infty$.

\subsection{Solution to the State-Evolution Equations}
Let us simplify the system of equations in~\cref{eq:system_equations_ridge} assuming $B=2$:

\paragraph{Overlaps $\mat{V}, \hat{\mat{V}}$} Note that the matrices $\mat{V}$ and $\hat{\mat{V}}$ are diagonal, so that we can denote them as $\mat{V}=\mathrm{Diag}(v_1, v_2)$ and $\hat{\mat{V}}=\mathrm{Diag}(\hat{v}_1, \hat{v}_2)$. This is due to the fact that the two estimators are independently computed. As such, combining the two equations for $\mat{V}$ and $\hat{\mat{V}}$ in~\cref{eq:system_equations_ridge}, one can write
\begin{align}
    \begin{bmatrix}
        v_1 & 0\\
        0 & v_2
    \end{bmatrix} = \begin{bmatrix}
        \frac{1}{\lambda + \alpha\mean{p_1}{\frac{p_1}{1+p_1v_1}}} & 0\\
        0 & \frac{1}{\lambda + \alpha\mean{p_2}{\frac{p_2}{1+p_2v_2}}}
    \end{bmatrix}.
\end{align}
Hence for $i=1, 2$, the overlap $v_i$ is given by the fixed-point equation
\begin{equation}
    v_i = \frac{1}{\lambda + \alpha\mean{p_i}{\frac{p_i}{1+p_iv_i}}}.
\end{equation}
Moreover, we have $\hat{v}_i = \alpha\mean{p_i}{\frac{p_i}{1+p_iv_i}}= \frac{1}{v_i}-\lambda$.

\paragraph{Overlaps $\vec{m}, \hat{\vec{m}}$} Next, we deduce $\vec{m}$ by combining the $\vec{m}$ and $\hat{\vec{m}}$ expressions from~\cref{eq:system_equations_ridge}:
\begin{align}
    \begin{bmatrix}
        m_{1}\\
        m_{2}
    \end{bmatrix} = \alpha\begin{bmatrix}
        \frac{\rho}{\lambda+\hat{v}_{1}}\mean{p_1}{\frac{p_1}{1+p_1v_{1}}}\\
        \frac{\rho}{\lambda+\hat{v}_{2}}\mean{p_2}{\frac{p_2}{1+p_2v_{2}}}
    \end{bmatrix} = \begin{bmatrix}
        \frac{\rho\hat{v}_{1}}{\lambda+\hat{v}_{1}}\\
        \frac{\rho\hat{v}_{2}}{\lambda+\hat{v}_{2}}
    \end{bmatrix},
\end{align}
so that $m_i=\frac{\rho\hat{v}_i}{\lambda+\hat{v}_i}=\rho(1-\lambda v_i)$, for $i=1,2$. Moreover, $\hat{m}_i=\hat{v}_i$.

\paragraph{Overlaps $\mat{Q}, \hat{\mat{Q}}$} One can leverage the fact that the matrices $\mat{Q}, \hat{\mat{Q}}$ are symmetric. Using the notation
\begin{align}
    \mat{Q}:= \begin{bmatrix}
        q_1 & q_{1,2}\\
        q_{1,2} & q_2
    \end{bmatrix},\quad \hat{\mat{Q}}:=\begin{bmatrix}
        \hat{q}_1 & \hat{q}_{1,2}\\
        \hat{q}_{1,2} & \hat{q}_2
    \end{bmatrix} \quad\text{ and }\quad \mat{Q}^{-1} := \begin{bmatrix}
        q_1^\prime & q_{1,2}^\prime\\
        q_{1,2}^\prime & q_2^\prime
    \end{bmatrix}
\end{align}
one can rewrite the equation for $\mat{Q}$ from~\cref{eq:system_equations_ridge} as
\begin{align}
    \begin{bmatrix}
        q_1 & q_{1,2}\\
        q_{1,2} & q_2
    \end{bmatrix} = \begin{bmatrix}
        \frac{\rho\hat{m}_1^2+\hat{q}_1}{(\lambda+\hat{v}_1)^2} & \frac{\rho\hat{m}_1\hat{m}_2+\hat{q}_{1,2}}{(\lambda+\hat{v}_1)(\lambda+\hat{v}_2)}\\
        \frac{\rho\hat{m}_1\hat{m}_2+\hat{q}_{1,2}}{(\lambda+\hat{v}_1)(\lambda+\hat{v}_2)} & \frac{\rho\hat{m}_2^2+\hat{q}_2}{(\lambda+\hat{v}_2)^2}
    \end{bmatrix}\iff \begin{cases}
        q_i = \frac{\rho\hat{m}_i^2+\hat{q}_i}{(\lambda+\hat{v}_i)^2} = \frac{1}{\rho} m_i^2 + v_i^2\hat{q}_i,\quad\text{for $i=1,2$}\\
        q_{1,2} = \frac{\rho\hat{m}_1\hat{m}_2+\hat{q}_{1,2}}{(\lambda+\hat{v}_1)(\lambda+\hat{v}_2)} = \frac{1}{\rho} m_1m_2 + v_1v_2\hat{q}_{1,2}
    \end{cases}.
\end{align}
The computations are slightly more involved for $\hat{\mat{Q}}$, but one can derive that
\begin{align}
    \mat{BQB}^\top = (m_1^2q_1^\prime + 2m_1m_2q_{1,2}^\prime+ m_2^2q_2^\prime)\mathbf{1}_2 + Q - \begin{bmatrix}
        \vec{m}^\top\\
        \vec{m}^\top
    \end{bmatrix}-\begin{bmatrix}
        \vec{m} & \vec{m}
    \end{bmatrix} \quad\text{ and }\quad v_{\star} = \rho-(m_1^2q_1^\prime + 2m_1m_2q_{1,2}^\prime+ m_2^2q_2^\prime),
\end{align}
and consequently the equation for $\hat{\mat{Q}}$ from~\cref{eq:system_equations_ridge} reads
\begin{align}
    \begin{bmatrix}
        \hat{q}_1 & \hat{q}_{1,2}\\
        \hat{q}_{1,2} & \hat{q}_2
    \end{bmatrix} &= \alpha \begin{bmatrix}
        \mean{p_1}{(\frac{p_1}{1+p_1v_1})^2}(\rho+\Delta-2m_1+q_1) & \mean{p_1, p_2}{\frac{p_1}{1+p_1v_1}\cdot\frac{p_2}{1+p_2v_2}}(\rho+\Delta-m_1-m_2+q_{1,2})\\
        \mean{p_1, p_2}{\frac{p_1}{1+p_1v_1}\cdot\frac{p_2}{1+p_2v_2}}(\rho+\Delta-m_1-m_2+q_{1,2}) & \mean{p_2}{(\frac{p_2}{1+p_2v_2})^2}(\rho+\Delta-2m_2+q_2)
    \end{bmatrix}\\
    &\iff \begin{cases}
        \hat{q}_i = \alpha\mean{p_i}{\left(\frac{p_i}{1+p_iv_i}\right)^2}(\rho+\Delta-2m_i+q_i),\quad\text{for $i=1,2$}\\
        \hat{q}_{1,2} = \alpha\mean{p_1, p_2}{\frac{p_1}{1+p_1v_1}\cdot\frac{p_2}{1+p_2v_2}}(\rho+\Delta-m_1-m_2+q_{1,2})
    \end{cases}.
\end{align}
Combining the equations for $q_i$ and $\hat{q}_i$ just derived, one can compute $q_i$ as
\begin{align}
    q_i = \frac{\frac{1}{\rho} m_i^2 + \alpha\mean{p_i}{\left(\frac{p_iv_i}{1+p_iv_i}\right)^2}(\rho+\Delta-2m_i)}{1-\alpha\mean{p_i}{\left(\frac{p_iv_i}{1+p_iv_i}\right)^2}},\quad \text{for $i=1,2$}
\end{align}
and similarly $q_{1,2}$ is given by
\begin{align}
    q_{1,2} = \frac{\frac{1}{\rho} m_1m_2 + \alpha\mean{p_1, p_2}{\frac{p_1v_1}{1+p_1v_1}\cdot\frac{p_2v_2}{1+p_2v_2}}(\rho+\Delta-m_1-m_2)}{1-\alpha\mean{p_1, p_2}{\frac{p_1v_1}{1+p_1v_1}\cdot\frac{p_2v_2}{1+p_2v_2}}}.
\end{align}

Let us collect these results in the following proposition:
\begin{proposition}\label{prop:ridge_scalar_overlaps}
    Consider two ridge estimators with sampling weights specified by $p_1, p_2$. The set of self-consistent equations in~\cref{eq:system_equations_ridge} gives a characterization of their overlaps in vector/matrix form for pair resampling. Using the notation
    \begin{equation}
        \mat{V}=\mathrm{Diag}(v_1, v_2),\quad
        \hat{\mat{V}}=\mathrm{Diag}(\hat{v}_1, \hat{v}_2),\quad
        \mat{Q}= \begin{bmatrix}
        q_1 & q_{1,2}\\
        q_{1,2} & q_2
    \end{bmatrix},\quad
    \hat{\mat{Q}}=\begin{bmatrix}
        \hat{q}_1 & \hat{q}_{1,2}\\
        \hat{q}_{1,2} & \hat{q}_2
    \end{bmatrix},
    \end{equation}
    the overlaps of interest can be simplified as follows: each $v_i$ is the unique solution to the fixed-point equation
\begin{equation}
    v_i = \frac{1}{\lambda + \alpha\mean{p_i}{\frac{p_i}{1+p_iv_i}}},
\end{equation}
while
\begin{align}
    m_i &= \rho(1-\lambda v_i),\\
    q_i &= \frac{\frac{1}{\rho} m_i^2 + \alpha\mean{p_i}{\left(\frac{p_iv_i}{1+p_iv_i}\right)^2}(\rho+\Delta-2m_i)}{1-\alpha\mean{p_i}{\left(\frac{p_iv_i}{1+p_iv_i}\right)^2}},\\
    q_{1,2} &= \frac{\frac{1}{\rho} m_1m_2 + \alpha\mean{p_1, p_2}{\frac{p_1v_1}{1+p_1v_1}\cdot\frac{p_2v_2}{1+p_2v_2}}(\rho+\Delta-m_1-m_2)}{1-\alpha\mean{p_1, p_2}{\frac{p_1v_1}{1+p_1v_1}\cdot\frac{p_2v_2}{1+p_2v_2}}},
\end{align}
where $\rho = \sfrac1d \|\wstar\|_2^2$ and $\Delta>0$.
\end{proposition}

\begin{remark}\label{remark:id_sampling_weights}
    When $p_1$ and $p_2$ are identically distributed according to some distribution $\mu$, we get $v_1=v_2\equiv v$, $m_1=m_2\equiv m$, and $q_1=q_2\equiv q$, with
\begin{align}\label{eq:system_equations_ridge_iid}
    \begin{cases}
        v &= \frac{1}{\lambda + \alpha\mean{p}{\frac{p}{1+pv}}}\\
        m &= \rho(1-\lambda v)\\
        q &= \frac{\frac{1}{\rho} m^2 + \alpha\mean{p}{\left(\frac{pv}{1+pv}\right)^2}(\rho+\Delta-2m)}{1-\alpha\mean{p}{\left(\frac{pv}{1+pv}\right)^2}},
    \end{cases}
\end{align}
where $p$ is a random variable distributed according to $\mu$.
\end{remark}
\begin{remark}\label{remark:indep_sampling_weights}
When $p_1, p_2$ are independent, the overlap $q_{12}$ can be simplified to
\begin{equation}
    q_{1,2} = \frac{\frac{1}{\rho} m_1m_2 + \alpha\mean{p_1}{\frac{p_1v_1}{1+p_1v_1}}\cdot\mean{p_2}{\frac{p_2v_2}{1+p_2v_2}}(\rho+\Delta-m_1-m_2)}{1-\alpha\mean{p_1}{\frac{p_1v_1}{1+p_1v_1}}\cdot\mean{p_2}{\frac{p_2v_2}{1+p_2v_2}}} = \frac{m_1m_2(\alpha\rho + \rho+\Delta-m_1-m_2)}{\alpha\rho^2 - m_1m_2}.
\end{equation}
\end{remark}

\paragraph{Residual Resampling} The system of equations for residual resampling in~\cref{eq:system_equations_residual_ridge} is almost identical to~\cref{eq:system_equations_ridge}, and in fact simpler as it does not involve expectations. Hence, following the same approach and notation as above, one can solve it to determine the overlaps of interests. 

\begin{proposition}\label{prop:ridge_residual_scalar_overlaps}
    Consider two ridge estimators. The set of self-consistent equations in~\cref{eq:system_equations_residual_ridge} gives a characterization of their overlaps in vector/matrix form for residual resampling. Using the notation
    \begin{equation}
        \mat{V}=\mathrm{Diag}(v_1, v_2),\quad
        \hat{\mat{V}}=\mathrm{Diag}(\hat{v}_1, \hat{v}_2),\quad
        \mat{Q}= \begin{bmatrix}
        q_1 & q_{1,2}\\
        q_{1,2} & q_2
    \end{bmatrix},\quad
    \hat{\mat{Q}}=\begin{bmatrix}
        \hat{q}_1 & \hat{q}_{1,2}\\
        \hat{q}_{1,2} & \hat{q}_2
    \end{bmatrix},
    \end{equation}
    the overlaps of interest are such that $v\equiv v_1=v_2, \: m\equiv m_1=m_2, \: q\equiv q_1=q_2$. In particular, $v$ is the unique solution to the fixed-point equation
\begin{equation}
    v = \frac{1}{\lambda + \frac{\alpha}{1+v}},
\end{equation}
while
\begin{align}
    m &= \rho(1-\lambda v),\\
    q &= \frac{\frac{1}{\rho} m^2 + \alpha\left(\frac{v}{1+v}\right)^2(\rho+\Delta-2m)}{1-\alpha\left(\frac{v}{1+v}\right)^2} = \frac{m^2(\alpha\rho +\rho+\Delta-2m)}{\alpha\rho^2-m^2},\\
    q_{1,2} &= \frac{\frac{1}{\rho} m^2 + \alpha\left(\frac{v}{1+v}\right)^2(\rho-2m)}{1-\alpha\left(\frac{v}{1+v}\right)^2} = \frac{m^2(\alpha\rho +\rho-2m)}{\alpha\rho^2-m^2},
\end{align}
where $\rho = \sfrac1d \|\wstar\|_2^2$ and $\Delta>0$.
\end{proposition}

\subsubsection{Full Resampling Overlaps}\label{sec:full_resampling_overlaps}
To compute overlaps between two independent learners performing ERM on their own dataset, we consider a single dataset of size $2n$ split evenly between the learners. This is achieved by using sampling weights $p_1, p_2$ with joint distribution given by $\mu(p_1, p_2) = \frac12\mathbbm{1}\{p_1=1, p_2=0\} + \frac12\mathbbm{1}\{p_1=0, p_2=1\}$. Since $p_1, p_2$ have the same marginals, \cref{remark:id_sampling_weights} applies. Note also that here we are in the high-dimensional regime with $\sfrac{2n}{d}\to2\alpha$. With this, the fixed-point equation for $v$ becomes $v = \frac{1}{\lambda + \frac{\alpha}{1+v}}$ and can be solved exactly. Overall, the overlaps are given by
\begin{align}
    \begin{cases}
        v &= \frac{1-\lambda-\alpha + \sqrt{(\alpha+\lambda -1)^2+4\lambda}}{2\lambda}\\
        m &= \rho(1-\lambda v)\\
        q &= \frac{\frac{1}{\rho}m^2 + \alpha\left(\frac{v}{1+v}\right)^2(\rho+\Delta-2m)}{1-\alpha\left(\frac{v}{1+v}\right)^2} = \frac{m^2(\alpha\rho +\rho+\Delta-2m)}{\alpha\rho^2-m^2}\\
        q_{1,2} &= \frac{m^2}{\rho}
    \end{cases}
\end{align}
by~\cref{prop:ridge_scalar_overlaps}.
In the following, we refer to these overlaps as $v_i^{\fr}, m_i^{\fr}, q_i^{\fr}$ and $q_{1,2}^{\fr}$.

\subsubsection{Residual Resampling Overlaps}\label{sec:residual_resampling_overlaps}
The overlaps are given by~\cref{prop:ridge_residual_scalar_overlaps}:
\begin{align}
    \begin{cases}
        v &= \frac{1-\lambda-\alpha + \sqrt{(\alpha+\lambda -1)^2+4\lambda}}{2\lambda}\\
        m &= \rho(1-\lambda v)\\
        q &= \frac{m^2(\alpha\rho +\rho+\Delta-2m)}{\alpha\rho^2-m^2}\\
        q_{1,2} &= \frac{m^2(\alpha\rho +\rho-2m)}{\alpha\rho^2-m^2}
    \end{cases}
\end{align}
In the following, we refer to these overlaps as $v_i^{\rr}, m_i^{\rr}, q_i^{\rr}$ and $q_{1,2}^{\rr}$.

\subsubsection{Subsampling Overlaps}\label{sec:subsampling_overlaps}
To compute overlaps between two independent learners that perform subsampling at rate $r_1, r_2$ of the same dataset, we must consider $p_1\sim\mathrm{Bern}(r_1)$ and $ p_2\sim\mathrm{Bern}(r_2)$ with $p_1$ independent of $p_2$. The fixed-point equations for $v_i$ become $v_i = \frac{1}{\lambda + \frac{\alpha r_i}{1+v_i}}$ and can be solved exactly to yield $v_i=\frac{1-\lambda-\alpha r_i + \sqrt{(\alpha r_i+\lambda -1)^2+4\lambda}}{2\lambda}$ for $i=1,2$. Note also that~\cref{remark:indep_sampling_weights} applies here. By~\cref{prop:ridge_scalar_overlaps}, we get
\begin{align}
    \begin{cases}
        v_i &= \frac{1-\lambda-\alpha r_i + \sqrt{(\alpha r_i+\lambda -1)^2+4\lambda}}{2\lambda}\\
        m_i &= \rho(1-\lambda v_i)\\
        q_i &= \frac{\frac{1}{\rho}m_i^2 + \alpha r_i\left(\frac{v_i}{1+v_i}\right)^2(\rho+\Delta-2m)}{1-\alpha r_i\left(\frac{v}{1+v}\right)^2} = \frac{m_i^2(\alpha\rho r_i +\rho+\Delta-2m_i)}{\alpha\rho^2 r_i-m_i^2}\\
        q_{1,2} &= \frac{m_1m_2(\alpha\rho + \rho+\Delta-m_1-m_2)}{\alpha\rho^2 - m_1m_2},
    \end{cases}
\end{align}
for $i=1,2$. In the following, we refer to these overlaps as $v_i^{\Ss}, m_i^{\Ss}, q_i^{\Ss}$ and $q_{1,2}^{\Ss}$.

\subsubsection{Pairs Bootstrap Overlaps}\label{sec:pairs_bootstrap_overlaps}
To compute overlaps between two independent learners that perform pairs bootstrap resampling of the same dataset, we must consider $p_1, p_2\stackrel{\text{i.i.d.}}{\sim}\text{Poi}(1)$, so that~\cref{remark:id_sampling_weights} and~\cref{remark:indep_sampling_weights} apply. By~\cref{prop:ridge_scalar_overlaps}, the overlaps are thus given by
\begin{align}\label{eq:bootstrap_overlaps}
    \begin{cases}
        v &= \frac{1}{\lambda + \alpha\mean{p}{\frac{p}{1+pv}}}\\
        m &= \rho(1-\lambda v)\\
        q &= \frac{\frac{1}{\rho}m^2 + \alpha\mean{p}{\left(\frac{pv}{1+pv}\right)^2}(\rho+\Delta-2m)}{1-\alpha\mean{p}{\left(\frac{pv}{1+pv}\right)^2}}\\
        q_{1,2} &= \frac{m^2(\alpha\rho + \rho+\Delta-2m)}{\alpha\rho^2 - m^2},
    \end{cases}
\end{align}
with $p\sim\mathrm{Poi}(1)$.
\begin{remark}
For $\lambda>0$, the variance is thus equal to
\begin{equation}
    \variancePairBootstrap = q-q_{1,2} = \frac{\frac{1}{\rho}m^2 + \alpha\mean{p}{\left(\frac{pv}{1+pv}\right)^2}(\rho+\Delta-2m)}{1-\alpha\mean{p}{\left(\frac{pv}{1+pv}\right)^2}} - \frac{m^2(\alpha\rho + \rho+\Delta-2m)}{\alpha\rho^2 - m^2},
\end{equation}
with $v$ and $m$ defined in~\cref{eq:bootstrap_overlaps}. Setting $\lambda=0$ (which only makes sense for $\alpha>1$), the variance becomes
    \begin{align}
    \variancePairBootstrap &= \frac{\rho + \alpha\mean{p}{\left(\frac{pv}{1+pv}\right)^2}(\Delta-\rho)}{1-\alpha\mean{p}{\left(\frac{pv}{1+pv}\right)^2}} - \frac{\alpha\rho - \rho+\Delta}{\alpha - 1}\\
    &= \Delta\left(\frac{\alpha\mean{p}{\left(\frac{pv}{1+pv}\right)^2}}{1-\alpha\mean{p}{\left(\frac{pv}{1+pv}\right)^2}}-\frac{1}{\alpha-1}\right)\\
    &= \Delta\left(\frac{1}{1-\alpha\mean{p}{\left(\frac{pv}{1+pv}\right)^2}}-\frac{\alpha}{\alpha-1}\right),
\end{align}
where $v$ is the unique solution to the fixed point equation $v = \frac{1}{\alpha\mean{p}{\frac{p}{1+pv}}}$. We thus recover Theorem 2 from~\cite{ElKaroui2018} since this is equivalent to writing
\begin{equation}
    \variancePairBootstrap = \Delta\left(\frac{\kappa}{1-\kappa-f(\kappa)}-\frac{1}{1-\kappa}\right),
\end{equation}
where $\kappa=\frac1\alpha$, $f(\kappa) := \mean{p}{\frac{1}{(1+pv)^2}}$, and $v$ is the unique solution of $\mean{p}{\frac{1}{1+pv}}=1-\kappa$.
\end{remark}
In the following, we refer to the overlaps as $v_i^{\pb}, m_i^{\pb}, q_i^{\pb}$ and $q_{1,2}^{\pb}$.

\subsubsection{Residual Bootstrap Overlaps}\label{sec:residual_bootstrap_overlaps}
To compute overlaps between two independent learners that perform bootstrap resampling, we follow the explanation in~\cref{appendix:residual_bootstrap}. It states that the overlaps for the residual bootstrap are given by those of the residual resampling, with $\rho$ replaced by $\Tilde{\rho} = q^{\fr}$ and $\Delta$ replaced by $\Tilde{\Delta} = \frac{\rho + \Delta - 2m^{\fr} + q^{\fr}}{(1+v^{\fr})^2}$. Hence, \Cref{prop:ridge_residual_scalar_overlaps} gives
\begin{align}\label{eq:residual_bootstrap_overlaps}
    \begin{cases}
        v &= \frac{1-\lambda-\alpha + \sqrt{(\alpha+\lambda -1)^2+4\lambda}}{2\lambda}\\
        m &= \Tilde{\rho}(1-\lambda v)\\
        q &= \frac{m^2(\alpha\Tilde{\rho} +\Tilde{\rho}+\Tilde{\Delta}-2m)}{\alpha\Tilde{\rho} ^2-m^2}\\
        q_{1,2} &= \frac{m^2(\alpha\Tilde{\rho} +\Tilde{\rho}-2m)}{\alpha\Tilde{\rho}^2-m^2}.
    \end{cases}
\end{align}
In the following, we refer to these overlaps as $v_i^{\rb}, m_i^{\rb}, q_i^{\rb}$ and $q_{1,2}^{\rb}$.

\subsubsection{Overlaps between Distinct Resampling Methods}\label{sec:other_overlaps}
Certain quantities of interest require to compute the correlation between two estimators which use different resampling methods. In the high-dimensional regime, this corresponds to the overlap $q_{1,2}$ where the sampling weights $p_1, p_2$ are independent. In that case, \cref{remark:indep_sampling_weights} applies and~\cref{prop:ridge_scalar_overlaps} yields
\begin{align}\label{eq:system_equations_ridge_iid_2}
    \begin{cases}
        v_i &= \frac{1}{\lambda + \alpha\mean{p_i}{\frac{p_i}{1+p_iv_i}}}\\
        m_i &= \rho(1-\lambda v_i)\\
        q_{12} &= \frac{m_1m_2(\alpha\rho + \rho+\Delta-m_1-m_2)}{\alpha\rho^2 - m_1m_2},
    \end{cases}
\end{align}
for $i=1,2$. In particular, the overlap between full resampling and pairs bootstrap is given by
\begin{equation}
    q_{1,2}^{\fr, \pb} := \frac{m^{\fr}m^{\pb}(\alpha\rho + \rho+\Delta-m^{\fr}-m^{\pb})}{\alpha\rho^2 - m^{\fr}m^{\pb}},
\end{equation}
the overlap between full resampling and subsampling at rate $r$ is given by
\begin{equation}
    q_{1,2}^{\fr, \Ss} := \frac{m^{\fr}m^{\Ss}(\alpha\rho + \rho+\Delta-m^{\fr}-m^{\Ss})}{\alpha\rho^2 - m^{\fr}m^{\Ss}}.
\end{equation}

\subsection{Large $\alpha$ rates}\label{appendix:large_alpha_rates}
In this section, we compute the rates of quantities of interest (variances, biases) in the $\alpha\to\infty$ limit, which are summarized in~\cref{table:large_alpha_rates}. The approach is mathematically standard: for each overlap, we compute its series expansion at $\alpha\to\infty$ up to a desired order. Let us illustrate this with an example. 

Consider the full resampling overlap $v^{\fr}$ computed in~\cref{sec:full_resampling_overlaps}:
\begin{equation}
    v^{\fr} = \frac{1-\lambda-\alpha + \sqrt{(\alpha+\lambda -1)^2+4\lambda}}{2\lambda}.
\end{equation}
To compute its series expansion at $\alpha\to\infty$, we substitute $\alpha$ with $\sfrac1\beta$ in the equation above, and then compute its Taylor series at $\beta\to0$. Letting
\begin{equation}
    h(\beta) \vcentcolon= \frac{1-\lambda-\frac1\beta + \sqrt{(\frac1\beta+\lambda -1)^2+4\lambda}}{2\lambda},
\end{equation}
one can apply this strategy and determine the Taylor expansion up to order 2 for $v^{\fr}$ by evaluating
\begin{align}
    \lim_{\beta\to0} h(\beta) &=\lim_{\beta\to0}\frac{\beta(1-\lambda)-1+\sqrt{(\beta(\lambda-1)+1)^2+4\lambda\beta^2}}{2\lambda\beta} = 0\\
    \lim_{\beta\to0} h^\prime(\beta) &= \lim_{\beta\to0} \frac{\frac{1}{\beta^2}-\frac{((\frac1\beta+\lambda -1)\frac{1}{\beta^2}}{\sqrt{(\frac1\beta+\lambda -1)^2+4\lambda}}}{2\lambda}=1\\
    \lim_{\beta\to0} h^{\prime\prime}(\beta) &= \lim_{\beta\to0}\frac{-\frac{2}{\beta ^3}+\frac{2 \left(\frac{1}{\beta }+\lambda -1\right)}{\beta ^3 \sqrt{\left(\frac{1}{\beta }+\lambda -1\right)^2+4 \lambda }}+\frac{1}{\beta ^4 \sqrt{\left(\frac{1}{\beta }+\lambda -1\right)^2+4 \lambda }}-\frac{\left(\frac{1}{\beta }+\lambda -1\right)^2}{\beta ^4 \left(\left(\frac{1}{\beta }+\lambda -1\right)^2+4 \lambda \right)^{3/2}}}{2 \lambda } = 2(1-\lambda),
\end{align}
from which we conclude that for $\beta\to0$,
\begin{equation}
    h(\beta) = h(\beta) + h^\prime(\beta)\beta +\frac12h^{\prime\prime}(\beta)\beta^2+O(\beta^3) = \beta + (1-\lambda)\beta^2+O(\beta^3)
\end{equation}
or equivalently, substituting back $\alpha=\sfrac{1}{\beta}$,
\begin{equation}
    v^{\fr} = \frac{1}{\alpha}+\frac{1-\lambda}{\alpha^2}+O\left(\frac{1}{\alpha^3}\right)
\end{equation}
for $\alpha\to\infty$. The computation of all overlaps are carried out in the same fashion, and we use the Mathematica software~\citep{Mathematica} to automate these computations.

\subsubsection{Full Resampling Rates}
From the overlaps computed in~\cref{sec:full_resampling_overlaps}, we retrieve the limiting behaviors
\begin{align}
    \begin{cases}
        v^{\fr}&\stackrel{\alpha\to\infty}{\simeq}\frac{1}{\alpha}+\frac{1-\lambda}{\alpha^2}+O\left(\frac{1}{\alpha^3}\right)\\
        m^{\fr}&\stackrel{\alpha\to\infty}{\simeq}\rho-\frac{\rho\lambda}{\alpha}+\frac{\rho\lambda(\lambda-1)}{\alpha^2}+O\left(\frac{1}{\alpha^3}\right)\\
        q^{\fr} &\stackrel{\alpha\to\infty}{\simeq} \rho+\frac{\Delta-2\lambda\rho}{\alpha}+\frac{\Delta(1-2\lambda)+\rho\lambda(3\lambda-2)}{\alpha^2}+O\left(\frac{1}{\alpha^3}\right)\\
        q_{1,2}^{\fr} &\stackrel{\alpha\to\infty}{\simeq} \rho-\frac{2\rho\lambda}{\alpha}+\frac{\rho\lambda(3\lambda-2)}{\alpha^2}+O\left(\frac{1}{\alpha^3}\right),
    \end{cases}
\end{align}
so that the variance is given by
\begin{equation}
    \varianceOnXY = q^{\fr}-q_{1,2}^{\fr} \stackrel{\alpha\to\infty}{\simeq} \frac{\Delta}{\alpha}+O\left(\frac{1}{\alpha^2}\right)
\end{equation}
and the bias is
\begin{equation}
    \biasOnXY = \rho+q_{1,2}^{\fr}-2m^{\fr} \stackrel{\alpha\to\infty}{\simeq} \frac{\rho\lambda^2}{\alpha^2}+O\left(\frac{1}{\alpha^3}\right).
\end{equation}

\subsubsection{Residual Resampling Rates}
From the overlaps computed in~\cref{sec:residual_resampling_overlaps}, we retrieve the limiting behaviors
\begin{align}
    \begin{cases}
        v^{\rr}&\stackrel{\alpha\to\infty}{\simeq}\frac{1}{\alpha}+\frac{1-\lambda}{\alpha^2}+O\left(\frac{1}{\alpha^3}\right)\\
        m^{\rr}&\stackrel{\alpha\to\infty}{\simeq}\rho-\frac{\rho\lambda}{\alpha}+\frac{\rho\lambda(\lambda-1)}{\alpha^2}+O\left(\frac{1}{\alpha^3}\right)\\
        q^{\rr} &\stackrel{\alpha\to\infty}{\simeq} \rho+\frac{\Delta-2\rho\lambda}{\alpha}+\frac{\Delta(1-2\lambda)+\lambda(3\lambda-2)}{\alpha^2}+O\left(\frac{1}{\alpha^3}\right)\\
        q_{1,2}^{\rr} &\stackrel{\alpha\to\infty}{\simeq} \rho-\frac{2\rho\lambda}{\alpha}+\frac{\rho\lambda(3\lambda-2)}{\alpha^2}+O\left(\frac{1}{\alpha^3}\right),
    \end{cases}
\end{align}
so that the variance is given by
\begin{equation}
    \varianceOnY = q^{\rr}-q_{1,2}^{\rr} \stackrel{\alpha\to\infty}{\simeq} \frac{\Delta}{\alpha}+O\left(\frac{1}{\alpha^2}\right)
\end{equation}
and the bias is
\begin{equation}
    \biasOnY = \rho+q_{1,2}^{\rr}-2m^{\rr} \stackrel{\alpha\to\infty}{\simeq} \frac{\rho\lambda^2}{\alpha^2}+O\left(\frac{1}{\alpha^3}\right).
\end{equation}

\subsubsection{Rates of Overlaps between Distinct Resampling Methods}
From the overlaps computed in~\cref{sec:other_overlaps}, we retrieve the limiting behaviors
\begin{align}
    \begin{cases}
        q_{1,2}^{\fr, \Ss} &\stackrel{\alpha\to\infty}{\simeq}\rho +\frac{r\Delta -\rho\lambda(r+1)}{r\alpha}+\frac{r^2\Delta + \rho\lambda(\lambda +r (\lambda +(\lambda -1) r)-1)-r\Delta  \lambda  (r+1)}{r^2\alpha^2}+O\left(\frac{1}{\alpha^3}\right)\\
        q_{1,2}^{\fr, \pb} &\stackrel{\alpha\to\infty}{\simeq} \rho +\frac{\Delta -2 \lambda  \rho }{\alpha }+\frac{\Delta(1-2  \lambda) +3 \rho\lambda(\lambda -1) }{\alpha ^2}+O\left(\frac{1}{\alpha^3}\right).
    \end{cases}
\end{align}

\subsubsection{Subsampling and Jackknife Rates}
From the overlaps computed in~\cref{sec:subsampling_overlaps}, we retrieve the limiting behaviors
\begin{align}
    \begin{cases}
        v_i^{\Ss}&\stackrel{\alpha\to\infty}{\simeq}\frac{1}{r_i\alpha}+\frac{1-\lambda}{r_i^2\alpha^2}+O\left(\frac{1}{\alpha^3}\right)\\
        m_i^{\Ss}&\stackrel{\alpha\to\infty}{\simeq}\rho-\frac{\rho\lambda}{r_i\alpha}+\frac{\rho\lambda(\lambda-1)}{r_i^2\alpha^2}+O\left(\frac{1}{\alpha^3}\right)\\
        q_i^{\Ss} &\stackrel{\alpha\to\infty}{\simeq} \rho+\frac{\Delta-2\rho\lambda}{r_i\alpha}+\frac{\Delta(1-2\lambda)+\rho\lambda(3\lambda-2)}{r_i^2\alpha^2}+O\left(\frac{1}{\alpha^3}\right)\\
        q_{1,2}^{\Ss} &\stackrel{\alpha\to\infty}{\simeq} \rho+\frac{\Delta r_1r_2 r-2\rho\lambda}{r_1r_2\alpha}+\frac{\Delta +\frac{(\lambda -1) \lambda  \rho }{r_1^2}+\frac{\lambda  (\lambda  \rho -\Delta r_2)}{r_1 r_2}+\frac{(\lambda -1) \lambda  \rho }{r_2^2}-\frac{\Delta  \lambda }{r_2}}{\alpha^2}+O\left(\frac{1}{\alpha^3}\right),
    \end{cases}
\end{align}
so that the variance when subsampling at rate $r_1=r_2\equiv r$ is given by
\begin{equation}
    \varianceSubsampling = \frac{q^{\Ss}-q_{1,2}^{\Ss}}{1-r} \stackrel{\alpha\to\infty}{\simeq} \frac{ \Delta}{\alpha  r}+O\left(\frac{1}{\alpha^2}\right).
\end{equation}
and the bias is
\begin{equation}
    \biasSubsampling = \frac{q_{1,2}^{\Ss} + q^{\fr}-2q_{1,2}^{\fr, \Ss}}{(1-r)^2} \stackrel{\alpha\to\infty}{\simeq} = \frac{\rho\lambda ^2}{\alpha ^2 r^2} + O\left(\frac{1}{\alpha^3}\right).
\end{equation}

The Jackknife variances and biases are computed by taking the limit $r\to 1$, and we get
\begin{equation}
    \varianceJackknife = \lim_{r\to 1}\frac{q^{\Ss}-q_{1,2}^{\Ss}}{1-r} \stackrel{\alpha\to\infty}{\simeq} \frac{ \Delta}{\alpha}+O\left(\frac{1}{\alpha^2}\right).
\end{equation}
and
\begin{equation}
    \biasJackknife = \lim_{r\to 1}\frac{q_{1,2}^{\Ss} + q^{\fr}-2q_{1,2}^{\fr, \Ss}}{(1-r)^2} \stackrel{\alpha\to\infty}{\simeq} = \frac{\rho\lambda ^2}{\alpha ^2} + O\left(\frac{1}{\alpha^3}\right).
\end{equation}

\subsubsection{Pairs Bootstrap Rates}\label{sec:pairs_bootstrap_rates}
The computation of rates in this case are less straightforward given that the overlaps depend on the evaluation of various expectations (see~\cref{sec:pairs_bootstrap_overlaps}). Let us consider $v^{\pb}$ first, which is given by the fixed-point equation
\begin{equation}
    v^{\pb} = \frac{1}{\lambda + \alpha\mean{p}{\frac{p}{1+pv^{\pb}}}}.
\end{equation}
We use the Ansatz that $v^{\pb}$ behaves as $\sfrac1\alpha$ in the $\alpha\to\infty$ limit, and hence write it as $v^{\pb}=\frac{\Tilde{v}}{\alpha}$. Since $\frac{1}{1+x}=1-x+O(x^2)$ for $x\to 0^+$, we get
\begin{align}\label{eq:v_pairs_bootstrap_approx}
    \Tilde{v} = \frac{\alpha}{\lambda + \alpha\mean{p}{\frac{p}{1+\frac{p\Tilde{v}}{\alpha}}}}\approx\frac{\alpha}{\lambda + \alpha\mean{p}{p(1-\frac{p\Tilde{v}}{\alpha})}} = \frac{\alpha}{\lambda + \alpha -2\Tilde{v}}.
\end{align}
This can be solved exactly and
\begin{equation}
    \Tilde{v} = \frac{\alpha+\lambda-\sqrt{(\alpha+\lambda)^2-8\alpha}}{4} \Rightarrow v^{\pb} = \frac{\alpha+\lambda-\sqrt{(\alpha+\lambda)^2-8\alpha}}{4\alpha}\stackrel{\alpha\to\infty}{\simeq}\frac{1}{\alpha}+\frac{2-\lambda}{\alpha^2}+O\left(\frac{1}{\alpha^3}\right).
\end{equation}
Overlaps $m^{\pb}$ and $q_{1,2}^{\pb}$ are thus given by
\begin{align}
    m^{\pb}&\stackrel{\alpha\to\infty}{\simeq}\rho-\frac{\rho\lambda}{\alpha}+\frac{\rho\lambda(\lambda-2)}{\alpha^2}+O\left(\frac{1}{\alpha^3}\right)\\
    q_{1,2}^{\pb}&\stackrel{\alpha\to\infty}{\simeq}\rho+\frac{\Delta-2\rho\lambda}{\alpha}+\frac{\Delta(1-2\lambda) +\rho\lambda(3\lambda-4)}{\alpha^2}+O\left(\frac{1}{\alpha^3 }\right).
\end{align}
Overlap $q^{\pb}$ involves the evaluation of  $\mean{p}{\left(\frac{pv^{\pb}}{1+pv^{\pb}}\right)^2}$, which can be computed using the same approximation as in~\cref{eq:v_pairs_bootstrap_approx}:
\begin{align}
    \mean{p}{\left(\frac{pv^{\pb}}{1+pv^{\pb}}\right)^2} &\approx \mean{p}{\left(pv^{\pb}(1-pv^{\pb})\right)^2} \\
    &= \mean{p}{(pv^{\pb})^2-2(pv^{\pb})^3+(pv^{\pb})^4}\\
    &=2(v^{\pb})^2-10(v^{\pb})^3+15(v^{\pb})^4,
\end{align}
where the last equality is obtained since $p\sim\Pois(1)$. This yields
\begin{equation}
    q^{\pb}\stackrel{\alpha\to\infty}{\simeq}1+\frac{2(\Delta -\rho\lambda)}{\alpha}+ \frac{2\Delta(1-2\lambda)+\rho\lambda(3\lambda-4)}{\alpha^2} + O\left(\frac{1}{\alpha^3 }\right).
\end{equation}
so that the variance in the $\alpha\to\infty$ limit is thus given by
\begin{equation}
    \variancePairBootstrap = q^{\pb} - q_{1,2}^{\pb}\stackrel{\alpha\to\infty}{\simeq}\frac{\Delta}{\alpha }+O\left(\frac{1}{\alpha^2}\right)
\end{equation}
and the bias is
\begin{equation}
    \biasPairBootstrap = q_{1,2}^{\pb} + q^{\fr}-2q_{1,2}^{\fr, \pb} \stackrel{\alpha\to\infty}{\simeq} = \frac{\rho\lambda^2}{\alpha^4} + O\left(\frac{1}{\alpha^5}\right).
\end{equation}

\subsubsection{Residual Bootstrap Rates}
From the overlaps computed in~\cref{sec:residual_bootstrap_overlaps}, we retrieve the limiting behaviors
\begin{align}
    \begin{cases}
        v^{\rb} &\stackrel{\alpha\to\infty}{\simeq}\frac{1}{\alpha}+\frac{1-\lambda}{\alpha^2}+O\left(\frac{1}{\alpha^3}\right)\\
        m^{\rb} &\stackrel{\alpha\to\infty}{\simeq}\rho+\frac{\Delta-3\rho\lambda}{\alpha}+\frac{\Delta(1-3\lambda) +3\rho\lambda(2\lambda-1)}{\alpha^2}+O\left(\frac{1}{\alpha^3}\right)\\
        q^{\rb} &\stackrel{\alpha\to\infty}{\simeq}\rho +\frac{2 (\Delta -2 \lambda  \rho )}{\alpha }+\frac{\Delta(1-6\lambda) +2 \rho\lambda  (5 \lambda -2)}{\alpha ^2}+O\left(\frac{1}{\alpha ^3}\right)\\
        q_{1,2}^{\rb} &\stackrel{\alpha\to\infty}{\simeq} \rho+\frac{\Delta-4\rho\lambda}{\alpha}+\frac{\Delta(1-4\lambda)+2\rho\lambda(5\lambda-2)}{\alpha^2}+O\left(\frac{1}{\alpha^3}\right),
    \end{cases}
\end{align}
so that the variance is
\begin{equation}
    \varianceResidualBootstrap = q^{\rb}-q_{1,2}^{\rb} \stackrel{\alpha\to\infty}{\simeq} \frac{\Delta}{\alpha }+O\left(\frac{1}{\alpha^2}\right)
\end{equation}
and the bias is
\begin{equation}
    \biasResidualBootstrap = q_{1,2}^{\rb} + q^{\fr}-2m^{\rb} \stackrel{\alpha\to\infty}{\simeq} \frac{\rho\lambda^2}{\alpha ^2}+O\left(\frac{1}{\alpha^3}\right).
\end{equation}

\subsubsection{Differences between Rates}
Recall that pairs bootstrap and subsampling aim to estimate bias and variance with respect to the joint distribution $p_{\theta}(y,\vec{x})$, while residual bootstrap seeks to estimate the bias and variance with respect to the conditional distribution $p_{\theta}(y|\vec{x})$. To understand how good each estimate of the bias and variance is, we compute for each resampling method the difference between their estimate and the true value. For the variances, this results in
\begin{align*}
    \left|\varianceSubsampling - \varianceOnXY\right|  &\stackrel{\alpha\to\infty}{\simeq} \frac{\Delta(1-r)}{\alpha  r}+\frac{\Delta \left((1-2 \lambda)(1- r^2)+r\right)}{\alpha ^2 r^2}+O\left(\frac{1}{\alpha^3}\right)\\
    \left|\varianceJackknife - \varianceOnXY\right|  &\stackrel{\alpha\to\infty}{\simeq} \frac{\Delta}{\alpha ^2}+O\left(\frac{1}{\alpha^3}\right)\\
    \left|\variancePairBootstrap - \varianceOnXY\right|  &\stackrel{\alpha\to\infty}{\simeq} \frac{\Delta(4 \lambda +7)}{\alpha ^3}+O\left(\frac{1}{\alpha^4}\right)\\
    \left|\varianceResidualBootstrap - \varianceOnY\right|  &\stackrel{\alpha\to\infty}{\simeq} \frac{\Delta}{\alpha ^2}+O\left(\frac{1}{\alpha^3}\right)
\end{align*}
while the biases are given by
\begin{align*}
    \left|\biasSubsampling - \biasOnXY\right|  &\stackrel{\alpha\to\infty}{\simeq} \frac{\rho\lambda^2(r^2-1)}{r^2\alpha^2} + \frac{\lambda ^2 \left(\rho  \left(2 \lambda -2 (\lambda -1) r^3-(3-2 \lambda ) r-2\right)-\Delta  r\right)}{r^3\alpha^3}+O\left(\frac{1}{\alpha^4}\right)\\
    \left|\biasJackknife - \biasOnXY\right|  &\stackrel{\alpha\to\infty}{\simeq} \frac{\lambda^2(\rho(2\lambda-3)-\Delta)}{\alpha^3}+O\left(\frac{1}{\alpha^4}\right)\\
    \left|\biasPairBootstrap - \biasOnXY\right|  &\stackrel{\alpha\to\infty}{\simeq} \frac{\rho\lambda^2}{\alpha^2}+O\left(\frac{1}{\alpha^3}\right)\\
    \left|\biasResidualBootstrap - \biasOnY\right|  &\stackrel{\alpha\to\infty}{\simeq} \frac{\lambda^2(2\lambda\rho-\Delta)}{\alpha ^3}+O\left(\frac{1}{\alpha^4}\right).
\end{align*}

\clearpage

\section{Asymptotics of prediction variance}
\label{appendix:other_variances}
The focus of our work is the variance of estimators with respect to the resampling of the training set. However, one can also be interested in computing the \textit{prediction variance}, often defined as 
\begin{equation}
    \Var_{\Vec{x}, y} \left( y - \hat{y}\left( \Vec{x}  \right) \right)
    \label{eq:preditive_variance}
\end{equation}
where now the training set is fixed, and the variance is taken with respect to the new test sample $\Vec{x}, y$. In a linear model where $\hat{y} = \werm^{\top} \Vec{x}$ and in our setting defined in~\cref{eq:def_model}, the prediction variance is equal to the test error of the ERM estimator. Indeed : 
\begin{align}
    \Var_{\Vec{x}, y} \left( y - \hat{y}\left( \Vec{x}  \right) | \dataset \right) &= \mathbb{E} \left[ ( y - \werm^{\top} \Vec{x} )^2 \right] + \mathbb{E} \left[ ( y - \werm^{\top} \Vec{x}) \right]^2 \\
    &= \mathbb{E} \left[ ( y - \werm^{\top} \Vec{x} )^2 \right] = \varepsilon_g
\end{align}
because $\mathbb{E} \left[ ( y - \werm^{\top} \Vec{x}) \right]^2 = 0$. In the case of Ridge regression,
\begin{equation}
    \varepsilon_g = \rho - 2 m^{\fr} + Q_{11}^{\fr} + \sigma^2.
\end{equation}
Note that at optimal $\lambda = \sigma^2$ ($\lambda = 1$ in our case), the performance of the ERM estimator is equal the posterior variance of the Bayes-optimal, as 
\begin{align}
\varianceBO &= \rho - q^{\bo} \\
&= \rho - 2 m^{\bo} + q^{\bo} \label{eq:nishimori_step}\\
&= \rho - 2 m^{\fr} + Q_{11}^{\fr},\label{eq:optimal_lambda_same_as_bo}
\end{align}
where~\cref{eq:nishimori_step} follows from the \textit{Nishimori condition} $m^{\bo} = q^{\bo}$, and~\cref{eq:optimal_lambda_same_as_bo} is due to the fact that $\werm = \mathbb{E} \left[ \Vec{\theta} | \dataset \right]$ for optimal $\lambda$.

\section{Additional Details for Numerical Experiments}
\label{appendix:numerics}
The state evolution equations for the resampling methods are written in the Julia language \citep{bezansonJuliaFreshApproach2017} and are available on the Github repository \texttt{https://github.com/SPOC-group/BootstrapAsymptotics} that also contains the code used to reproduce the plots. The code leverages libraries such as \texttt{NLSolvers.jl} for optimization \citep{mogensenOptimMathematicalOptimization2018}, \texttt{QuadGK.jl}  and \texttt{HCubature.jl} for integration \citep{johnsonQuadGKJlGauss2013,johnsonHCubatureJlPackage2017,genzRemarksAlgorithm0061980}, \texttt{MLJLinearModels.jl} for estimation of GLMs \citep{JuliaAIMLJLinearModelsJl2023}, as well as various utilities for statistical functions \citep{JuliaStatsStatsFunsJl2024,JuliaStatsLogExpFunctionsJl2023}, performance \citep{JuliaArraysStaticArraysJl2024} and plotting \citep{breloffPlotsJl2024}.
The code to compute the posterior variance of the Bayes-optimal estimator is written in Rust and is available at \texttt{https://github.com/spoc-group/double\_descent\_uncertainty}. All the experiments were run on a computer with the following specifications: 16 GB RAM, Apple M1 Pro CPU.

\subsection{Effects of finite $B$}
In~\cref{sec:discussions}, we studied the behavior of resampling methods in the limit $B \to \infty$. However, in practice $B$ is usually not very large, and the finiteness of $B$ has an impact on the estimated bias and variances. Indeed : 
\begin{align*}
    \widehat{\Var} &= \frac{1}{dB}\sum\limits_{b=1}^{B}\left\lVert \hatw_{b}-
    \frac{1}{B}\sum\limits_{b=1}^{B}\hatw_{b}\right\lVert^{2} = \frac{1}{dB} \sum_{b = 1}^B \| \hatw_b - \mathbb{E}_{\dataset^{\star}} \left[ \hatw \right]\|^2 + \frac{1}{d}\| \mathbb{E}_{\dataset^{\star}} \left[ \hatw \right] - \frac{1}{B} \sum_{b = 1}^B \hatw_b \|^2
\end{align*}
where second term vanishes as $B \to \infty$. Note that our framework allows us to compute the $\widehat{\Var}(B)$ for a finite number of Bootstrap resamples $B$, as we get asymptotically 
\begin{equation*}
    \widehat{\Var}(B) = \frac{B - 1}{B} \lim_{B \to \infty} \widehat{\Var}
\end{equation*}
where $\widehat{\Var}$ is the variance plotted in~\cref{fig:variance_ridge} and \cref{fig:variance_logistic}.

Likewise, the estimator of the bias with finite $B$ can be computed and equates 
\begin{align*}
    \widehat{\Bias}(B) = \widehat{\Bias} + \frac{1}{B} \widehat{\Var}
\end{align*}
where $\frac{1}{B} \widehat{\Var}$ is due to finite sampling and vanishes as $B \to \infty$. Note that the overlaps computed with our state-evolution equations allow us to compute $\widehat{\Bias}(B)$ at any $B$.

\end{document}